\documentclass{article}
\pdfoutput=1

\usepackage[final]{neurips_2022}

\usepackage[utf8]{inputenc} 
\usepackage[T1]{fontenc}    
\usepackage{url}            
\usepackage{booktabs}       
\usepackage{amsfonts}       
\usepackage{nicefrac}       
\usepackage{microtype}      
\usepackage{xcolor}         

\usepackage[pagebackref=true,breaklinks=true,letterpaper=true,colorlinks,bookmarks=false]{hyperref}

\usepackage{multirow}
\usepackage{boldline}
\usepackage{comment}
\usepackage{graphicx}
\usepackage{subfigure}
\usepackage{enumitem}
\usepackage{natbib}
\setcitestyle{square,numbers}
\usepackage{amsmath}
\usepackage{graphicx}
\usepackage{array}
\usepackage{arydshln} 
\usepackage{wrapfig}
\usepackage{lipsum}
\usepackage{adjustbox}
\usepackage{caption}

\newcommand{\ie}{\textit{i}.\textit{e}.}
\newcommand{\eg}{\textit{e}.\textit{g}.}
\newcommand{\etc}{\textit{etc}.}

\usepackage{color,soul}
\newcolumntype{C}[1]{>{\centering\arraybackslash}p{#1}}

\title{Promising or Elusive? Unsupervised Object Segmentation from Real-world Single Images}

\author{%
Yafei Yang \quad Bo Yang \vspace{0.2cm}  \\
vLAR Group, The Hong Kong Polytechnic University\\
\tt \small{ya-fei.yang@connect.polyu.hk \quad bo.yang@polyu.edu.hk}\\
}

\begin{document}

\maketitle

\begin{abstract}
In this paper, we study the problem of unsupervised object segmentation from single images. We do not introduce a new algorithm, but systematically investigate the effectiveness of existing unsupervised models on challenging real-world images. We firstly introduce four complexity factors to quantitatively measure the distributions of object- and scene-level biases in appearance and geometry for datasets with human annotations. With the aid of these factors, we empirically find that, not surprisingly, existing unsupervised models catastrophically fail to segment generic objects in real-world images, although they can easily achieve excellent performance on numerous simple synthetic datasets, due to the vast gap in objectness biases between synthetic and real images. By conducting extensive experiments on multiple groups of ablated real-world datasets, we ultimately find that the key factors underlying the colossal failure of existing unsupervised models on real-world images are the challenging distributions of object- and scene-level biases in appearance and geometry. Because of this, the inductive biases introduced in existing unsupervised models can hardly capture the diverse object distributions. Our research results suggest that future work should exploit more explicit objectness biases in the network design.

\end{abstract}

\section{Introduction}
The capability of automatically identifying individual objects from complex visual observations is a central aspect of human intelligence \cite{Spelke1992}. It serves as the key building block for higher-level cognition tasks such as planning and reasoning \cite{Greff2020}. In last years, a plethora of models have been proposed to segment objects from single static images in an unsupervised fashion: from the early AIR \cite{Eslami2016} and MONet \cite{Burgess2019} to the recent SPACE \cite{Lin2020}, SlotAtt \cite{Locatello2020}, GENESIS-V2 \cite{Engelcke2021}, \etc{} They jointly learn to represent and segment multiple objects from a single image, without needing any human annotations in training. This process is often called perceptual grouping/binding or object-centric learning. These methods and their variants have achieved impressive segmentation results on numerous synthetic scene datasets such as dSprites \cite{Matthey2017} and CLEVR \cite{Johnson2017}. Such advances come with great expectations that the unsupervised techniques would likely close the gap with fully-supervised methods for real-world visual understanding. However, few work has systematically investigated the true potential of the emerging unsupervised object segmentation models on complex real-world images such as COCO dataset \cite{Lin2014}. This naturally raises an essential question:

\textit{Is it promising or even possible to segment generic objects from real-world single images using (existing) unsupervised methods?} 

\textbf{What is an object?} 
To answer the above question involves another fundamental question: what is an object? Exactly 100 years ago in Gestalt psychology, Wertheimer \cite{Wertheimer1923} first introduced a set of Gestalt principles such as proximity, similarity and continuation to heuristically define visual data as objects. However, these factors are highly subjective, whilst the real-world generic objects are far more complex with extremely diverse appearances and shapes. Therefore, it is practically impossible to quantitatively define what is an object, \ie{}, the objectness, from visual inputs (\eg{}, a set of image pixels). Nevertheless, to thoroughly understand whether unsupervised methods can truly learn objectness akin to the psychological process of humans, it is vital to investigate the underlying factors that potentially facilitate or otherwise hinder the ability of unsupervised models. In this regard, by drawing on Gestalt principles, we instead define a series of new factors to quantitatively measure the complexity of objects and scenes in Section \ref{sec:complexity_factors}. By taking into account both the appearance and geometry of objects and scenes, our complexity factors explicitly assess the difficulty of segmenting a specific object. For example, a chair with colorful textures tend to have higher complexity than a single-color ball for unsupervised methods. With the aid of these factors, we extensively study whether and how existing unsupervised models can discover objects in Section \ref{sec:experimental_res}. 

\textbf{What is the problem of unsupervised object segmentation from single images?} A large number of models \cite{Yuan2022} aim to tackle the problem of unsupervised object segmentation from single images. They share several key problem settings: 1) all training images do not have any human annotations; 2) every single image has multiple objects; 3) each image is treated as a static data point without any dynamic or temporal information; 4) all models are trained from scratch without requiring any pretrained networks on additional datasets. Ultimately, the goal of these models is to segment all individual objects as accurate as the ground truth human annotations. In this paper, we regard these settings as the basic and necessary part of unsupervised object segmentation from single images, and empirically evaluate how successfully the existing models can exhibit on real-world images.

\begin{figure}[t]
\setlength{\abovecaptionskip}{ 0 pt}
\setlength{\belowcaptionskip}{ -6 pt}
\centering
   \includegraphics[width=0.7\linewidth]{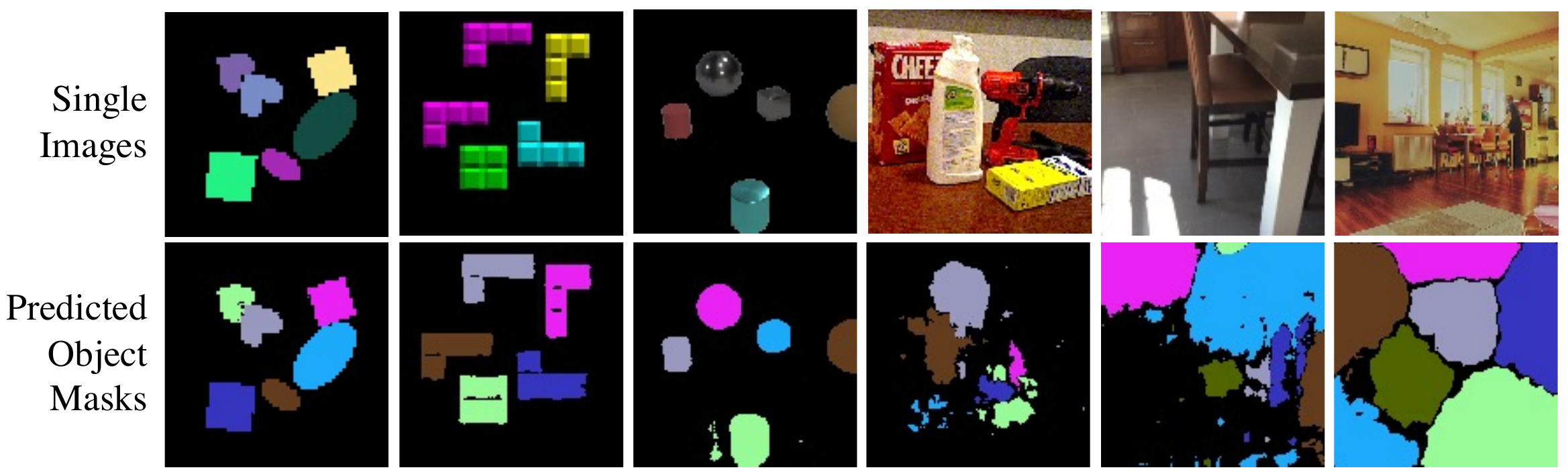}
\caption{The failure of SlotAtt \cite{Locatello2020} on three real-world images (right-hand side), although it can perfectly segment simple objects on three synthetic images (left-hand side).}
\label{fig:opening}
\vspace{-0.2cm}
\end{figure}

\textbf{Contributions and findings.} This paper addresses the essential question regarding the potential of unsupervised segmentation of generic objects from real-world single images. Our contributions are:
\begin{itemize}[leftmargin=*]
\setlength{\itemsep}{1pt}
\setlength{\parsep}{1pt}
\setlength{\parskip}{1pt}
    \item We firstly introduce 4 complexity factors to quantitatively measure the difficulty of objects and scenes. These factors are key to investigate the true potential of existing unsupervised models.
    \item We extensively evaluate current unsupervised approaches in a large-scale experimental study. We implement 4 representative methods and train more than 130 models on 6 curated datasets from scratch. The datasets, code and pretrained models are available at \url{https://github.com/vLAR-group/UnsupObjSeg}   
    \item We analyze our experimental results and find that: 1) existing unsupervised object segmentation models cannot discover generic objects from single real-world images, although they can achieve outstanding performance on synthetic datasets, as qualitatively illustrated in Figure \ref{fig:opening}; 2) the challenging distributions of both object- and scene-level biases in appearance and geometry from real-world images are the key factors incurring the failure of existing models; 3) the inductive biases introduced in existing unsupervised models are fundamentally not matched with the objectness biases exhibited in real-world images, and therefore fail to discover the real objectness.  
\end{itemize}

\textbf{Related Work.} Recently, ClevrTex {\cite{Karazija2021}} and the concurrent work {\cite{Papa2022}} also study unsupervised object segmentation on single images. Through evaluation on (complex) synthetic datasets only, both works focus on benchmarking the effectiveness of particular network designs of baselines. By comparison, our paper aims to explore what and how the objectness distribution gaps between synthetic and real-world images incur the failure of existing models. The recent work {\cite{Weis2021}} which investigates video object discovery is orthogonal to our work as the motion signals do not exist in single images.     

\textbf{Scope of this research.} This paper does not investigate unsupervised object discovery on saliency maps \cite{Wang2021}, static multi-views or dynamic videos \cite{Yuan2022}. Recent methods \cite{Caron2021,Henaff2022} requiring pretrained models on monolithic object images such as ImageNet \cite{Russakovsky2015} are not evaluated as well.

\clearpage
\section{Complexity Factors}\label{sec:complexity_factors}

\setlength{\columnsep}{10pt}
\begin{wrapfigure}{R}{0.35\textwidth}
\setlength{\abovecaptionskip}{ -1 pt}
\setlength{\belowcaptionskip}{ -8 pt}
\raisebox{0pt}[\dimexpr\height-1.\baselineskip\relax]{
   \centering
   \includegraphics[width=1\linewidth]{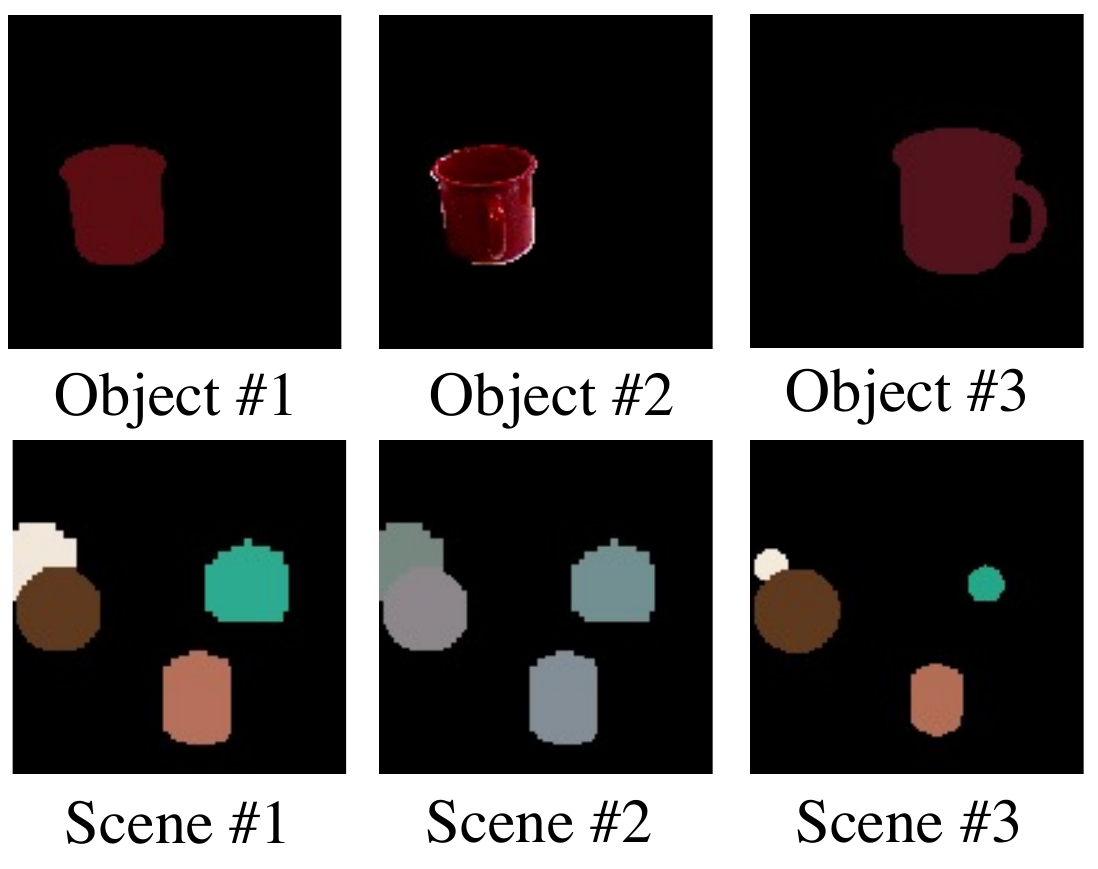}
}
\caption{\small{Complexity in appearance and geometry for objects and scenes.}}
\label{fig:complexity_factors}
\end{wrapfigure}

As illustrated in the top row of Figure \ref{fig:complexity_factors}, an individual object, represented by a set of color pixels painted within a mask, can vary significantly given different types of appearance and geometric shape. A specific scene, represented by a set of objects placed within an image, can also differ vastly given different types of relative appearance and geometric layout between objects, as illustrated in the bottom row. Unarguably, such variation and complexity of appearance and geometry in both object level and scene level directly affects human's ability to precisely separate all objects. Naturally, the performance of unsupervised segmentation models are also expected to be influenced by the variation. In this regard, we carefully define the following two groups of factors to quantitatively describe the complexity of different datasets.  

\subsection{Object-level Complexity Factors}
As to a specific object, all its information can be described by appearance and geometry. Therefore we define the below two factors to measure the complexity of appearance and geometry respectively. Notably, both factors are nicely invariant to the object scale. 

\begin{itemize}[leftmargin=*]
\setlength{\itemsep}{1pt}
\setlength{\parsep}{1pt}
\setlength{\parskip}{1pt}
    \item \textbf{Object Color Gradient:} This factor aims to calculate how frequently the appearance changes within the object mask. In particular, given the RGB image and mask of an object, we firstly convert RGB into grayscale and then apply Sobel filter \cite{Sobel1973} to compute the gradients horizontally and vertically for each pixel within the mask. The final gradient value is obtained by averaging out all object pixels. Note that, the object boundary pixels are removed to avoid the interference of background. Numerically, the higher this factor is, the more complex texture and/or lighting effect the object has, and therefore it is likely harder to segment.
    \item \textbf{Object Shape Concavity:} This factor is designed to evaluate how irregular the object boundary is. Particularly, given an object (binary) mask, denoted as $\boldsymbol{M}_{obj}\in \mathbb{R}^{H\times W}$, we firstly find the smallest convex polygon mask ($\boldsymbol{M}_{cvx} \in \mathbb{R}^{H\times W}$) that surrounds the object mask using an existing algorithm \cite{Eddins2011}, and then the object shape concavity value is computed as: $1-\sum\boldsymbol{M}_{obj} / \sum\boldsymbol{M}_{cvx}$. Clearly, the higher this factor is, the more irregular object shape is, and segmentation is more tricky.   
\end{itemize}

\subsection{Scene-level Complexity Factors}
As to a specific image, in addition to the object-level complexity, the spatial and appearance relationships between all objects can also incur extra difficulty for segmentation. We define the following two factors to quantify the complexity of relative appearance and geometry between objects in an image. 

\begin{itemize}[leftmargin=*]
\setlength{\itemsep}{1pt}
\setlength{\parsep}{1pt}
\setlength{\parskip}{1pt}
    \item \textbf{Inter-object Color Similarity:} This factor intends to assess the appearance similarity between all objects in the same image. Specifically, we firstly calculate the average color for each object, and then compute the pair-wise Euclidean distances of object colors, obtaining a $K\times K$ matrix where $K$ represents the object number. The \textit{average color distance} is calculated by averaging the matrix excluding diagonal entries, and the final inter-object color similarity is computed as: $1-$\textit{average color distance}$/(255\times\sqrt{3})$. Intuitively, the higher this factor is, the more similar all objects appear to be, the less distinctive each object is, and it is harder to separate each object.
    \item \textbf{Inter-object Shape Variation:} This factor aims to measure the relative geometry diversity between all objects in the image. We firstly calculate the diagonal length of bounding box for each object, and then compute the pair-wise absolute differences for all object diagonal lengths, obtaining a $K\times K$ matrix. The final inter-object shape variation is the average of the matrix excluding diagonal entries. The higher this factor, the objects within an image have more diverse and imbalanced sizes, and therefore segmenting both gigantic and tiny objects is likely more challenging. 
\end{itemize}

\begin{figure}[t]
\setlength{\abovecaptionskip}{ 2 pt}
\setlength{\belowcaptionskip}{ -8 pt}
\centering
   \includegraphics[width=0.7\linewidth]{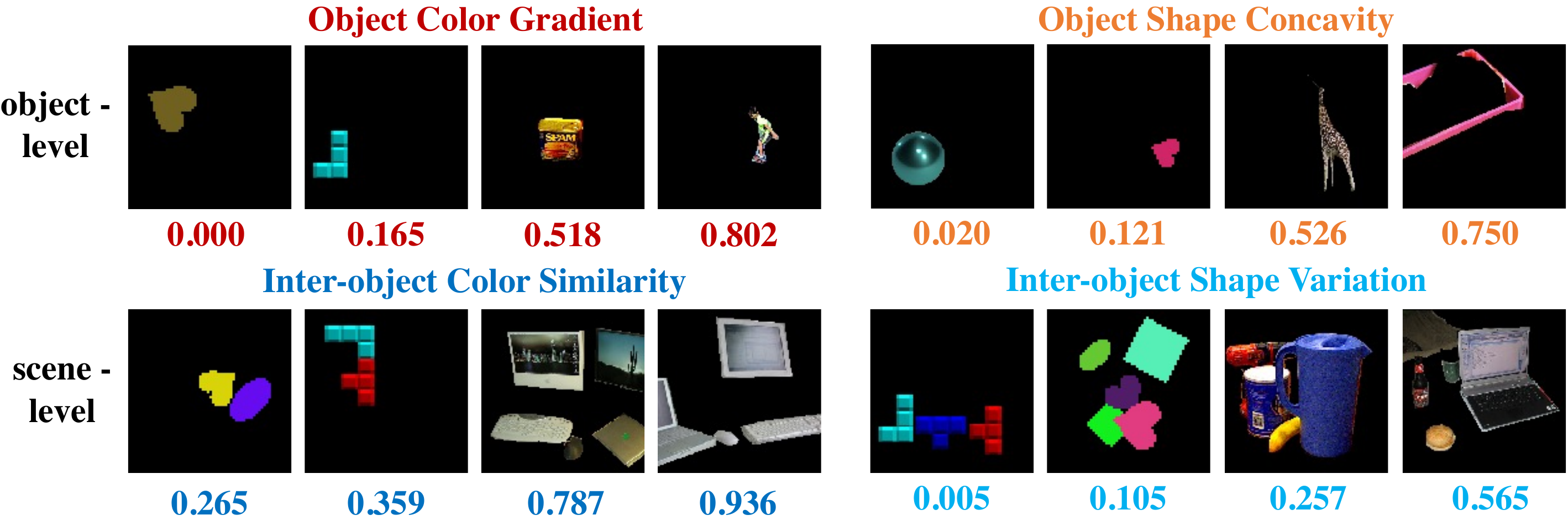}
\caption{Sample objects and scenes for the four factors at different complexity values. All complexity values are normalized to the range of $[0,1]$.}
\label{fig:factors_illustration}
\vspace{-0.3cm}
\end{figure}

By capturing the appearance and geometry in both object and scene levels, the four factors are designed to quantify the complexity of objects and images. For illustration, Figure {\ref{fig:factors_illustration}} shows sample images for the four factors at different values. The higher the values, the more complex the objects and scenes. In fact, these factors are carefully selected from more than 10 candidates because they are empirically more suitable to differentiate the gaps between synthetic and real-world 
images, and they eventually serve as key indicators to diagnose existing unsupervised models in Section \ref{sec:experimental_res}. 
Calculation details of the four factors and other candidates are in appendix. 

\section{Experimental Design}
\subsection{Considered Methods}
A range of works have explored unsupervised object segmentation in recent years. They are typically formulated as (variational) autoencoders (AE/VAE) \cite{Kingma2014} or generative adversarial networks (GAN) \cite{Goodfellow2014}. GAN based models \cite{Chen2019,Arandjelovic2019,Bielski2019,VanSteenkiste2020,Azadi2020,Voynov2021,Abdal2021} are usually limited to identifying a single foreground object and can hardly discover multiple objects due to the training instabilities, therefore not considered in this paper. As shown in Table \ref{tab:existing_models}, the majority of existing models are based on AE/VAE and can be generally divided into two groups according to the object representation: 
\begin{itemize}[leftmargin=*]
\setlength{\itemsep}{1pt}
\setlength{\parsep}{1pt}
\setlength{\parskip}{1pt}
\item \textbf{Factor-based models}: Each object is represented by explicit factors such as size, position, appearance, \etc{}, and the whole image is a spatial organization of multiple objects. Basically, such representation explicitly enforces objects to be bounded within particular regions. 
\item \textbf{Layer-based models}: Each object is represented by an image layer, \ie{}, a binary mask, and the whole image is a spatial mixture of multiple object layers. Intuitively, this representation does not have strict spatial constrains, and instead is more flexible to cluster similar pixels as objects. 
\end{itemize}

In order to decompose the input images into objects, these approaches introduce different types of network architecture, loss functions, and regularization terms as inductive biases. These biases broadly include: 1) variational encoding which encourages the disentanglement of  latent variables; 2) iterative inference which likely ends up with better scene representations over occlusions; 3) object relationship regularization such as depth estimation and autoregressive prior which aims at capturing the dependency of multiple objects; and many other biases. With different combinations of these biases, many methods have shown outstanding performance in synthetic datasets. Among them, we select 4 representative models for our investigation: 1) AIR \cite{Eslami2016}, 2) MONet \cite{Burgess2019}, 3) IODINE \cite{Greff2019}, and 4) SlotAtt \cite{Locatello2020}. We also add the fully-supervised Mask R-CNN {\cite{He2017a}} as an additional baseline for comprehensive comparison. 
Implementation details are provided in appendix.

\begin{table}
\centering
\setlength{\abovecaptionskip}{2 pt}
\footnotesize
\caption{Existing unsupervised models for object segmentation on single images. Each model includes different inductive biases, such as variational autoencoding (VAE), iterative inference (Iter), object relationship regularization (Rel), \etc{}}
\tabcolsep= 0.18cm 
\begin{tabular}{rl|ccc||rl|ccc} \hline
\multicolumn{2}{c}{\multirow{2}{*}{Factor-based Models}} & \multicolumn{3}{c||}{Inductive Biases} & \multicolumn{2}{c}{\multirow{2}{*}{Layer-based Models}} & \multicolumn{3}{c}{Inductive Biases} \\
\multicolumn{2}{c}{}                                     & VAE & Iter & Rel                     & \multicolumn{2}{c}{}                    & VAE & Iter & Rel  \\ \hlineB{3}
CST-VAE \cite{Huang2016}   & ICLRW'16                    &\checkmark& &                      & Tagger \cite{Greff2016} & NIPS'16       & & \checkmark & \\  
AIR \cite{Eslami2016}      & NIPS'16                     &\checkmark& &                      & RC \cite{Greff2016} & ICLRW'16          & & \checkmark & \\  
SPAIR \cite{Crawford2019}  & AAAI'19                     &\checkmark& &                      & NEM \cite{Greff2017} & NIPS'17          & & \checkmark & \\
SuPAIR \cite{Stelzner2019} & ICML'19                     &\checkmark& &                      & MONet \cite{Burgess2019} & arXiv'19     & \checkmark & & \\  
GMIO  \cite{Yuan2019}      & ICML'19                     &\checkmark&\checkmark&           & IODINE \cite{Greff2019} & ICML'19       & \checkmark & \checkmark & \\ 
ASR  \cite{Xu2019}         & NeurIPS'19                  &\checkmark& &                      & ECON \cite{VonKugelgen2020} & ICLRW'20  & \checkmark & & \checkmark \\ 
SPACE \cite{Lin2020}       & ICLR'20                     &\checkmark& &                      & GENESIS \cite{Engelcke2020} & ICLR'20   & \checkmark & & \checkmark \\ 
GNM \cite{Jiang2020}       & NeurIPS'20                  &\checkmark& &\checkmark           & SlotAtt \cite{Locatello2020} & NeurIPS'20 & & \checkmark & \\ 
SPLIT \cite{Charakorn2020} & arXiv'20                    &\checkmark& &                      & GENESIS-V2 \cite{Engelcke2021} & NeurIPS'21 & \checkmark & & \checkmark \\
OCIC \cite{Anciukevicius2020} & arXiv'20                 &  \checkmark & & \checkmark           & R-MONet \cite{ShengxinQian2021} & arXiv'21 & \checkmark & & \checkmark \\
GSGN \ \cite{Deng2021}     & ICLR'21                     &  \checkmark & & \checkmark           & CAE \cite{Lowe2022} & arXiv'22 & & & \checkmark \\ \hline
\end{tabular}
\label{tab:existing_models}
\vspace{-0.1cm}
\end{table}

\subsection{Considered Datasets}
We consider two groups of datasets for extensive benchmarking and analysis: 1) three commonly-used synthetic datasets: dSprites \cite{Matthey2017}, Tetris \cite{Kabra2019} and CLEVR \cite{Johnson2017}, 2) three real-world datasets: YCB \cite{Calli2017}, ScanNet \cite{Dai2017}, and COCO \cite{Lin2014}, representing the small-scale, indoor- and outdoor-level real scenes respectively. Naturally, objects and scenes in different datasets tend to have very different types of biases. For example, the objects in dSprites tend to have the single-color bias, while COCO does not. Generally, the object-level biases can be divided as: 1) appearance biases including different textures and lighting effects, and 2) geometry biases including the object shape and occlusions. Similarly, the scene-level biases include: 1) appearance biases such as the color similarity between all objects, and 2) geometry biases such as the diversity of all object shapes. In fact, our complexity factors introduced in Section \ref{sec:complexity_factors} are designed to well capture these biases.
Table \ref{tab:dataset_biases} qualitatively summarizes the biases of selected datasets. We may hypothesize that the large gaps of biases between synthetic and real-world datasets would have a huge impact on the effectiveness of existing models. 

\begin{table}[tb]
\centering
\setlength{\abovecaptionskip}{1 pt}
\setlength{\belowcaptionskip}{2 pt}
\footnotesize
\caption{The object- and scene-level biases in appearance and geometry of the considered datasets.}
\tabcolsep= 0.14cm 
\begin{tabular}{lr|lll|lll} \hline
&     &\multicolumn{3}{c|}{Synthetic Datasets} &  \multicolumn{3}{c}{Real-world Datasets}   \\ 
&     &dSprites \cite{Matthey2017} & Tetris \cite{Kabra2019}  & CLEVR \cite{Johnson2017} & YCB \cite{Calli2017} & ScanNet \cite{Dai2017} & COCO \cite{Lin2014} \\ \hline
\multicolumn{8}{c}{\textit{Object-level Biases}} \\ \hlineB{3}  
\multirow{2}{*}{Appearance} & Texture: & simple   & simple  & simple  & diverse   & simple  & diverse  \\  
                            & Lighting: & no       & no      & synthetic      & real   & real  & real  \\ \hdashline
\multirow{2}{*}{Geometry}   & Shape:    & simple     & simple    & simple  & simple   & diverse  & diverse  \\ 
                            & Occlusion: & minor & no    & minor & severe & severe  & severe \\  \hline
\multicolumn{8}{c}{\textit{Scene-level Biases}} \\ \hlineB{3}  
Appearance                  & Similarity: & low   & low  & high  & high   & high  & high  \\  \hdashline
Geometry                    & Diversity:  & low   & low    & low  & high   & high  & high  \\  \hline
\end{tabular}
\label{tab:dataset_biases}
\vspace{-0.3cm}
\end{table}

To guarantee the fairness and consistency of all experiments, we carefully prepare all six datasets using the following same protocols. Preparation details for each dataset are provided in appendix.
\begin{itemize}[leftmargin=*]
\setlength{\itemsep}{1pt}
\setlength{\parsep}{1pt}
\setlength{\parskip}{1pt}
    \item All images are rerendered or cropped with the same resolution of $128\times 128$. 
    \item Each image has about 2 to 6 solid objects with a blank background. 
    \item Each dataset has about 10000 images for training, 2000 images for testing.
\end{itemize}

\subsection{Considered Metrics}
Having the six representative datasets and four existing unsupervised methods at hand, we choose the following metrics to evaluate the object segmentation performance: 1) AP score which is widely used for object detection and segmentation \cite{Everingham2015}, 2) PQ score which is used to measure non-overlap panoptic segmentation \cite{Kirillov2019}, and 3) Precision and Recall scores. A predicted mask is considered correct if its IoU against a ground truth mask is above 0.5. All objects are treated as a single class. The blank background is not taken into account for fair comparison. To compute AP, we simply treat the mean value of the soft object mask as the object confidence score. Note that, the alternative metrics ARI \cite{Rand1971} and segmentation covering (SC) \cite{Arbelaez2011} are not considered as they can be easily saturated. 

\section{Key Experimental Results}\label{sec:experimental_res}
\subsection{Can current unsupervised models succeed on real-world datasets?}\label{sec:exp_main}
\begin{figure}[t]
    \setlength{\abovecaptionskip}{ 1 pt}
    \setlength{\belowcaptionskip}{ 2 pt}
    \centering
       \includegraphics[width=1\linewidth]{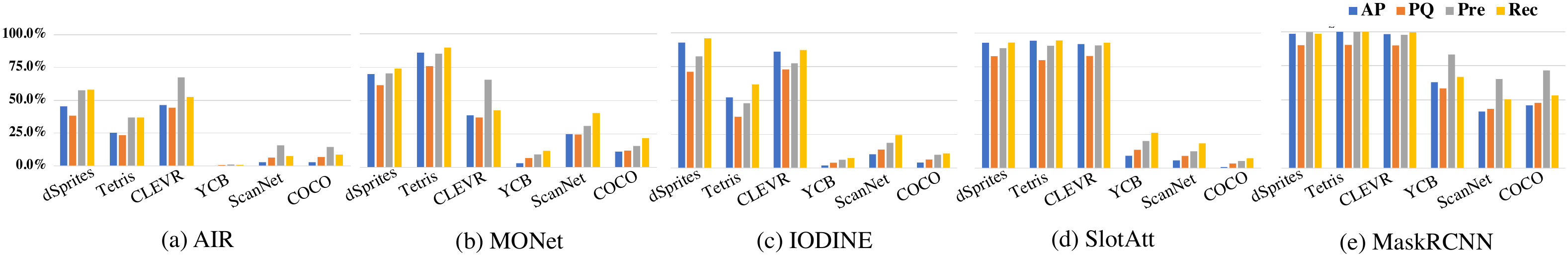}
    \caption{{Quantitative results of object segmentation from the five methods on six datasets.}}
    \label{fig:res_main}
    \vspace{-0.4cm}
\end{figure}

First of all, we evaluate all baselines on our six datasets separately. In particular, we train each model from scratch on each dataset separately. For fair evaluation, we carefully tune the hyperparameters of each model on every dataset and fully optimize the networks until convergence. {Figure {\ref{fig:res_main}}} compares the quantitative results. It can be seen that all methods demonstrate satisfactory segmentation results on synthetic datasets, especially the recent strong baselines IODINE and SlotAtt. However, not surprisingly, all {unsupervised} methods fail catastrophically on the three real-world datasets. 

\textbf{Preliminary Diagnosis:} In order to diagnose the colossal failure, we hypothesize that it is because of the huge gaps in objectness biases between two types of datasets. In this regard, we quantitatively compute the distributions of our four complexity factors on the six datasets. In particular, the two object-level factors, \ie{}, Object Color Gradient and Object Shape Concavity, are computed for each object of the six training splits. The two scene-level factors, \ie{}, Inter-object Color Similarity and Inter-object Shape Variation, are computed for each image of the six training splits. 

\begin{figure}[hb]
\vspace{-0.2cm}
    \setlength{\abovecaptionskip}{ 1 pt}
    \setlength{\belowcaptionskip}{ -6 pt}
    \centering
       \includegraphics[width=1\linewidth]{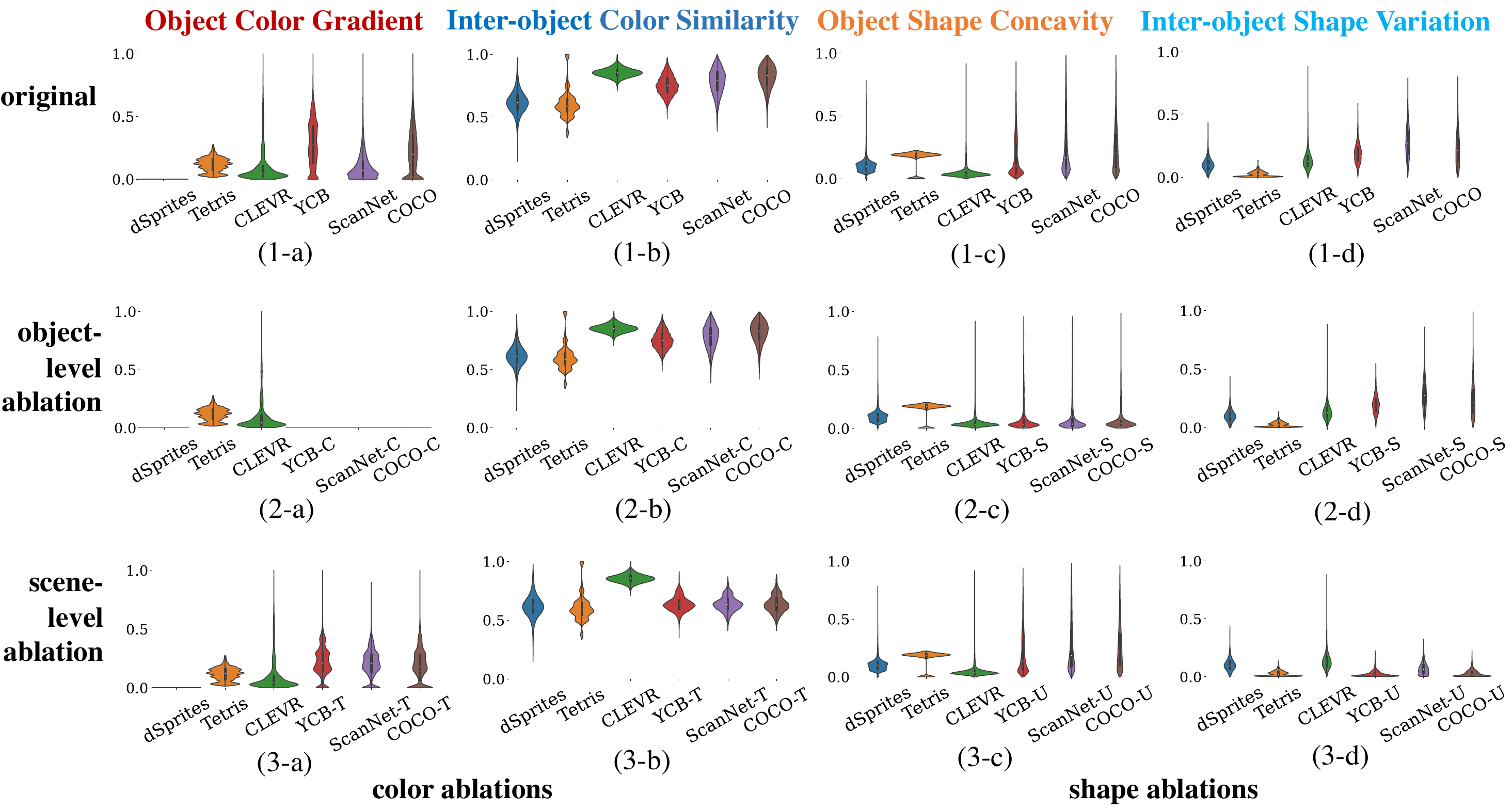}
    \caption{{Distributions of the four complexity factors. The \textit{top row} shows the distributions of original  synthetic/real-world datasets in Sec {\ref{sec:exp_main}}. The \textit{2nd row} shows the distributions of object-level ablated datasets in Sec {\ref{sec:exp_object_factors}}. The \textit{3rd row} shows distributions of scene-level ablated datasets in Sec {\ref{sec:exp_scene_factors}}.}}
    \label{fig:factors_distributions}
    \vspace{-0.1cm}
\end{figure}

From the distributions shown in {the top row of} Figure \ref{fig:factors_distributions}, we can see that, 1) for the two object-level factors {(Subfigs 1-a and 1-c)}, the three synthetic datasets tend to have extremely lower scores than the real-world datasets, which means that the synthetic objects are more likely have uniform colors and convex shapes; 2) for the two scene-level factors {(Subfigs 1-b and 1-d)}, the images in synthetic datasets tend to include less similar objects in terms of color, which means that multiple objects in real-world scenes are less distinctive in appearance. In addition, the multiple objects in synthetic scenes tend to have similar sizes, whereas real-world scenes usually have diverse object scales in single images. To validate whether these distribution biases are the true reasons incurring the failure, we conduct extensive ablative experiments in Sections \ref{sec:exp_object_factors}, \ref{sec:exp_scene_factors}
and \ref{sec:exp_joint_factors}. 

\subsection{How do object-level factors affect current models?}\label{sec:exp_object_factors}
In this section, we aim to verify to what extent the distributions of object-level factors affect the segmentation performance. In particular, we conduct the following three ablative experiments. 

\setlength{\columnsep}{10pt}
\begin{wrapfigure}{R}{0.33\textwidth}
\setlength{\abovecaptionskip}{ 1 pt}
\setlength{\belowcaptionskip}{ -16 pt}
\centering
\raisebox{0pt}[\dimexpr\height-1.\baselineskip\relax]{
   \centering
   \includegraphics[width=1\linewidth]{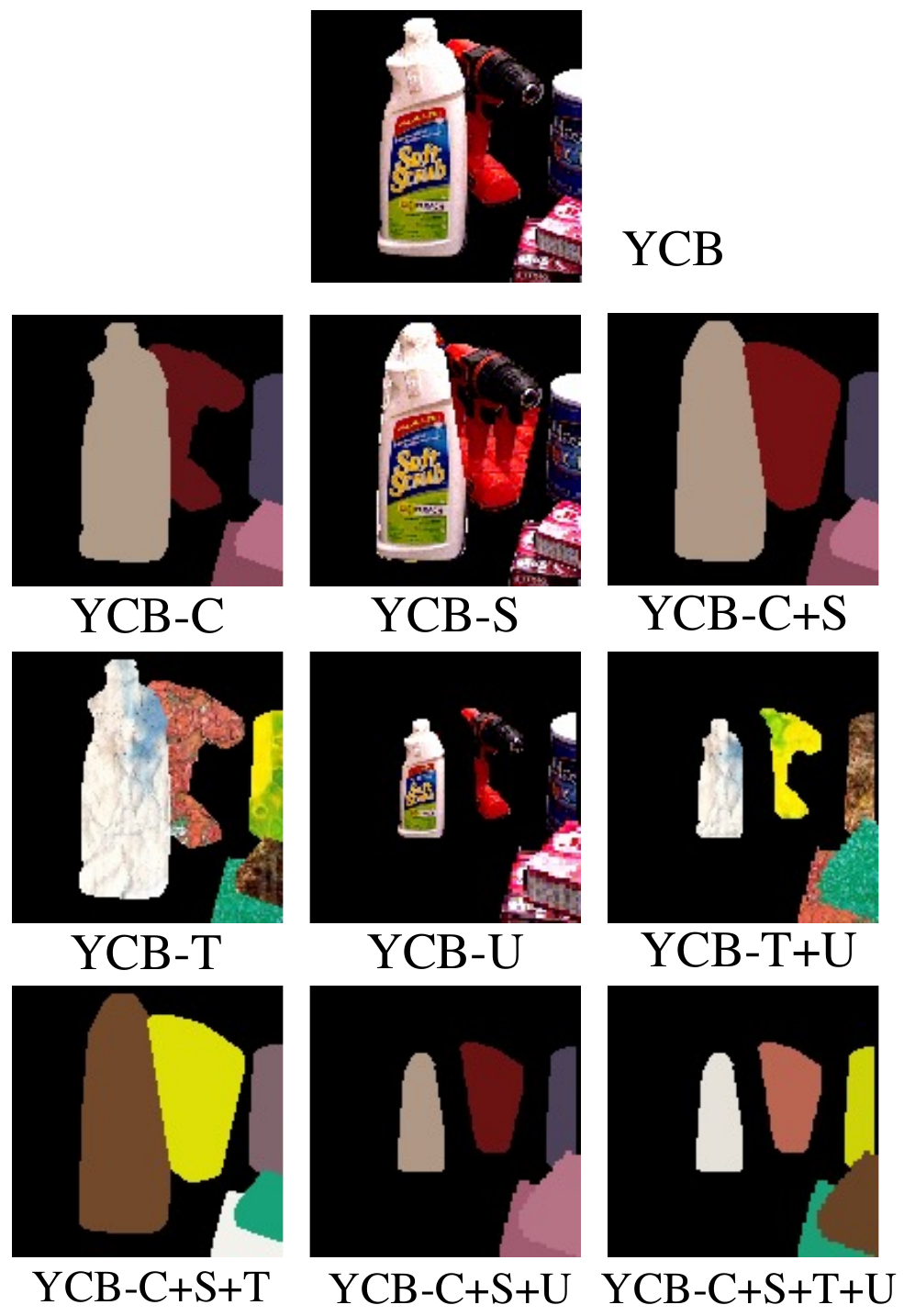}
}

\caption{{Sample images with different ablations. The \textit{top row} shows an original image from YCB dataset. The \textit{2nd row} shows examples from three ablated datasets in Sec \ref{sec:exp_object_factors}. The \textit{3rd row} presents examples from three ablated datasets in Sec \ref{sec:exp_scene_factors}. The last row shows examples from three ablated datasets in Sec \ref{sec:exp_joint_factors}}} 
\label{fig:ablations_illustration}
\vspace{-0.3cm}
\end{wrapfigure}

\begin{itemize}[leftmargin=*]
\setlength{\itemsep}{1pt}
\setlength{\parsep}{1pt}
\setlength{\parskip}{1pt}
    \item \textit{Ablation of Object Color Gradient}: For each object of the three real-world datasets, we only replace all pixel colors by its average color $rgb$ value within each object mask, without touching the object shapes. In this way, the \textbf{c}olor gradients of each object are totally erased, thus removing the potential impact of Object Color Gradient. The three ablated datasets are: YCB-C / ScanNet-C / COCO-C. 
    \item \textit{Ablation of Object Shape Concavity}: For each object in the real-world datasets, we find the smallest convex hull \cite{Eddins2011} for its object mask and then fill the empty pixels by shifting the original object pixels. Basically, this ablation aims to only reduce the irregularity of object {\textbf{s}}hapes, yet retaining the distributions of color gradients. The ablated datasets are: YCB-S / ScanNet-S / COCO-S.  
    \item \textit{Ablation of both Object Color Gradient and Shape Concavity}: We simply combine the above two ablation for each real-world object, getting datasets: YCB-C+S / ScanNet-C+S / COCO-C+S.
\end{itemize}

For illustration, the 2nd row of Figure \ref{fig:ablations_illustration}  shows example images of three ablated datasets: YCB-C / YCB-S / YCB-C+S.

In the 2nd row of Figure \ref{fig:factors_distributions} (Subfigs 2-a/2-b), we calculate new distributions of both Object Color Gradient and Inter-object Color Similarity on the datasets YCB-C / ScanNet-C / COCO-C. We can see that the object-level gradients become all zeros in Subfig 2-a, even simpler than  three synthetic datasets. Yet, the distributions of Inter-object Color Similarity are almost the same as original YCB / ScanNet / COCO, \ie{}, Subfig 2-b is similar to 1-b. 

Similarly, the 2nd row of of Figure {\ref{fig:factors_distributions}} (Subfigs 2-c/2-d) shows that the distributions of Object Shape Concavity of ablated datasets now become similar to the synthetic datasets, while the distributions of Inter-object Shape Variation keep the same, \ie{}, Subfig 2-d is similar to Subfig 1-d. Note that, for the ablation of both Object Color Gradient and Shape Concavity, the distributions will be the same as shown in the 2nd row of Figure \ref{fig:factors_distributions}. Having the three groups of object-level ablated real-world datasets, we then evaluate segmentation performance of the baselines separately.

\begin{figure}[t]
    \setlength{\abovecaptionskip}{ 4 pt}
    \setlength{\belowcaptionskip}{ -8 pt}
    \centering
       \includegraphics[width=0.95\linewidth]{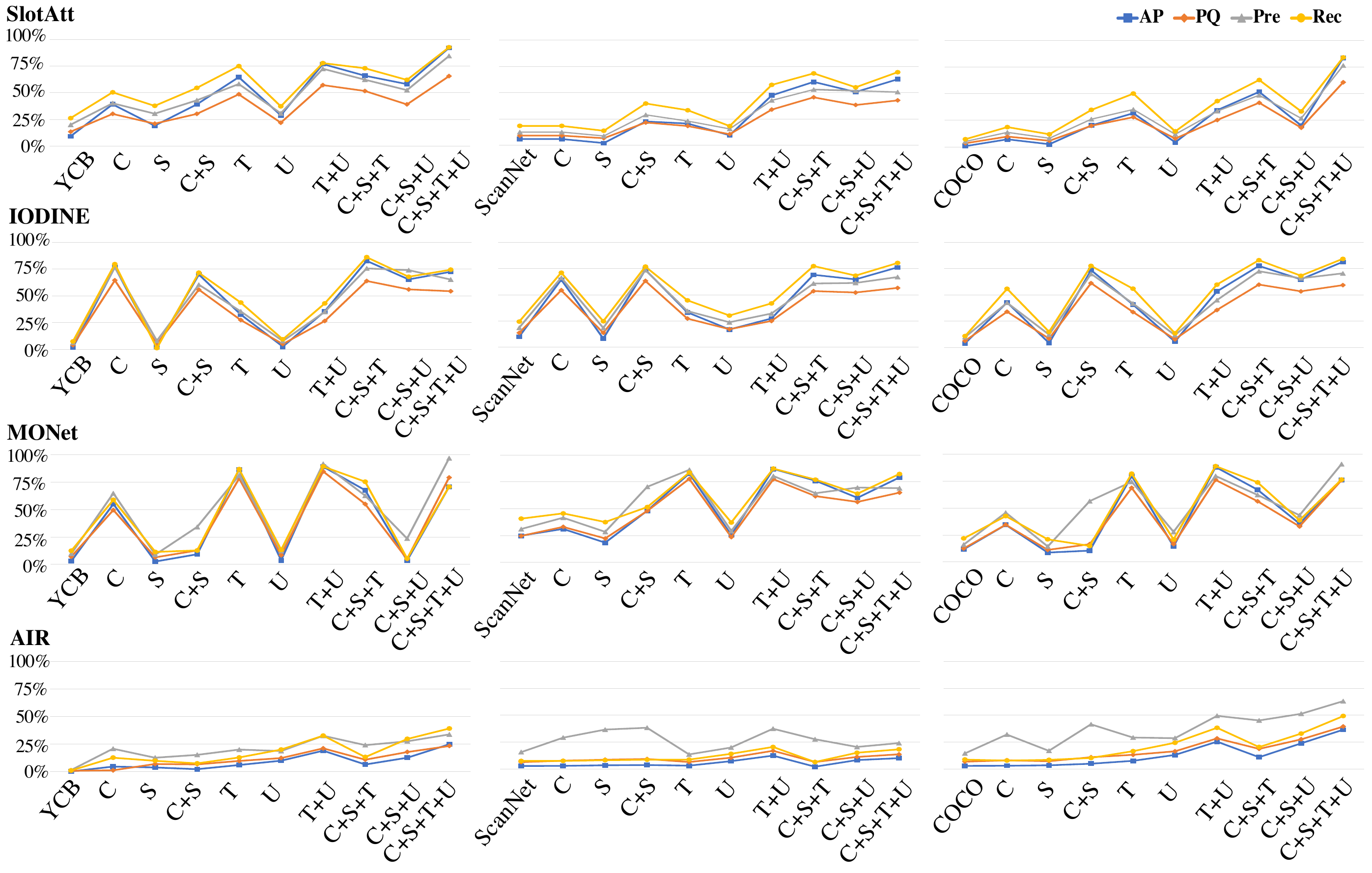}
    \caption{{Quantitative results of baselines on the three real-world datasets and their variants. The letters C/S/C+S represent the three ablated datasets in Sec {\ref{sec:exp_object_factors}}; T/U/T+U represent the three ablated datasets in Sec {\ref{sec:exp_scene_factors}}; C+S+T/C+S+U/C+S+T+U represent the three ablated datasets in Sec {\ref{sec:exp_joint_factors}}.}}
    \label{fig:res_ablations}
    \vspace{-0.2cm}
\end{figure}

\textbf{Brief Analysis:} As shown in Figures \ref{fig:res_ablations} \& \ref{fig:qualitative_object_scene_level_ablation}, we can see that: 1) Once the pixels of real-world objects are replaced by its mean color, \ie{}, without any color gradients, the object segmentation performance has been significantly improved for almost all methods. 2) Reducing the irregularity of real-world objects can also improve the object segmentation, although not significantly. 3) Overall, these results show that existing methods are more likely to learn the objectness represented by uniform colors and/or regular objects. However, comparing with Figure \ref{fig:res_main}, the results of current ablated datasets in Figure \ref{fig:res_ablations} still lag behind the synthetic datasets. This means that there must be some other factors that also potentially affect the object segmentation of existing models. More results are in appendix.

\subsection{How do scene-level factors affect current models?}\label{sec:exp_scene_factors}
In this section, we turn to investigate to what extent the distributions of scene-level factors affect the segmentation performance. Particularly, we conduct the following three ablative experiments.

\begin{wrapfigure}{R}{0.38\textwidth}
\setlength{\abovecaptionskip}{ -1 pt}
\setlength{\belowcaptionskip}{ -10 pt}
\centering
\raisebox{0pt}[\dimexpr\height-1.\baselineskip\relax]{
   \centering
   \includegraphics[width=1\linewidth]{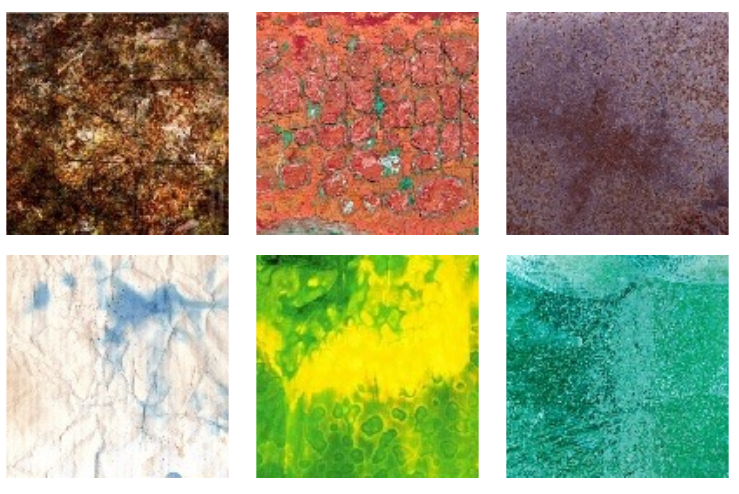}
}
\caption{Six selected texture images from DTD \cite{Cimpoi2014}.} 
\label{fig:dtd_texture_0}
\vspace{-0.3cm}
\end{wrapfigure}

\begin{itemize}[leftmargin=*]
\setlength{\itemsep}{1pt}
\setlength{\parsep}{1pt}
\setlength{\parskip}{1pt}
    \item \textit{Ablation of Inter-object Color Similarity}: In each image of the three real-world datasets, we replace all object textures by a set of new distinctive textures from the existing DTD database \cite{Cimpoi2014}, as shown in Figure \ref{fig:dtd_texture_0}. In this way, the multiple objects look more distinctive in appearance, while the per-object texture gradients are roughly preserved. The ablated datasets are denoted: YCB-T / ScanNet-T / COCO-T.   
    \item \textit{Ablation of Inter-object Shape Variation}: In each image of the real-world datasets, we normalize the scales of  multiple objects by shrinking or expanding the diagonal length of their bounding boxes, such that the new object sizes tend to be uniform. For each object, its shape and texture are linearly scaled up or down. Basically, this aims to remove the diversity of object sizes within single images. The ablated datasets are denoted as: YCB-U / ScanNet-U / COCO-U.   
    \item \textit{Ablation of both Inter-object Color Similarity and Shape Variation}: We simply combine the above two ablation strategies for each real-world image. Ablation details are in appendix. The ablated datasets are denoted as: YCB-T+U / ScanNet-T+U / COCO-T+U.   
\end{itemize}

\setlength{\columnsep}{10pt}
For illustration, the 3rd row of Figure \ref{fig:ablations_illustration} shows example images of three ablated datasets: YCB-T / YCB-U / YCB-T+U.

In the 3rd row of Figure \ref{fig:factors_distributions} (Subfigs 3-a/3-b), we calculate new distributions of both Object Color Gradient and Inter-object Color Similarity on the ablated datasets YCB-T / ScanNet-T / COCO-T. We can see that the distributions of scene-level color similarity become more similar to synthetic datasets in Subfig 3-b, while the distributions of object color gradients are still similar as original real-world datasets YCB / ScanNet / COCO, \ie{}, Subfig 3-a is similar to 1-a.

Similarly, the 3rd row of Figure {\ref{fig:factors_distributions}} shows that the distributions of Inter-object Shape Variation of ablated datasets now become similar to the synthetic datasets in Subfig 3-d, whereas the distributions of Object Shape Concavity are still the same as original real-world datasets YCB / ScanNet / COCO, \ie{}, Subfig 3-c is similar to Subfig 1-c. Note that, for the ablation of both Inter-object Color Similarity and Shape Variation, the distributions will be the same as the 3rd row of Figure \ref{fig:factors_distributions}. Having these three groups of scene-level ablated real-world datasets, we then separately evaluate each baseline.

\begin{figure}[ht]
    \setlength{\abovecaptionskip}{ 4 pt}
    \setlength{\belowcaptionskip}{ 6 pt}
    \centering
      \includegraphics[width=1\linewidth]{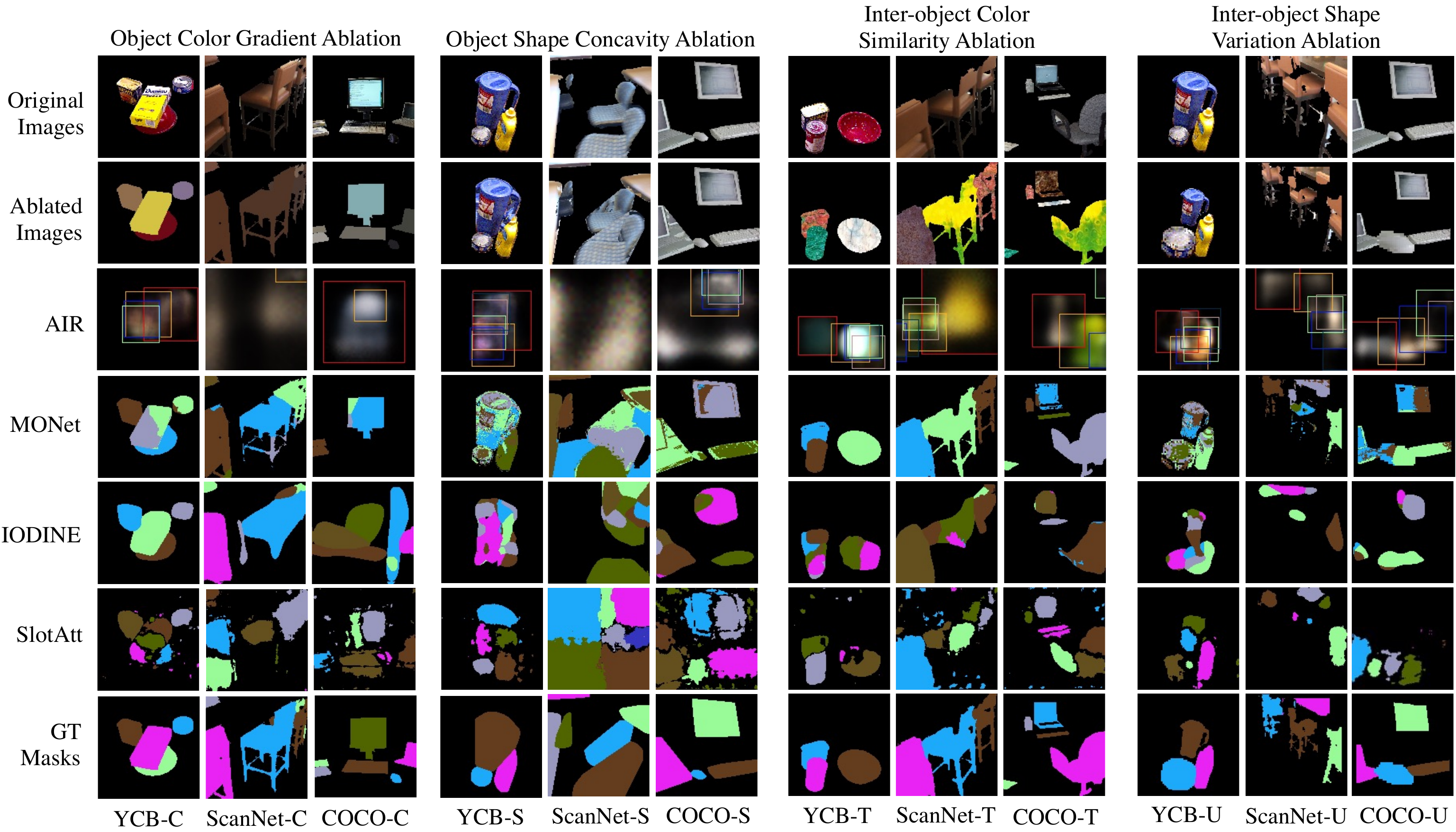}
    \caption{Qualitative results of four representative methods on multiple ablation datasets in Sec \ref{sec:exp_object_factors} and Sec \ref{sec:exp_scene_factors}.}
    \label{fig:qualitative_object_scene_level_ablation}
    \vspace{-0.2cm}
\end{figure}

\textbf{Brief Analysis:} As shown in Figures \ref{fig:res_ablations} \& \ref{fig:qualitative_object_scene_level_ablation}, we can see that: 
1) Once the textures of real-world images are replaced by more distinctive textures, \ie{}, with lower similarity between object appearance, the segmentation performance has been surprisingly boosted remarkably for almost all methods. 2) Normalizing object sizes over images can also reasonably improve the segmentation performance. 3) Overall, these results clearly show that existing unsupervised models significantly favor objectness with distinctive appearance in single images. However, compared with Figure \ref{fig:res_main}, the results on current scene-level ablated datasets are still inferior to synthetic datasets, meaning that the scene-level factors alone are not enough to explain the performance gap. More results are in appendix.

\subsection{How do object- and scene-level factors jointly affect current models?} \label{sec:exp_joint_factors}
In this section, we aim to study how the object- and scene-level factors jointly affect the segmentation performance. In particular, we conduct the following three ablative experiments.

\vspace{-0.2cm}
\begin{itemize}[leftmargin=*]
\setlength{\itemsep}{1pt}
\setlength{\parsep}{1pt}
\setlength{\parskip}{1pt}
    \item \textit{Ablation of Object Color Gradient, Object Shape Concavity and Inter-object Color Similarity}: In each image of the three real-world datasets, we replace the object color by averaging all pixels of the distinctive texture, and also replace the object shape with a simple convex hull. The ablated datasets are denoted: {YCB-C+S+T / ScanNet-C+S+T / COCO-C+S+T}.   
    \item \textit{Ablation of Object Color Gradient, Object Shape Concavity and Inter-object Shape Variation}: In each image of the three real-world datasets, we replace the object color by averaging its own texture, and modify the object shape as convex hull following by size normalization. The ablated datasets are denoted as: {YCB-C+S+U / ScanNet-C+S+U / COCO-C+S+U}.   
    \item \textit{Ablation of all four factors}: We aggressively combine all four ablation strategies and we get datasets: {YCB-C+S+T+U / ScanNet-C+S+T+U / COCO-C+S+T+U}.  
\end{itemize}\vspace{-0.2cm}
For illustration, the 4th row of Figure \ref{fig:ablations_illustration} shows example images of three ablated datasets: YCB-C+S+T / YCB-C+S+U / YCB-C+S+T+U.

Since these ablations are conducted independently, the new distributions of four complexity factors on current jointly ablated datasets are exactly the same as the second and third rows of Figure \ref{fig:factors_distributions}.

\textbf{Brief Analysis:} As shown in Figure \ref{fig:res_ablations}, we can see that: 1) Combining the two object-level factors and either of scene-level factors to ablate the real-world datasets, the segmentation performance can be improved as we expect, especially for IODINE and SlotAtt. 2) If the challenging real-world objects and images are ablated in both object- and scene-level, the segmentation performance of all unsupervised models achieves the same level with three synthetic datasets as shown in Figure \ref{fig:res_main}. 3) Overall, these three groups of experiments demonstrate that the colossal failure of unsupervised models on real-world images involves both object- and scene-level dataset biases. More experiments and results are in appendix.

\begin{figure}[ht]
    \setlength{\abovecaptionskip}{ 2 pt}
    \setlength{\belowcaptionskip}{ 2 pt}
    \centering
      \includegraphics[width=0.9\linewidth]{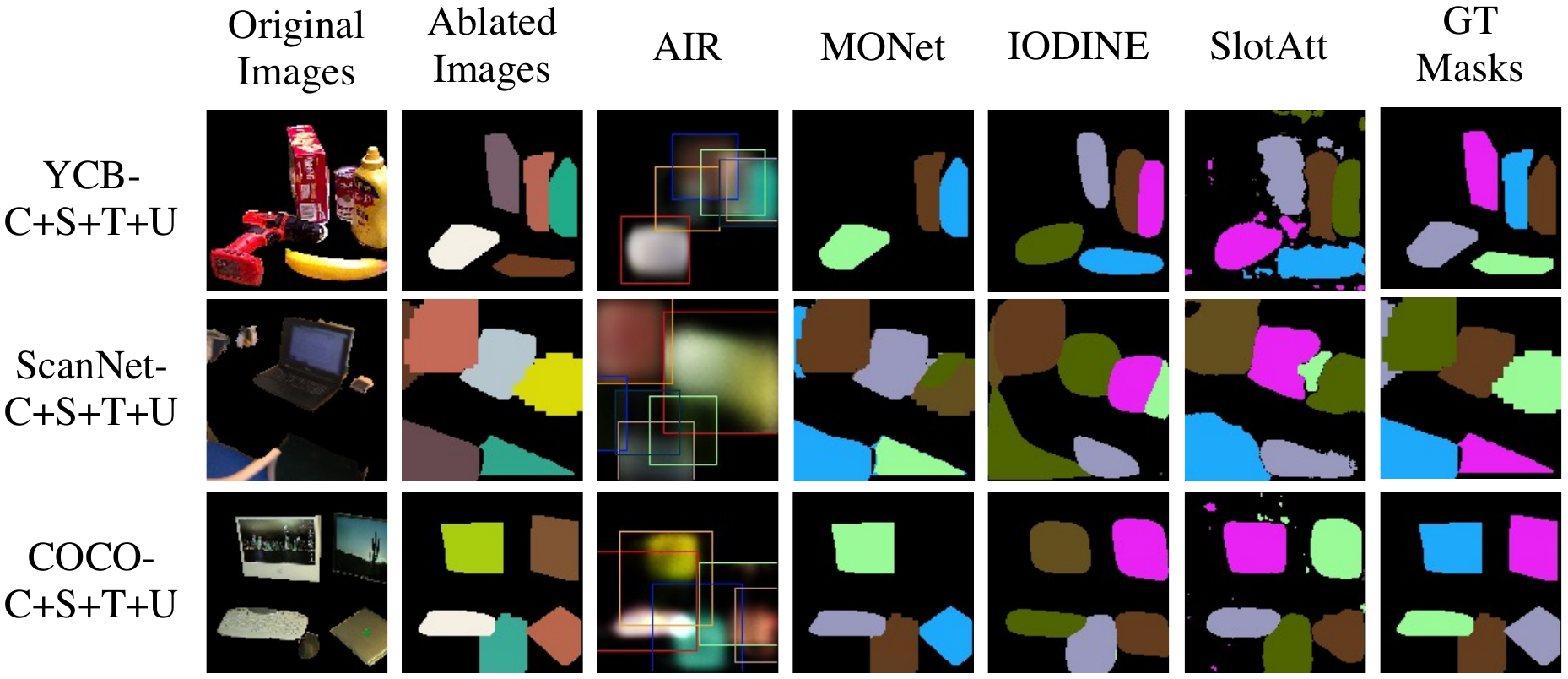}
    \caption{Qualitative results on fully ablated real-world datasets (YCB-C+S+T+U / ScanNet-C+S+T+U / COCO-C+S+T+U).}
    \label{fig:qualitative_full_ablation}
    \vspace{-0.2cm}
\end{figure}

\subsection{Why do current unsupervised models fail on real-world datasets?}\label{sec:exp_why_fail}
As demonstrated in Sections \ref{sec:exp_object_factors}/\ref{sec:exp_scene_factors}/\ref{sec:exp_joint_factors}, once the complexity factors are removed from the challenging real-world datasets, existing unsupervised models can perform as excellent as on the synthetic datasets, as qualitatively illustrated in Figure {\ref{fig:qualitative_full_ablation}}. From this,  we can safely conclude that the inductive biases designed in existing unsupervised models are far from able to match with and fully capture the true and complex objectness biases exhibited in real-world images. Nevertheless, from Figure {\ref{fig:res_ablations}}, each baseline tends to favor different objectness biases. In particular,  

\begin{itemize}[leftmargin=*]
\setlength{\itemsep}{2pt}
\setlength{\parsep}{1pt}
\setlength{\parskip}{1pt}
    \item AIR \cite{Eslami2016}: 
    {As a factor based model, AIR has a strong spatial-locality bias. Despite its poor segmentation performance across all datasets, there is a notable improvement when inter-object shape variation is ablated from real-world datasets (U / T+U / C+S+U / C+S+T+U). More convincingly, even when all other three factors are ablated (C+S+T), it can be hardly improved. These observations show that object shape variation is a significant factor for AIR.}
    \item MONet \cite{Burgess2019}:   
    {MONet is more sensitive to color-related factors than shape-related factors. The ablations of object color gradient and inter-object color similarity significantly improve its performance, while ablations of object shape concavity and inter-object shape variation make little differences. For the two color-related factors, the scene-level one is more important than the object-level factor. From this, we can see that MONet has a strong dependency on color. Similar colors tend to be grouped together while different colors are separated apart. Furthermore, the ablation on object color gradient alleviates over-segmentation whereas the ablation on inter-object color similarity alleviates under-segmentation. We conjecture that under-segmentation can be a more severe issue for MONet on real-world datasets, leading to a larger sensitivity on the scene-level color factor.}
    \item IODINE \cite{Greff2019}:
    {IODINE also has a heavy dependency on both object- and scene-level color-related factors. However, different from MONet, the ablation on object color gradient brings better performance than inter-object color similarity. We speculate it is because the regularization on shape latent alleviates under-segmentation by biasing towards more regular shapes. In this way, over-segmentation is the key issue, making the object color gradient a dominant factor.} 
    \item SlotAtt \cite{Locatello2020}:
    {The ablations on all four factors increase the performance of SlotAtt at different levels, among which object- and scene-level color-related factors are more significant. We conjecture that it is because the feature embeddings used by Slot Attention module are learnt from both pixel colors and coordinates, which contributes to its sensitivity to both shape and color factors.}
\end{itemize}

\section{Conclusions}
We systematically show that existing unsupervised methods are practically impossible to segment generic objects from single real-world images, and investigate the underlying factors that incur the catastrophic failure. 
With the aid of our carefully designed four object- and scene-level complexity factors, we conduct extensive experiments on multiple groups of ablated real-world objects and images, and safely conclude that the distributions of both object- and scene-level biases in appearance and geometry of real-world datasets are particularly diverse and indiscriminative, such that current unsupervised models cannot segment real objects. Based on this finding, we suggest two main directions for future study: 1) To exploit more discriminative objectness biases such as object motions which expressively describe the ownership of visual pixels as recently explored in \cite{Tangemann2021,Chen2022,Bear2020} for 2D images and in \cite{Song2022} for 3D point clouds. 2) To leverage pretrained features from single-object-dominant datasets which explicitly regard each image as an object as recently studied in \cite{Caron2021,Henaff2022}, although such settings are no longer purely unsupervised.    

\textbf{Broader Impact:} One limitation of this work is the lack of study about foreground-background segmentation, but our work can still have a positive impact for researchers to design more practically useful models. A potential negative impact could be using our findings to attack existing models.  

\textbf{Acknowledgements:} This work was partially supported by Shenzhen Science and Technology Innovation Commission (JCYJ20210324120603011).

\clearpage
{
\vspace*{-15mm}
\small
\bibliographystyle{abbrv}
\bibliography{references}
}

\clearpage


\begin{enumerate}

\item For all authors...
\begin{enumerate}
  \item Do the main claims made in the abstract and introduction accurately reflect the paper's contributions and scope?
    \answerYes{}
  \item Did you describe the limitations of your work?
    \answerYes{See Section 5.}
  \item Did you discuss any potential negative societal impacts of your work?
    \answerYes{See Section 5.}
  \item Have you read the ethics review guidelines and ensured that your paper conforms to them?
    \answerYes{}
\end{enumerate}

\item If you are including theoretical results...
\begin{enumerate}
  \item Did you state the full set of assumptions of all theoretical results?
    \answerNA{}
        \item Did you include complete proofs of all theoretical results?
    \answerNA{}
\end{enumerate}

\item If you ran experiments...
\begin{enumerate}
  \item Did you include the code, data, and instructions needed to reproduce the main experimental results (either in the supplemental material or as a URL)?
    \answerYes{See \textbf{Contributions and finding} in Section 1.}
  \item Did you specify all the training details (e.g., data splits, hyperparameters, how they were chosen)?
    \answerYes{See \textbf{Implementation Details} in Appendix.}
    \item Did you report error bars (e.g., with respect to the random seed after running experiments multiple times)?
    \answerYes{In the tables for qunatitative results, we report standard deviation of performances over three runs.}
    \item Did you include the total amount of compute and the type of resources used (e.g., type of GPUs, internal cluster, or cloud provider)?
    \answerYes{See \textbf{Implementation Details} in Appendix.}
\end{enumerate}

\item If you are using existing assets (e.g., code, data, models) or curating/releasing new assets...
\begin{enumerate}
  \item If your work uses existing assets, did you cite the creators?
    \answerYes{}
  \item Did you mention the license of the assets?
    \answerNA{}
  \item Did you include any new assets either in the supplemental material or as a URL?
    \answerYes{See \textbf{Contributions and finding} in Section 1.}
  \item Did you discuss whether and how consent was obtained from people whose data you're using/curating?
    \answerNA{}
  \item Did you discuss whether the data you are using/curating contains personally identifiable information or offensive content?
    \answerNA{}
\end{enumerate}

\item If you used crowdsourcing or conducted research with human subjects...
\begin{enumerate}
  \item Did you include the full text of instructions given to participants and screenshots, if applicable?
    \answerNA{}
  \item Did you describe any potential participant risks, with links to Institutional Review Board (IRB) approvals, if applicable?
    \answerNA{}
  \item Did you include the estimated hourly wage paid to participants and the total amount spent on participant compensation?
    \answerNA{}
\end{enumerate}

\end{enumerate}

\clearpage
\appendix
\section{Appendix}

\subsection{Details of the Four Complexity Factors}

We have introduced four complexity factors in both object- and scene-level appearance and geometry. Details of these factors are as follows.

\begin{itemize}[leftmargin=*]
\setlength{\itemsep}{1pt}
\setlength{\parsep}{1pt}
\setlength{\parskip}{1pt}
\item \textbf{Object Color Gradient:}
    As shown in Figure \ref{fig:object_color_gradient}, given an RGB image of an object, we first convert it to grayscale by applying $Y = 0.299R + 0.587G + 0.114B$. Then Sobel filter \cite{Sobel1973} with kernel size $3 \times 3$ is applied horizontally and vertically to compute the image gradient for each pixel. Since this factor should only be related to the appearance inside an object regardless of the background, we remove the gradients at the object boundary. The object boundary is computed from its mask, following the practice of \cite{Cheng2021a}. The factor score is calculated as the average gradient of all object pixels that are not on the boundary. This value ranges between 0 and 255, which is then divided by 255 so as to be normalized to the range of $[0,1]$.
    \begin{figure}[ht]
    \setlength{\abovecaptionskip}{ -0 pt}
    \setlength{\belowcaptionskip}{ -4 pt}
    \centering
       \includegraphics[width=0.8\linewidth]{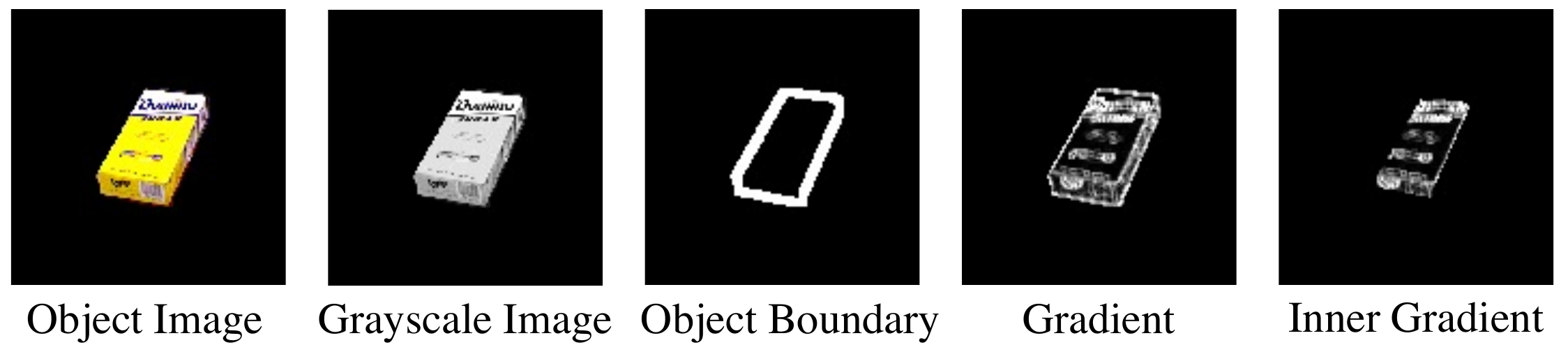}
    \caption{The calculation of Object Color Gradient.}
    \label{fig:object_color_gradient}
    \end{figure}
    
    \item \textbf{Object Shape Concavity:}
    As shown in Figure \ref{fig:object_shape_concavity}, given the binary mask of an object ($\boldsymbol{M}_{obj}\in \mathbb{R}^{H\times W}$), we first calculate its smallest surrounding convex polygon mask ($\boldsymbol{M}_{cvx} \in \mathbb{R}^{H\times W}$) using the existing algorithm \cite{Eddins2011}. Basically, this polygon mask is the smallest region that can cover any lines between two points on the original object mask. The object shape concavity factor is calculated as $1-\sum\boldsymbol{M}_{obj} / \sum\boldsymbol{M}_{cvx}$. This factor naturally takes a value between 0 and 1. No further normalization is needed.
    \begin{figure}[ht]
    \setlength{\abovecaptionskip}{ -0 pt}
    \setlength{\belowcaptionskip}{ -4 pt}
    \centering
       \includegraphics[width=0.5\linewidth]{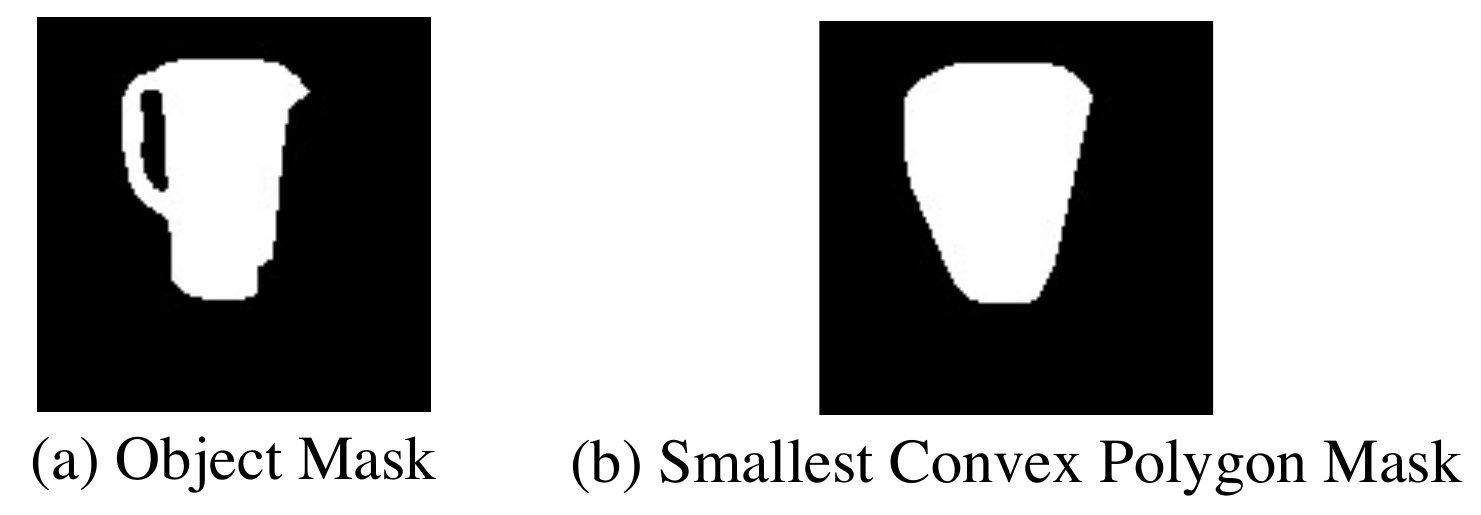}
    \caption{An illustration of the Object Shape Concavity.}
    \label{fig:object_shape_concavity}
    \vspace{-0.25cm}
    \end{figure}
    
    \item \textbf{Inter-object Color Similarity:}
    As shown in Figure \ref{fig:inter_object_color_similarity}, we first calculate the average RGB color of each object. Each color corresponds to a point in the RGB space. We then calculate the Euclidean distance between each pair of object colors. The average value of all pairwise distances is divided by $255 \times \sqrt{3}$, which is the largest distance between two colors in the RGB space, so as to be normalized to the range of $[0, 1]$. The final score for inter-object color similarity is calculated as $1 - normalized\;RGB\;distance$.
    \begin{figure}[ht]
    \setlength{\abovecaptionskip}{ -0 pt}
    \setlength{\belowcaptionskip}{ -4 pt}
    \centering
       \includegraphics[width=0.5\linewidth]{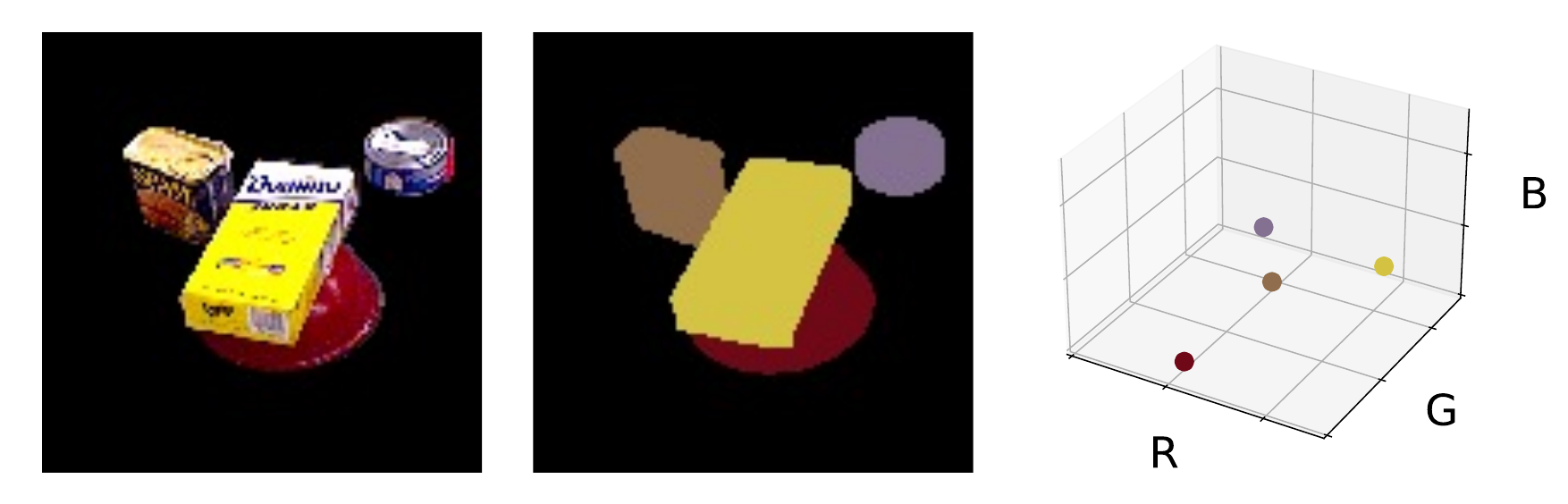}
    \caption{An illustration of the Inter-object Color Similarity.}
    \label{fig:inter_object_color_similarity}
    \end{figure}
    
    \item \textbf{Inter-object Shape Variation:} We first compute an axis-aligned bounding box for each object, and calculate its diagonal length. Then the differences of diagonal lengths between each pair of objects are calculated. The average value of such pairwise differences is the raw value of inter-object shape variation. This raw value is divided by $255 \times \sqrt{2}$, which is the largest possible difference of diagonal length, so as to be normalized to the range of $[0, 1]$.
\end{itemize}

\subsection{Other Candidates of Complexity Factors }
In addition to the primary four complexity factors, we also explore other potential complexity factors to quantitatively measure the distributions of object- and scene-level biases in appearance and geometry. Basically, we aim to consider as many aspects as possible to investigate key factors underlying the distribution gaps between synthetic and real-world datasets. 
However, we empirically find that these candidate factors do not show significant discrepancy between synthetic and real-world datasets. Details are shown below.

\subsubsection{Candidates of Object-level Complexity Factors}

\begin{itemize}[leftmargin=*]
\setlength{\itemsep}{1pt}
\setlength{\parsep}{1pt}
\setlength{\parskip}{1pt}
    \item \textbf{Object Color Count:} 
    This factor is defined as the total number of unique colors within an object mask. Basically, this is to simply measure the diversity of object color. 
    \item \textbf{Object Color Entropy:}
    Inspired by Shannon entropy \cite{shannon1948mathematical}, we calculate the entropy value at each pixel by applying a $3 \times 3$ filter on the grayscale image concerted from RGB. In particular, for each pixel, its color value becomes a discrete value in $[0, 255]$. We compute its entropy score: $H(x) = -\sum_{i=1}^n p(x_i) \log_2 p(x_i)$, where $p(x_i)$ denotes the probability of a specific color value $x_i$ within the $3\times 3$ neighbourhood. Basically, this factor aims to measure the color diversity within $3\times 3$ image patches. The higher this factor, the more frequently the object color changes in small local areas. 
    \item \textbf{Object Shape Non-rectangularity:}
    Given the binary mask of an object ($\boldsymbol{M}_{obj}\in \mathbb{R}^{H\times W}$), we first calculate its axis-aligned bounding box ($\boldsymbol{M}_{bbox} \in \mathbb{R}^{H\times W}$). Object shape non-rectangularity is calculated as $1-\sum\boldsymbol{M}_{obj} / \sum\boldsymbol{M}_{bbox}$. Similar to object shape concavity, this factor is also designed to measure the complexity of object shapes. However, this factor is more likely to be affected by the object orientation since it takes axis-aligned bounding boxes as reference.
    \item \textbf{Object Shape Incompactness:}
    There are two similar methods to quantify the compactness of object shapes. The first one is Polsby–Popper test \cite{polsby1991third}: $PP(\boldsymbol{M}_{obj}) = 4\pi A(\boldsymbol{M}_{obj}) /P(\boldsymbol{M}_{obj})^2 $. The other is Schwartzberg \cite{schwartzberg1965reapportionment} compactness score: $S(\boldsymbol{M}_{obj}) = (2 \pi \sqrt{A(\boldsymbol{M}_{obj}) / \pi}) / P(\boldsymbol{M}_{obj})$. In both formula, $P(\boldsymbol{M}_{obj})$ is the object perimeter and $A(\boldsymbol{M}_{obj})$ is the object area. For simplicity, we choose $PP(\boldsymbol{M}_{obj})$ to calculate the object shape incompactness score: $1 - PP(\boldsymbol{M}_{obj})$. 
    \item \textbf{Object Shape Discontinuity:}
    Given an object mask ($\boldsymbol{M}_{obj}\in \mathbb{R}^{H\times W}$), we first find the largest connected component ($\boldsymbol{M}_{lcc}\in \mathbb{R}^{H\times W}$) in its binary mask. The discontinuity of shape is calculated as: $1 - \sum\boldsymbol{M}_{lcc} / \sum\boldsymbol{M}_{obj}$. This factor is to evaluate how continuous an object shape is. 
    \item \textbf{Object Shape Decentralization:} Given an object mask, we first calculate its centroid (${\Bar{x}, \Bar{y} }$) by averaging all pixel coordinates in the object. Then, the second moment of this object is calculated as: $\sum_x\sum_y (x - \Bar{x})^2 (y - \Bar{y})^2$, where $(x, y)$ is the coordinates of pixels within the object. The higher this factor, the object shape is less likely to be centralized.  
\end{itemize}
\vspace{-0.2cm}

\begin{figure}[ht]
    \vspace{-0.2cm}
    \setlength{\abovecaptionskip}{ -2 pt}
    \setlength{\belowcaptionskip}{ -8 pt}
    \centering
       \includegraphics[width=1\linewidth]{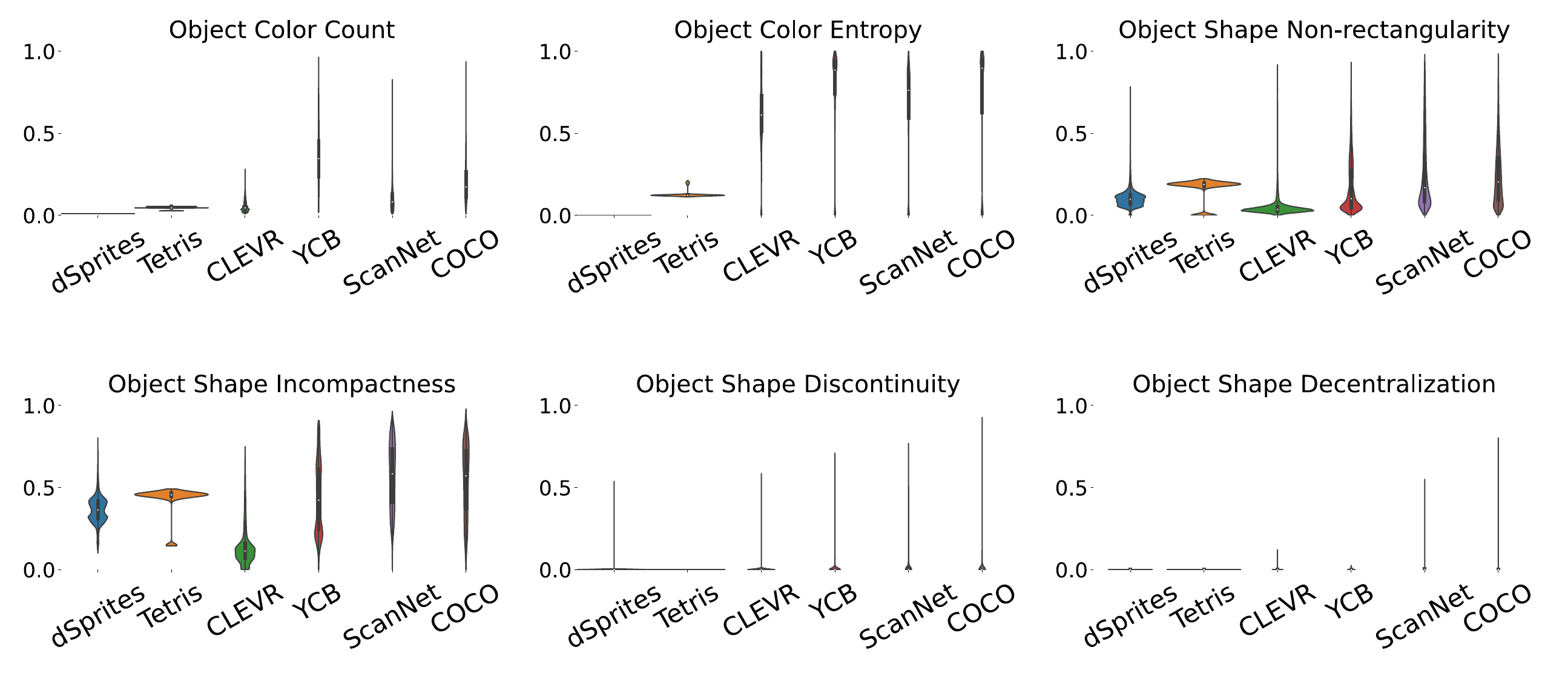}
    \caption{Distributions of Object-level Complexity Factors candidates.}
    \label{fig:app_candidate_object_level_factors}
\end{figure}

As shown in Figure \ref{fig:app_candidate_object_level_factors}, we compare the distributions of the  object-level factor candidates on both synthetic and real-world datasets. It can be seen that the majority of these factors do not show significant gaps between the simple synthetic and the challenging real-world datasets. Therefore, we do not conduct relevant ablation experiments.

\subsubsection{Candidates of Scene-level Complexity Factors }

\begin{itemize}[leftmargin=*]
\setlength{\itemsep}{1pt}
\setlength{\parsep}{1pt}
\setlength{\parskip}{1pt}
    \item \textbf{Inter-object Color Similarity with Chamfer Distance:}
    In the calculation of this factor, we first convert each pixel into a point in RGB space. In this way, each object can be represented by a point set in the RGB space. This factor is calculated between each pair of objects by measuring the Chamfer distance of two point sets in the RGB space. Since Chamfer distance is an asymmetric measurement, we calculate and average out the bidirectional Chamfer distances. Compared with Euclidean distance, this measurement favors the most similar colors between two objects.  
    \item \textbf{Inter-object Color Similarity with Hausdorff Distance:} This factor is similar to the previous one. The only difference is that we replace Chamfer distance with Hausdorff distance. Hausdorff distance is also a directed and asymmetric measurement, so the final score is the average of distance values in both direction.
    \item \textbf{Inter-object Shape Similarity over Boundaries:}
    For each object mask, we first find its boundary using the method in \cite{Cheng2021a}, and then crop it with its axis-aligned bounding box. Each bounding box is scaled and fit into a unit box with its original aspect ratio. Lastly, we calculate the IoU between the boundaries of two objects to measure their shape similarity.
    \item \textbf{Inter-object Shape Entropy between Boundaries:} 
    We first combine all object masks into a single image by assigning different indices to different objects. Then we compute the entropy of each pixel with a $3 \times 3$ filter. The final factor score is calculated by averaging all non-zero entropy values. Note that, the interior part of objects and background will not be considered because their entropy values will always be zeros. Basically, this factor is designed to evaluate how crowded an image is. The higher this factor, the more objects are spatially adjacent.
    \item \textbf{Inter-object Proximity between Centroids:}
    We first calculate the centroid (${\Bar{x}, \Bar{y} }$) of each object by averaging all pixel coordinates in the object mask. Euclidean distances between object centroids are then computed pair-wisely before they are averaged to be the final factor score. This factor is designed to measure the spatial proximity of multiple objects in a single image.
    \item \textbf{Inter-object Proximity with Chamfer Distance:}
    In order to measure the spatial proximity between objects, we also calculate the spatial Chamfer distance between objects. Specifically, each object is represented by a set of $x-y$ coordinates, and then the average of pair-wise bidirectional Chamfer distances is calculated as the proximity score for each image.
\end{itemize}

\begin{figure}[ht]
    \setlength{\abovecaptionskip}{ -2 pt}
    \setlength{\belowcaptionskip}{ -8 pt}
    \centering
       \includegraphics[width=1\linewidth]{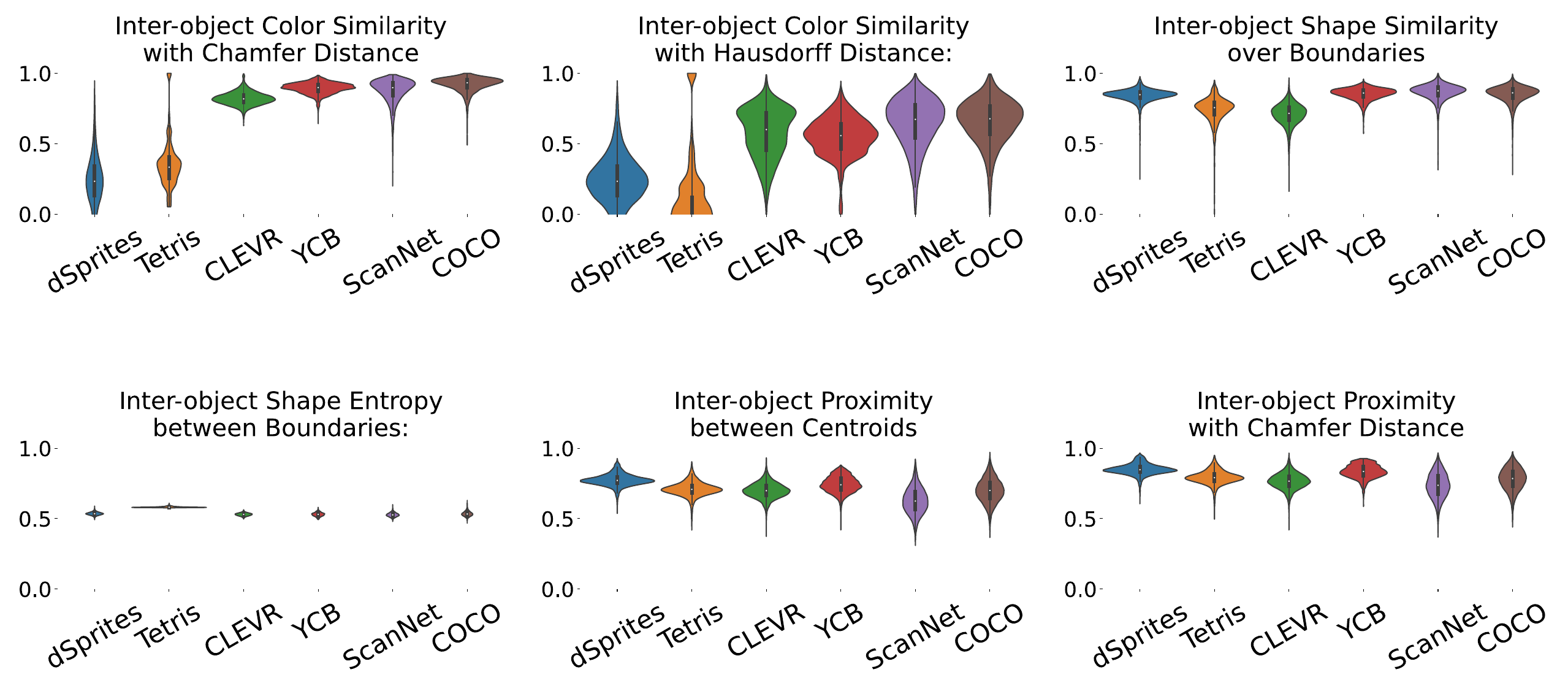}
      \vspace{0.05cm}
    \caption{Distributions of Scene-level Complexity Factors candidates.}
    \label{fig:app_candidate_scene_level_factors}
\end{figure}

As shown in Figure \ref{fig:app_candidate_scene_level_factors}, we compare the distributions of the scene-level factor candidates on both synthetic and real-world datasets. We can see that both the inter-object color similarity with Chamfer and Hausdorff distances share similar distributions gaps with our primary inter-object color similarity factor defined in Section \ref{sec:complexity_factors}. The remaining four candidate factors relating to inter-object shape complexity do not show significant distribution gaps between the synthetic and real-world datasets. In this regard, we choose not to conduct ablation experiments on these six candidate factors. 

\clearpage
\subsection{Implementation Details}
    \vspace{-0.2cm}
    In this section, we present the implementation details of four representative models.

\textbf{AIR \cite{Eslami2016}}
\vspace{-0.2cm}
\begin{itemize}[leftmargin=*]
    \setlength{\itemsep}{1pt}
    \setlength{\parsep}{1pt}
    \setlength{\parskip}{1pt}
    \item \textbf{Source Code:} We refer to \url{https://pyro.ai/examples/air.html} and \url{https://github.com/addtt/attend-infer-repeat-pytorch} for the implementation.
    \item \textbf{Important Adaptations:} We use an additional parameter to weight the KL divergence loss and reconstruction loss. For each experiment, we choose the weight for KL divergence from 1, 10, 25 and 50. The highest AP score is kept.
    \item \textbf{Training Details:} All experiments of AIR \cite{Eslami2016} are conducted with a batch size of 64. The learning rate is set to $1\mathrm{e}{-4}$ for training inference networks and decoders, $1\mathrm{e}{-3}$ for baselines which is the same as the original paper. Since the number of objects in our datasets ranges between 2 and 6, we set the maximum number of steps at inference to be 6 for all experiments. All models are trained on a single GPU for 1000 epochs. We perform evaluation every 50 epochs and select the one with the highest AP score.
\end{itemize}
        
\textbf{MONet \cite{Burgess2019}}
\vspace{-0.2cm}
\begin{itemize}[leftmargin=*]
    \setlength{\itemsep}{1pt}
    \setlength{\parsep}{1pt}
    \setlength{\parskip}{1pt}
    \item \textbf{Source Code:} We refer to \cite{Engelcke2020}'s re-implementation at: \url{https://github.com/applied-ai-lab/genesis}.
    \item \textbf{Important Adaptations:} We train MONet \cite{Burgess2019} with the GECO objective \cite{rezende2018taming} following the protocol mentioned in \cite{Engelcke2020}.
    \item \textbf{Training Details:} All experiments on MONet \cite{Burgess2019} are conducted with a batch size of 32 and learning rate of $1\mathrm{e}{-4}$. Since the maximum number of components is 7, including 1 background and 6 objects, we set the number of steps to be 7 for all experiments. All models are trained on a single GPU for 200 epochs with the training loss converged. We perform evaluation every 10 epochs and select the one with the highest AP score.
\end{itemize}
    
\textbf{IODINE \cite{Greff2019}}
\vspace{-0.2cm}
\begin{itemize}[leftmargin=*]
    \setlength{\itemsep}{1pt}
    \setlength{\parsep}{1pt}
    \setlength{\parskip}{1pt}
    \item \textbf{Source Code:} We use the official implementation at: \url{https://github.com/deepmind/deepmind-research/tree/master/iodine} .
    \item \textbf{Important Adaptations:} The architecture is set the same as what is used for CLEVR dataset \cite{Johnson2017} in the original paper \cite{Greff2019}.
    \item \textbf{Training Details:} Since we use a single GPU for the training of all models, the batch size is adjusted to be 4 and learning rate $0.0001 \times \sqrt{1/8}$. The number of slots $K$ is set as 7 and the inference iteration $T$ as 5. We train each model for $500K$ iterations until the loss is fully converged.
\end{itemize}
    
\textbf{SlotAtt \cite{Locatello2020}}
\vspace{-0.2cm}
\begin{itemize}[leftmargin=*]
    \setlength{\itemsep}{1pt}
    \setlength{\parsep}{1pt}
    \setlength{\parskip}{1pt}
    \item \textbf{Source Code:} We use the official implementation at: \url{https://github.com/google-research/google-research/tree/master/slot_attention}.
    \item \textbf{Important Adaptations:} The architecture is set the same as what is used for CLEVR dataset \cite{Johnson2017} in the original paper \cite{Locatello2020}.
    \item \textbf{Training Details:} All experiments of SlotAtt \cite{Locatello2020} are conducted with a batch size of 32 and learning rate selected from [$4\mathrm{e}{-4}$, $4\mathrm{e}{-5}$]. The number of slots $K$ is set as 7 and the number of iterations $T$ is set as 3. All models are trained on a single GPU for $500K$ iterations until the loss is fully converged.
\end{itemize}

\textbf{{Mask R-CNN} \cite{He2017a}}
\vspace{-0.2cm}
\begin{itemize}[leftmargin=*]
    \setlength{\itemsep}{1pt}
    \setlength{\parsep}{1pt}
    \setlength{\parskip}{1pt}
    \item \textbf{{Source Code:}} {We use the implementation at:} \url{https://github.com/matterport/Mask_RCNN}.
    \item \textbf{{Important Adaptations:}} {We use the same settings as training for COCO in above repository}. 
    \item \textbf{{Training Details:}} {Training for all datasets starts from the pre-trained COCO weights (mask\_rcnn\_coco.h5) from} \url{https://github.com/matterport/Mask_RCNN/releases}. {All models are trained on a single GPU for 30 epochs until the loss is fully converged.}
\end{itemize}    

\clearpage
\subsection{Details of six Benchmark Datasets}
In this section, we present the details of three synthetic and three real-world datasets.

\textbf{dSprites \cite{Matthey2017}}
To generate a specific image for this dataset, we first sample a random integer $K$ from a uniform distribution with interval $[2, 6]$ as the number of objects in that image. Then, $K$ object shapes are selected from the binary dsprites dataset \cite{Matthey2017} also in a uniformly random manner. Each object is assigned with a random RGB color by sampling three random integers from a uniform distribution with interval $[0, 255]$. In total, we generate 10000 images for training, 2000 for testing.
    
\textbf{Tetris \cite{Kabra2019}}
For each image in this dataset, we first sample a random integer $K$ from a uniform distribution with interval $[2, 6]$ as the number of objects in this image. To render one tetris-like object onto the canvas, we randomly pick up a tetris object from a randomly selected image from \cite{Kabra2019}. Each object is resized to be $88 \times 88$ and then placed onto the canvas. The position of each object is also sampled from a uniform distribution with 2 criteria: 1) all objects shall be on the canvas with complete shapes; 2) all objects shall not overlap with each other.
    
\textbf{CLEVR \cite{Johnson2017}}
We first generate CLEVR images following \url{https://github.com/facebookresearch/clevr-dataset-gen}, where the number of objects per image is restricted between 3 to 6. 
Given generated images with a resolution $640 \times 480$, we perform center-cropping and then resize them to be $128 \times 128$. Then, we remove tiny objects which have less than 35 pixels from each image. Subsequently, the images with less than 2 objects are removed. Being consistent with previous 2 synthetic datasets, all images have a black background.
    
\textbf{YCB \cite{Calli2017}}
We sample single frames from the YCB video dataset \cite{Calli2017} every 20 images. Given the sampled frames with a resolution of $640 \times 480$, we first center crop and then resize them to be $128 \times 128$. Then, the images consisting of less than 2 or more than 6 objects are removed. Similarly, all background pixels are replaced by a black color.
    
\textbf{ScanNet \cite{Dai2017}}
We sample single frames from the ScanNet dataset \cite{Dai2017} every 20 images. Given the selected frames with a resolution of $1296 \times 968$, we first center crop the images with a size of $800 \times 800$ and then resize them to be $128 \times 128$. For each resized image, we remove objects that contain more than $128 \times 128 \times 0.2$ pixels or less than $128 \times 128 \times 0.007$ pixels. The images with less than 2 or more than 6 are also dropped. All background pixels are replaced by the black color.
    
\textbf{COCO \cite{Lin2014}}
Given images in COCO-2017 \cite{Lin2014} with various resolutions, we first center crop and then resize images to be $128 \times 128$. For each resized image, we use the same criteria applied to ScanNet to remove too large and too small objects. The images with $2\sim 6$ objects are kept.
All background pixels are replaced by the black color. Since the number of images that could meet all requirements is less than 2,000 (only 1,597) in the official validation split, we additionally select 403 different images from its official training split for testing. There is no overlap between our training and testing splits.

\begin{figure}[ht]
\setlength{\abovecaptionskip}{ -0 pt}
\setlength{\belowcaptionskip}{ -8 pt}
\centering
   \includegraphics[width=0.8\linewidth]{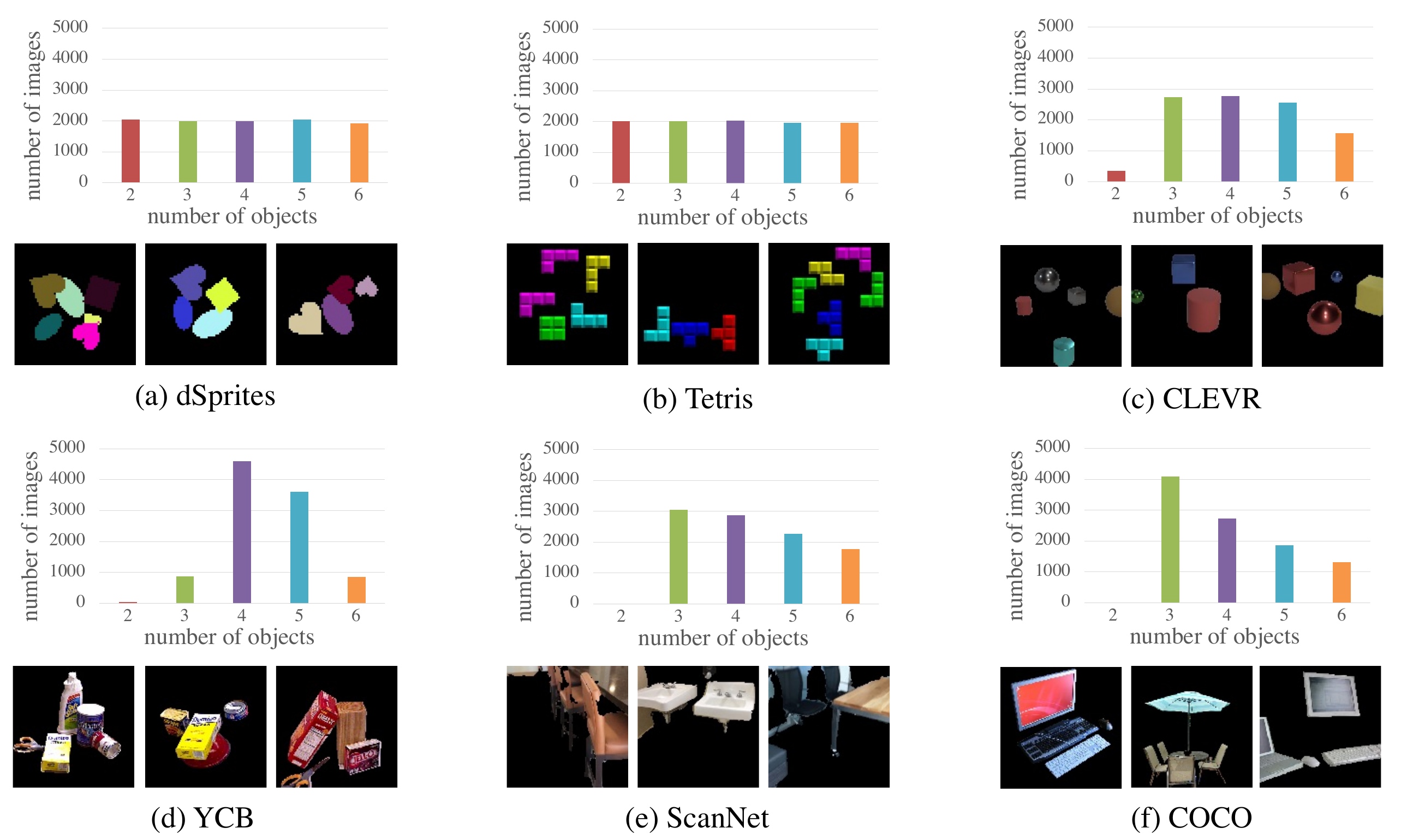}
\caption{Example images and the statistics of the six benchmark datasets.}
\label{fig:app_datasets}
\end{figure}

Figure \ref{fig:app_datasets} shows three example images for each dataset. The distribution of images with different number of objects is also presented.  

\clearpage
\subsection{Details and Results of Main Experiments}
\vspace{-0.15cm}
In this section, we present the result of four unsupervised methods and one supervised method on six datasets. Both quantitative evaluation and qualitative results are provided.
\subsubsection{Four Unsupervised Methods}
\begin{figure}[ht]
    \setlength{\abovecaptionskip}{ 0 pt}
    \setlength{\belowcaptionskip}{ -2 pt}
    \centering
      \includegraphics[width=0.85\linewidth]{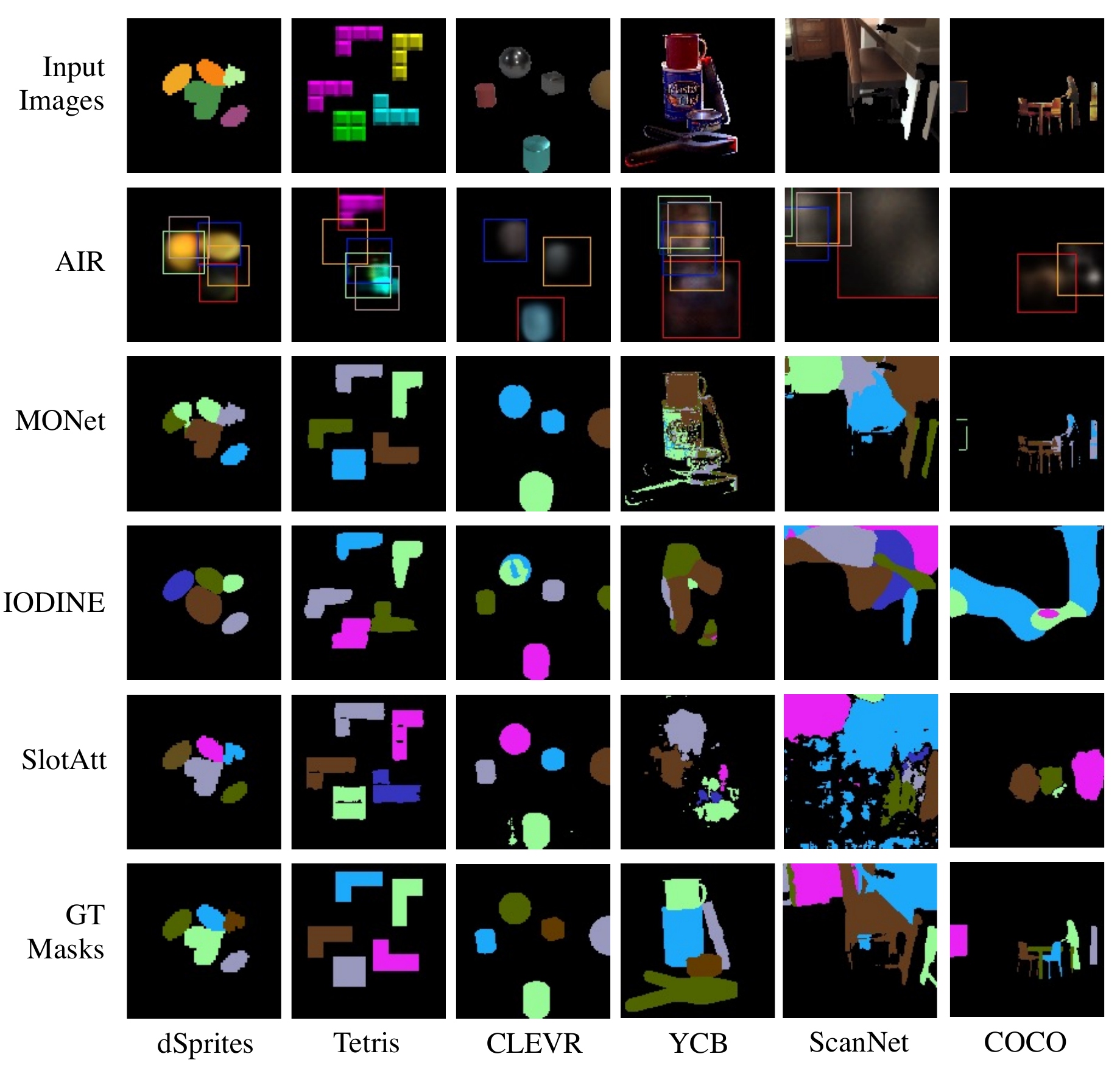}
    \caption{Qualitative results of object segmentation from the four methods on six datasets.}
    \label{fig:app_main_experiments}
    \vspace{-0.25cm}
\end{figure}

Figure \ref{fig:app_main_experiments} shows the qualitative results of 4 representative methods on 6  datasets. The results of AIR \cite{Eslami2016} are presented with predicted bounding boxes while others are represented with predicted segmentation masks. Different colors in a segmentation mask indicate different objects. For the same object, the assigned color may not be the same. We can see that all methods give reasonable results on the three synthetic datasets in spite of some performance gaps between them due to the different capability of these methods. However, they all fail on the three real-world datasets. Specifically, AIR \cite{Eslami2016} cannot reconstruct reasonable images. MONet \cite{Burgess2019} always performs segmentation based on color. IODINE \cite{Greff2019} cannot recover meaningful object shapes and locations. SlotAtt \cite{Locatello2020} can roughly locate objects but cannot identify the real object shapes. 

\setlength{\abovecaptionskip}{-10 pt}
\setlength{\belowcaptionskip}{4 pt}
\begin{table}[tbh]
\tiny
\vspace{-0.25cm}
  \caption{Quantitative results of object segmentation from the four methods on six datasets. Standard deviations of performance are calculated over 3 runs (marked with \textcolor{blue}{{blue}}{).}}
  \tabcolsep= 0.1cm 
  \label{tab:ablation_main_res}
  \centering
  \begin{tabular}{r c|c|c} \toprule
  & dSprites  & Tetris  & CLEVR   \\ \midrule
              & AP / PQ / Pre / Rec  & AP / PQ / Pre / Rec & AP / PQ / Pre / Rec  \\ 
AIR \cite{Eslami2016} & 45.4 \textcolor{blue}{1.8} / 38.2 \textcolor{blue}{3.0} / 57.6 \textcolor{blue}{7.4} / 58.1 \textcolor{blue}{7.5} & 25.2 \textcolor{blue}{13.9} / 23.4 \textcolor{blue}{12.4} / 36.8 \textcolor{blue}{20.9} / 39.9 \textcolor{blue}{12.9} & 46.4 \textcolor{blue}{14.0} / 44.3 \textcolor{blue}{12.4} / 67.4 \textcolor{blue}{9.9} / 52.5 \textcolor{blue}{15.9} \\ 
MONet \cite{Burgess2019} & 69.7 \textcolor{blue}{4.1} / 61.6 \textcolor{blue}{6.0} / 70.4 \textcolor{blue}{8.1} / 73.9 \textcolor{blue}{1.9} & 85.9 \textcolor{blue}{13.0} / 75.8 \textcolor{blue}{13.6} / 85.1 \textcolor{blue}{16.4}/ 89.7  \textcolor{blue}{8.2} & 39.0 \textcolor{blue}{8.5} / 37.3 \textcolor{blue}{6.3} / 65.6 \textcolor{blue}{11.8} / 42.8 \textcolor{blue}{10.8}  \\ 
IODINE \cite{Greff2019} & 92.9  \textcolor{blue}{4.3}/ 71.3  \textcolor{blue}{6.1}/ 82.6  \textcolor{blue}{2.3}/ 96.0  \textcolor{blue}{5.2} & 52.2 \textcolor{blue}{2.3} / 37.9 \textcolor{blue}{4.6} / 48.0 \textcolor{blue}{2.3} / 61.8 \textcolor{blue}{1.7} & 82.8 \textcolor{blue}{2.8} / 73.0 \textcolor{blue}{5.7} / 77.5 \textcolor{blue}{3.1} / 87.4 \textcolor{blue}{2.0} \\ 
SlotAtt \cite{Locatello2020} & 92.8  \textcolor{blue}{1.4} / 82.8  \textcolor{blue}{1.6} / 88.8 \textcolor{blue}{3.4} / 92.9  \textcolor{blue}{1.6} & 94.3 \textcolor{blue}{1.2} / 79.9 \textcolor{blue}{6.4} / 90.5 \textcolor{blue}{3.3} / 94.4 \textcolor{blue}{1.3} & 91.7 \textcolor{blue}{6.4} / 82.9 \textcolor{blue}{10.9} / 90.8 \textcolor{blue}{9.7} / 92.7 \textcolor{blue}{5.3} \\ 
\midrule
& YCB  & ScanNet  & COCO   \\ \midrule
              & AP / PQ / Pre / Rec  & AP / PQ / Pre / Rec & AP / PQ / Pre / Rec  \\ 
AIR \cite{Eslami2016} & 0.0 \textcolor{blue}{0.1} / 0.6 \textcolor{blue}{0.3} / 1.1   \textcolor{blue}{0.4} / 0.8 \textcolor{blue}{0.2} & 2.7 \textcolor{blue}{1.4} / 6.3 \textcolor{blue}{1.7} / 15.6 \textcolor{blue}{2.8} / 7.3 \textcolor{blue}{1.6} & 2.7 \textcolor{blue}{0.1} / 6.7 \textcolor{blue}{0.5} / 14.3 \textcolor{blue}{2.6} / 8.6 \textcolor{blue}{0.8} \\
MONet \cite{Burgess2019} & 3.1 \textcolor{blue}{1.6} / 7.0 \textcolor{blue}{2.6} / 9.8 \textcolor{blue}{3.6} / 1.2 \textcolor{blue}{0.8} & 24.8 \textcolor{blue}{1.6} / 24.6 \textcolor{blue}{1.6} / 31.0 \textcolor{blue}{1.6} / 40.7 \textcolor{blue}{1.8} & 11.8 \textcolor{blue}{2.0} / 12.5 \textcolor{blue}{1.1}/ 16.1 \textcolor{blue}{0.9}/ 21.9  \textcolor{blue}{1.7} \\
IODINE \cite{Greff2019} & 1.8 \textcolor{blue}{0.2} / 3.9 \textcolor{blue}{1.3} / 6.2 \textcolor{blue}{2.0} / 7.3 \textcolor{blue}{1.9} & 10.1 \textcolor{blue}{2.9} / 13.7 \textcolor{blue}{2.7} / 18.6 \textcolor{blue}{4.2} / 24.4 \textcolor{blue}{3.8} & 4.0 \textcolor{blue}{1.2} / 6.3 \textcolor{blue}{1.2} / 9.9 \textcolor{blue}{1.8} / 10.8 \textcolor{blue}{2.0} \\ 
SlotAtt \cite{Locatello2020} & 9.2 \textcolor{blue}{0.4} / 13.5 \textcolor{blue}{0.9} / 20.0 \textcolor{blue}{1.3} / 26.2 \textcolor{blue}{6.8} & 5.7 \textcolor{blue}{0.3} / 9.0 \textcolor{blue}{1.5} / 12.4 \textcolor{blue}{2.5} / 18.3 \textcolor{blue}{2.7} & 0.8 \textcolor{blue}{0.3} / 3.5 \textcolor{blue}{1.2} / 5.3 \textcolor{blue}{1.7} / 7.3 \textcolor{blue}{2.2} \\ \bottomrule

  \end{tabular}
  \vspace{-0.2cm}
\end{table}

\subsubsection{One Supervised Method}
\begin{figure}[hb]
\vspace{-0.2cm}
\centering
\begin{minipage}{.6\textwidth}
  \centering
  \captionsetup{width=0.98\textwidth}
  \includegraphics[width=1\linewidth]{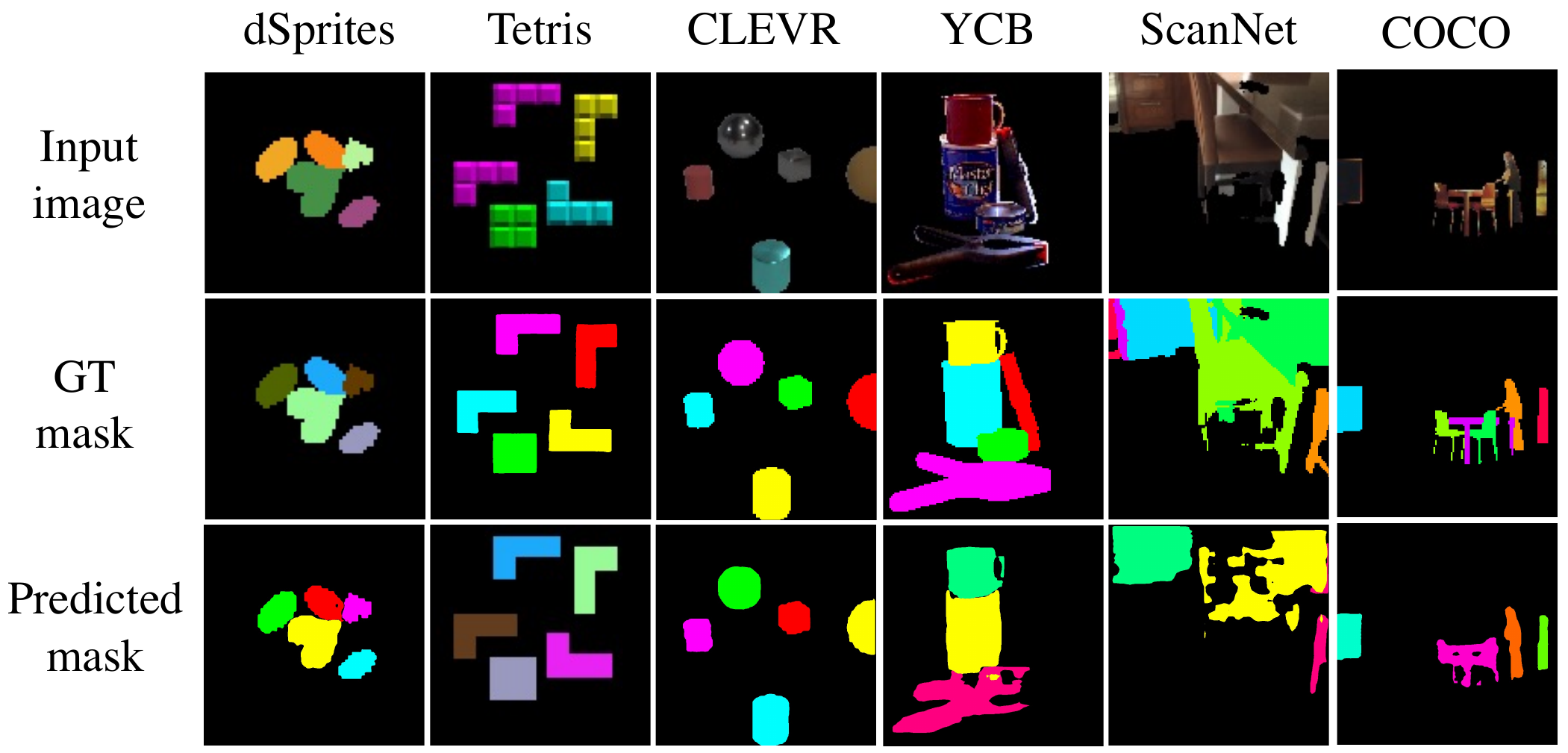}
\end{minipage}%
\hspace{0.1cm}
\begin{minipage}{.25\textwidth}
\tabcolsep= 0.1cm 
  \label{tab:maskrcn_tab}
  \centering
  \small
  \begin{tabular}{r c} \toprule
      &  MaskRCNN \\ \midrule
              & AP / PQ / Pre / Rec   \\ 
      dSprites  & 98.4 / 90.2 / 99.6 / 98.4 \\ 
      Tetris   & 99.8 / 90.3 / 99.8 / 99.8 \\
      CLEVR    & 98.2 / 90.0 / 97.8 / 99.5 \\ \hdashline
      YCB    & 62.9 / 58.4 / 83.3 / 66.9 \\
      ScanNet   & 41.4 / 43.3 / 65.2 / 50.5  \\
      COCO   & 46.0 / 47.9 / 71.7 / 53.2 \\ \bottomrule
  \end{tabular}
  \vspace{-0.5cm}
\end{minipage}
\vspace{0.5cm}
\caption[width=0.85\linewidth]{{Qualitative and quantitative results of Mask-RCNN on six datasets.}}
\label{fig:mask_rcnn_res}
\end{figure}

Apart from above four unsupervised approaches, we also include Mask R-CNN \cite{He2017a} as a supervised object segmentation baseline. As shown in Figure \ref{fig:mask_rcnn_res}, Mask R-CNN exhibits quite successful segmentation on three synthetic datasets. Real-world datasets are more challenging for Mask R-CNN,  but the results are much better than unsupervised baselines.

\subsection{Details and Results of Ablation Experiments}
In this section, we shows details of ablated datasets including example images. Both qualitative and quantitative evaluation are also presented. Table \ref{tab:ablation_lookup} is a look-up table for individual ablated factor.
\begin{table}[h]
   \caption{{A look-up table for ablations.}}
   \tabcolsep= 0.1cm 
  \label{tab:}
   \centering
 \begin{tabular}{ p{1cm}||p{1cm}|p{1cm}||p{1cm}|p{1cm}||p{1cm}|p{1cm}|p{1cm}|p{1cm}  }
     \hline
     \multicolumn{1}{c||}{}& \multicolumn{2}{c||}{Which level} &  \multicolumn{2}{c||}{Which aspect} &  \multicolumn{4}{c}{Target factor} \\
     \hline
     \multicolumn{1}{c||}{Ablation}& \multicolumn{1}{c|}{Object} &  \multicolumn{1}{c||}{Scene} &  \multicolumn{1}{c|}{Color}  &  \multicolumn{1}{c||}{Shape} &  \multicolumn{1}{C{.09\textwidth}|}{Object Color Gradient} &  \multicolumn{1}{C{.11\textwidth}|}{Object Shape Concavity} & \multicolumn{1}{C{.11\textwidth}|}{Inter-object Color Similarity} & \multicolumn{1}{C{.11\textwidth}}{Inter-object Shape Variation}
     \\ 
     \hline 
     C & \checkmark &  & \checkmark &  & \checkmark & & &\\\hline 
     S & \checkmark &  & & \checkmark & & \checkmark & & \\ \hline 
     T &  & \checkmark & \checkmark & & &  &\checkmark  &  \\ \hline 
     U &  & \checkmark & & \checkmark & & & & \checkmark \\ 
     \hline
 \end{tabular}\label{tab:ablation_lookup}
 \end{table}

\subsubsection{Ablations on Object-level Factors}
\paragraph{Ablated Datasets}

\begin{itemize}[leftmargin=*]
\setlength{\itemsep}{1pt}
\setlength{\parsep}{1pt}
\setlength{\parskip}{1pt}
    \item \textit{Ablation of Object Color Gradient}: In the three ablated datasets: YCB-C / ScanNet-C / COCO-C, each object is represented by a single average color. 
    \item \textit{Ablation of Object Shape Concavity}: In the three ablated datasets: YCB-S / ScanNet-S / COCO-S, each object is represented by a convex shape. 
    \item \textit{Ablation of both Object Color Gradient and Shape Concavity}: In the three ablated datasets: YCB-C+S / ScanNet-C+S / COCO-C+S, each object is represented by a convex shape with a single average color. 
\end{itemize}

\begin{figure}[ht]
    \setlength{\abovecaptionskip}{ 0 pt}
    \setlength{\belowcaptionskip}{ -2 pt}
    \centering
      \includegraphics[width=0.5\linewidth]{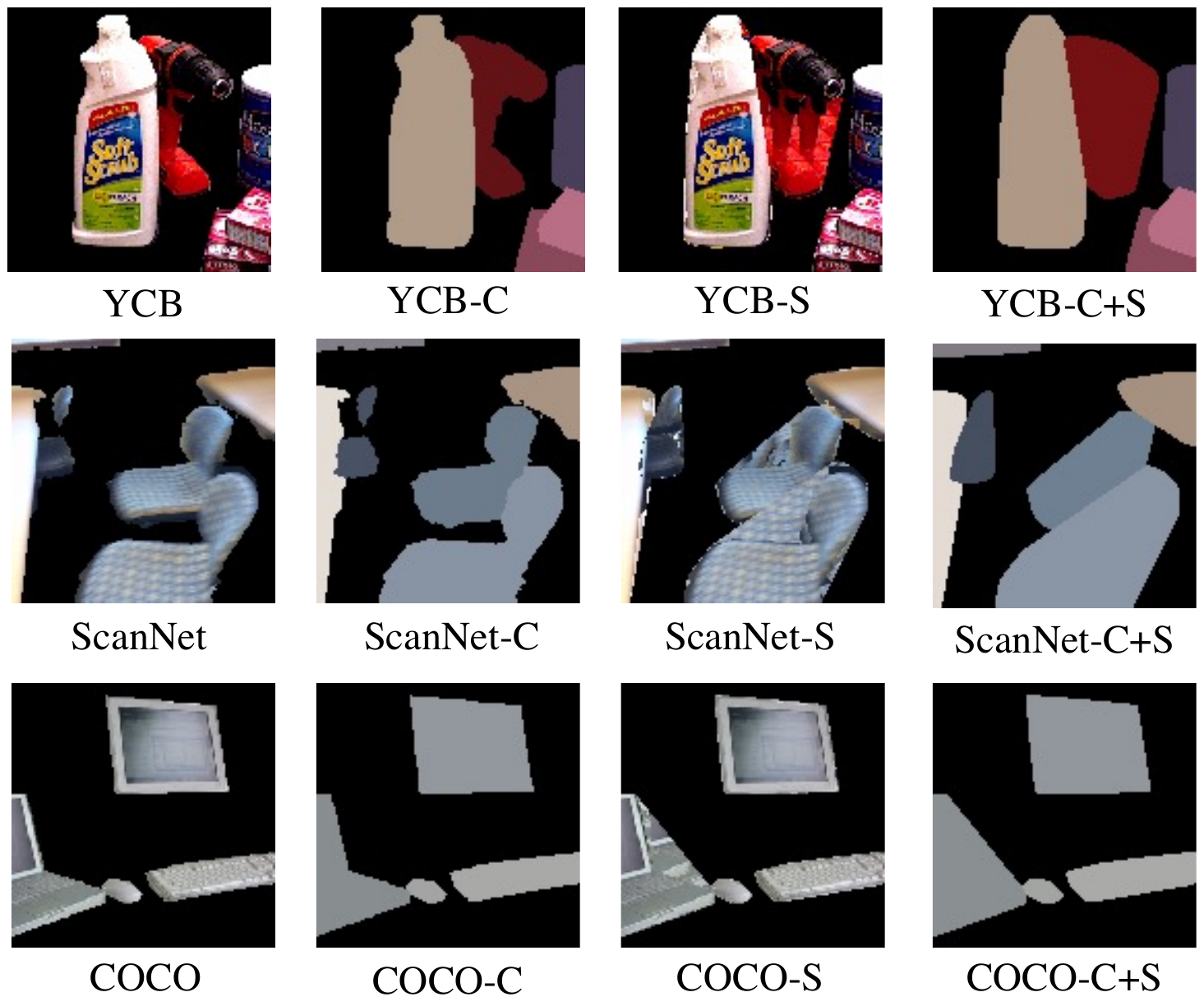}
      \vspace{0.1cm}
    \caption{Example images of real-world datasets ablated with object-level factors.}
    \label{fig:dataset_S_C}
     \vspace{-0.4cm}
\end{figure}

Figure \ref{fig:dataset_S_C} shows example images from the three real-world datasets with different types of object-level factor ablations. 

\begin{figure}[bht]
    \setlength{\abovecaptionskip}{ 0 pt}
    \setlength{\belowcaptionskip}{ -2 pt}
    \centering
      \includegraphics[width=0.9\linewidth]{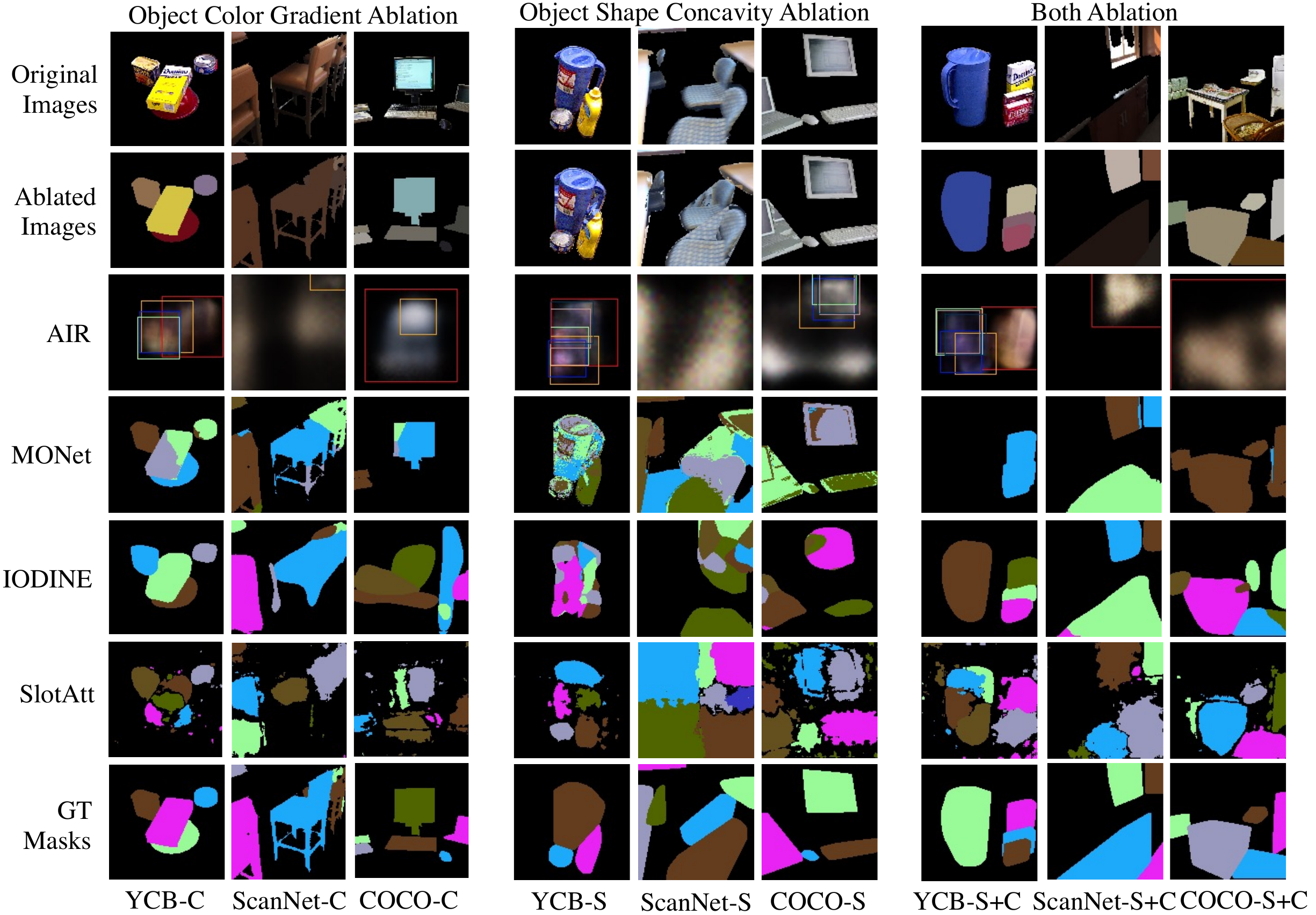}
      \vspace{0.1cm}
    \caption{Qualitative results on the datasets ablated with object-level factors.}
    \label{fig:app_S_C_experiments}
     \vspace{-.5cm}
\end{figure}

\paragraph{Qualitative and Quantitative Results}
As shown in Figure \ref{fig:app_S_C_experiments} and Table \ref{tab:app_S_C_experiments}, all four methods have a significant improvement in segmentation performance on the ablated datasets with object color gradients being removed. By comparison, the ablation of object shape concavity is less effective. This shows that both object-level factors are relevant to the success of object segmentation, although object color gradient is more important for the four methods. Specifically, MONet \cite{Burgess2019} and IODINE \cite{Greff2019} are more sensitive to color gradient compared with the other two methods. 

\setlength{\abovecaptionskip}{-8 pt}
\setlength{\belowcaptionskip}{-4 pt}
\begin{table}[h]
\tiny
  \caption{Quantitative results on the datasets ablated with object-level factors. Standard deviations of performance are calculated over 3 runs (marked with \textcolor{blue}{blue}). }
  \tabcolsep= 0.1cm 
  \label{tab:app_S_C_experiments}
  \centering
  \begin{tabular}{r c|c|c} \toprule
      & YCB-C & ScanNet-C & COCO-C  \\ \midrule
              & AP / PQ / Pre / Rec  & AP / PQ / Pre / Rec & AP / PQ / Pre / Rec \\ 
AIR \cite{Eslami2016} & 4.4 \textcolor{blue}{0.2} / 1.0 \textcolor{blue}{7.3} / 20.7 \textcolor{blue}{7.1} / 12.5 \textcolor{blue}{1.2} & 2.7 \textcolor{blue}{1.0} / 7.6 \textcolor{blue}{2.1} / 29.1 \textcolor{blue}{20.2} / 7.4 \textcolor{blue}{1.5} & 2.9 \textcolor{blue}{1.8} / 7.9 \textcolor{blue}{2.5} / 32.0 \textcolor{blue}{29.3} / 7.7 \textcolor{blue}{4.5} \\ 
MONet \cite{Burgess2019} & 55.1 \textcolor{blue}{6.0} / 49.5 \textcolor{blue}{5.2} / 64.8 \textcolor{blue}{1.7} / 58.8 \textcolor{blue}{7.8} & 31.1 \textcolor{blue}{15.9} / 33.1 \textcolor{blue}{9.6} / 41.6 \textcolor{blue}{10.3} / 45.6 \textcolor{blue}{11.1} & 34.5 \textcolor{blue}{7.2} / 34.6 \textcolor{blue}{3.3} / 45.8 \textcolor{blue}{2.0} / 42.6 \textcolor{blue}{11.0}\\ 
IODINE \cite{Greff2019} & 76.8 \textcolor{blue}{0.3} / 64.4 \textcolor{blue}{0.1} / 76.5 \textcolor{blue}{0.1} / 79.5 \textcolor{blue}{0.0} & 64.4 \textcolor{blue}{6.4} / 54.3 \textcolor{blue}{5.9} / 66.3 \textcolor{blue}{4.6} / 71.0 \textcolor{blue}{5.9} & 42.8 \textcolor{blue}{14.7} / 34.0 \textcolor{blue}{11.8} / 42.5 \textcolor{blue}{14.4} / 55.8 \textcolor{blue}{11.9}\\ 
SlotAtt \cite{Locatello2020} & 39.2 \textcolor{blue}{1.9} / 30.2 \textcolor{blue}{0.0} / 40.2 \textcolor{blue}{0.8} / 50.4 \textcolor{blue}{0.7} & 5.7 \textcolor{blue}{8.2} / 9.0 \textcolor{blue}{6.1} / 12.4 \textcolor{blue}{7.6} / 18.3 \textcolor{blue}{8.9} & 7.2 \textcolor{blue}{1.9} / 9.8  \textcolor{blue}{1.7}/ 13.9 \textcolor{blue}{2.2} / 18.7 \textcolor{blue}{3.4}\\ 
\midrule
& YCB-S  & ScanNet-S  & COCO-S   \\ \midrule
              & AP / PQ / Pre / Rec  & AP / PQ / Pre / Rec & AP / PQ / Pre / Rec  \\
AIR \cite{Eslami2016} & 3.6 \textcolor{blue}{3.0} / 6.8 \textcolor{blue}{3.2} / 12.5 \textcolor{blue}{6.7} / 9.7 \textcolor{blue}{2.7} & 3.3 \textcolor{blue}{2.5} / 8.5 \textcolor{blue}{1.1} / 36.6 \textcolor{blue}{23.0} / 8.1 \textcolor{blue}{8.5} & 3.3 \textcolor{blue}{0.0} / 7.1 \textcolor{blue}{1.1} / 16.8 \textcolor{blue}{5.6} / 8.3 \textcolor{blue}{1.0}\\ 
MONet \cite{Burgess2019} & 2.5 \textcolor{blue}{1.1} / 6.4 \textcolor{blue}{1.8} / 8.9 \textcolor{blue}{2.6} / 11.3 \textcolor{blue}{3.2} & 18.3 \textcolor{blue}{10.4} / 22.2 \textcolor{blue}{3.9} / 28.4 \textcolor{blue}{4.4} / 37.5 \textcolor{blue}{5.3} & 8.7 \textcolor{blue}{3.2} / 11.2 \textcolor{blue}{1.7} / 14.4 \textcolor{blue}{3.0} / 21.0 \textcolor{blue}{0.7}\\ 
IODINE \cite{Greff2019} & 2.5 \textcolor{blue}{2.5} / 5.0 \textcolor{blue}{5.0} / 7.9 \textcolor{blue}{7.8} / 1.1 \textcolor{blue}{1.0} & 8.3 \textcolor{blue}{1.1} / 13.7 \textcolor{blue}{0.4} / 18.4 \textcolor{blue}{0.4} / 24.8 \textcolor{blue}{0.7} & 4.2 \textcolor{blue}{0.4} / 8.3 \textcolor{blue}{0.4} / 12.1 \textcolor{blue}{0.3} / 14.8 \textcolor{blue}{0.6} \\ 
SlotAtt \cite{Locatello2020} & 19.1 \textcolor{blue}{0.7} / 21.0 \textcolor{blue}{1.7} / 30.4 \textcolor{blue}{2.8} / 37.7 \textcolor{blue}{0.8} & 1.9 \textcolor{blue}{8.5} / 6.6 \textcolor{blue}{7.5} / 8.7 \textcolor{blue}{9.8} / 13.7 \textcolor{blue}{11.9} & 2.6 \textcolor{blue}{1.0} / 5.9 \textcolor{blue}{0.9} / 8.3 \textcolor{blue}{1.3} / 11.9 \textcolor{blue}{10.9}\\ 
\midrule
& YCB-C+S  & ScanNet-C+S  & COCO-C+S   \\ \midrule
              & AP / PQ / Pre / Rec  & AP / PQ / Pre / Rec & AP / PQ / Pre / Rec  \\ 
AIR \cite{Eslami2016} & 2.0 \textcolor{blue}{0.3} / 6.4 \textcolor{blue}{0.2} / 15.1 \textcolor{blue}{5.2} / 7.3 \textcolor{blue}{4.1} & 3.5 \textcolor{blue}{1.0} / 9.1 \textcolor{blue}{0.5} / 38.3 \textcolor{blue}{25.8} / 8.5 \textcolor{blue}{5.9} & 4.8 \textcolor{blue}{0.1} / 10.8 \textcolor{blue}{0.1} / 41.5 \textcolor{blue}{10.2} / 10.1 \textcolor{blue}{1.7}\\ 
MONet \cite{Burgess2019} & 9.1 \textcolor{blue}{13.6} / 12.7 \textcolor{blue}{10.5} / 34.1 \textcolor{blue}{4.6} / 12.6 \textcolor{blue}{24.0} & 48.1 \textcolor{blue}{1.0} / 47.9 \textcolor{blue}{6.8} / 70.4 \textcolor{blue}{23.3} / 51.5 \textcolor{blue}{6.5} & 10.5 \textcolor{blue}{24.3} / 16.5 \textcolor{blue}{18.1} / 56.5 \textcolor{blue}{6.9} / 14.9 \textcolor{blue}{28.2}\\ 
IODINE \cite{Greff2019} & 69.9 \textcolor{blue}{30.9} / 55.7 \textcolor{blue}{29.6} / 60.3 \textcolor{blue}{28.3} / 71.5 \textcolor{blue}{26.4} & 73.8 \textcolor{blue}{11.4} / 63.2 \textcolor{blue}{10.1} / 73.5 \textcolor{blue}{6.1} / 76.8 \textcolor{blue}{10.8} & 73.5 \textcolor{blue}{2.5} / 61.2 \textcolor{blue}{2.4} / 70.0 \textcolor{blue}{2.9} / 77.6 \textcolor{blue}{1.2} \\ 
SlotAtt \cite{Locatello2020} & 39.1 \textcolor{blue}{22.3} / 30.2 \textcolor{blue}{13.3} / 42.9 \textcolor{blue}{12.2} / 54.5 \textcolor{blue}{12.7} & 22.5 \textcolor{blue}{7.6} / 21.5 \textcolor{blue}{6.2} / 28.8 \textcolor{blue}{6.0} / 39.6 \textcolor{blue}{5.4} & 20.3 \textcolor{blue}{3.0} / 19.9 \textcolor{blue}{0.4} / 26.1 \textcolor{blue}{0.4} / 34.8 \textcolor{blue}{0.7} \\ \bottomrule
               
  \end{tabular}
  \vspace{-0.4cm}
\end{table}

\clearpage
\subsubsection{Ablations on Scene-level Factors}
\paragraph{Ablated Datasets}

\begin{figure}[hb]
\centering
\begin{minipage}{.35\textwidth}
  \centering
  \captionsetup{width=0.95\textwidth}
  \vspace{3.23cm}
  \includegraphics[width=0.95\linewidth]{figs/app_texture_source.pdf}
  \vspace{0.53cm}
  \caption[width=0.85\linewidth]{Six selected texture images from DTD \cite{Cimpoi2014}.}
  \label{fig:dtd_texture}
\end{minipage}%
\begin{minipage}{.6\textwidth}
  \centering
  \captionsetup{width=0.9\textwidth}
  \vspace{0.25cm}
  \includegraphics[width=0.9\linewidth]{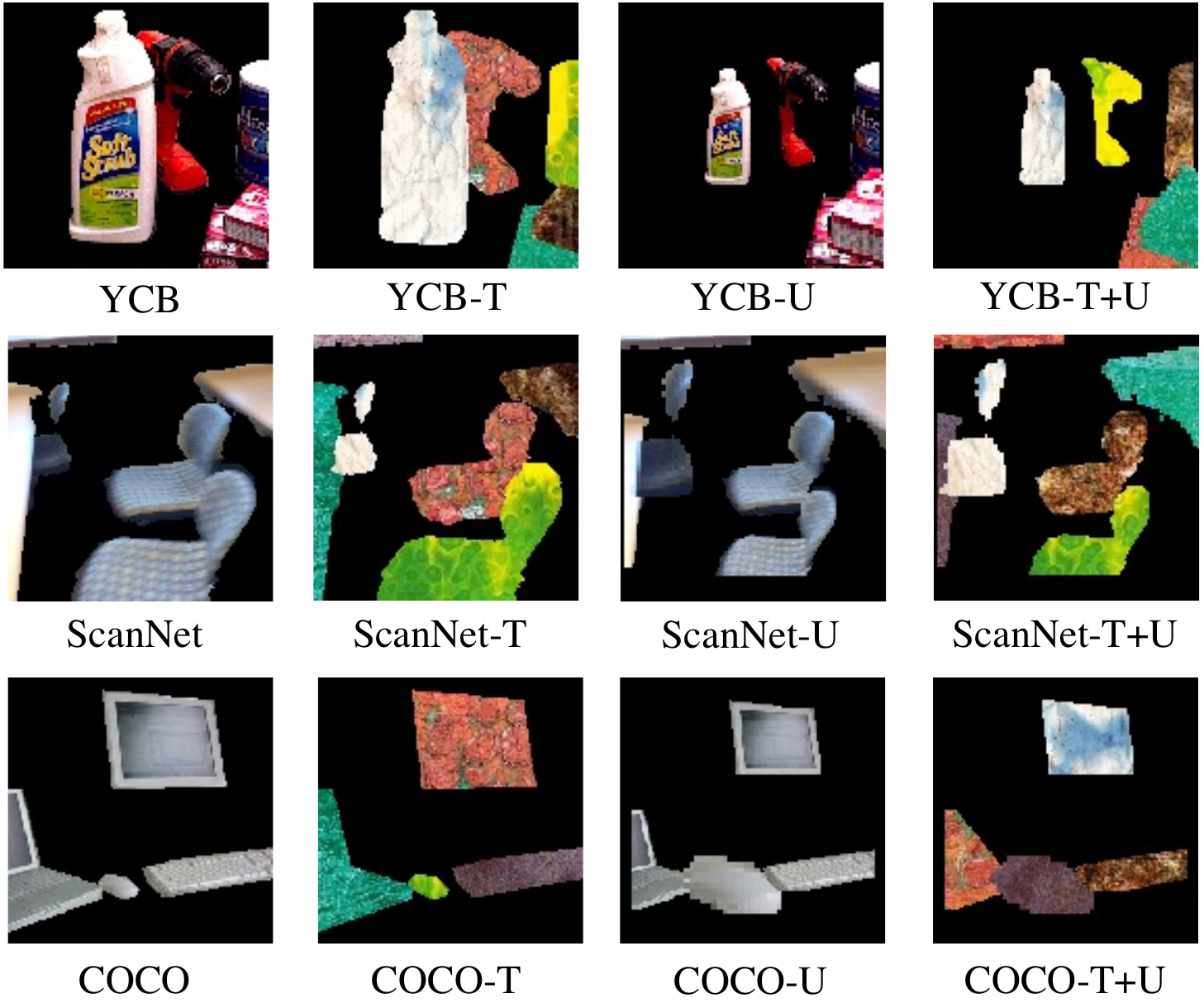}
  \vspace{0.35cm}
  \caption[width=0.85\linewidth]{Example images of datasets ablated with scene-level factors.}
  \label{fig:dataset_T_U}
\end{minipage}
\end{figure}

\begin{itemize}[leftmargin=*]
\setlength{\itemsep}{1pt}
\setlength{\parsep}{1pt}
\setlength{\parskip}{1pt}
    \item \textit{Ablation of Inter-object Color Similarity}: In the three ablated datasets: YCB-T / ScanNet-T / COCO-T, each object's original texture is replaced by one of the selected 6 distinctive textures from the DTD database \cite{Cimpoi2014}, as shown in Figure \ref{fig:dtd_texture}. 
    In particular, the 6 textures are chosen from the `blotchy' category, and we deliberately select those with distinctive colors. The texture images are center-cropped and resized to a resolution of $128 \times 128$. Each object appearance is replaced by one of the 6 texture images, and each object in a single image is assigned with a different texture. 
    \item \textit{Ablation of Inter-object Shape Variation}: In the three ablated datasets: YCB-U / ScanNet-U / COCO-U, the size of each object is uniformly scaled. In particular, before generating the ablated datasets, we first calculate the average scale of objects in each dataset by calculating the mean of object bounding box diagonals (YCB: 60 / ScanNet: 70 / COCO: 57, in pixels). Each object in a specific dataset is scaled up or down to approach the average scale in this dataset. The shape and appearance are also linearly scaled, so that the shape variation is the only factor changed in the new dataset.
    \item \textit{Ablation of both Inter-object Color Similarity and Shape Variation}: In the three ablated datasets: YCB-T+U / ScanNet-T+U / COCO-T+U, each object's texture is replaced and shape is uniformly scaled.    
\end{itemize}

Figure \ref{fig:dataset_T_U} shows example images from the three real-world datasets with different types of scene-level factor ablations.

\begin{figure}[ht]
    \setlength{\abovecaptionskip}{ 0 pt}
    \setlength{\belowcaptionskip}{ -2 pt}
    \centering
      \includegraphics[width=1\linewidth]{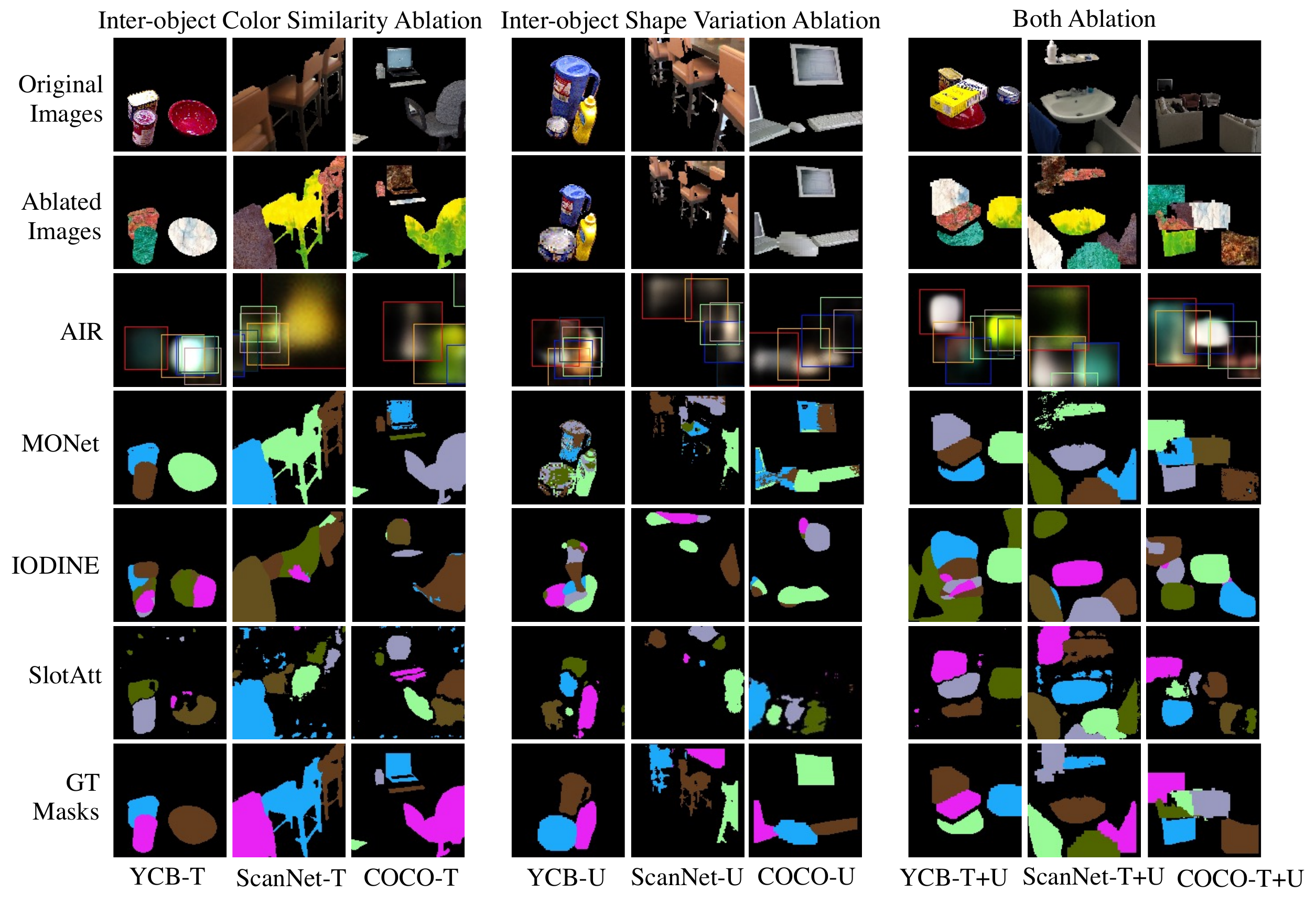}
    \caption{Qualitative results on the datasets ablated with scene-level factors.}
    \label{fig:app_T_U_experiments}
\end{figure}

\paragraph{Qualitative and Quantitative Results}
As shown in Figure \ref{fig:app_T_U_experiments} and Table \ref{tab:app_T_U_experiments}, all methods have a significant improvement in object segmentation on the ablated datasets with reduced inter-object color similarity. The effect of uniform scale ablation is limited standalone. However, when both ablations are combined, there is a larger performance gain compared with purely texture replaced ablation. Specifically, AIR \cite{Eslami2016} is sensitive to both scene-level factors, since only the combination of two factors can result in a significant improvement. MONet \cite{Burgess2019} and IODINE \cite{Greff2019}, however, are almost only sensitive to the inter-object color similarity. Both factors have a noticeable effect on the segmentation performance of SlotAtt \cite{Locatello2020}.

\clearpage
\setlength{\abovecaptionskip}{-6 pt}
\setlength{\belowcaptionskip}{4 pt}
\begin{table}[ht]
\tiny
  \caption{Quantitative results on the datasets ablated with scene-level factors. Standard deviations of performance are calculated over 3 runs (marked with \textcolor{blue}{blue}). }
  \tabcolsep= 0.06cm 
  \label{tab:app_T_U_experiments}
  \centering
  \begin{tabular}{r c|c|c} \toprule
      & YCB-T & ScanNet-T & COCO-T  \\ \midrule
              & AP / PQ / Pre / Rec  & AP / PQ / Pre / Rec & AP / PQ / Pre / Rec \\ 
AIR \cite{Eslami2016} & 5.9 \textcolor{blue}{3.7} / 9.6 \textcolor{blue}{3.3} / 19.9 \textcolor{blue}{10.3} / 12.7 \textcolor{blue}{1.3} & 2.9 \textcolor{blue}{1.2} / 6.3 \textcolor{blue}{0.1} / 13.4 \textcolor{blue}{3.7} / 8.6 \textcolor{blue}{3.2} & 7.4 \textcolor{blue}{3.1} / 13.0 \textcolor{blue}{4.0} / 29.1 \textcolor{blue}{9.8} / 16.4 \textcolor{blue}{4.3} \\ 
MONet \cite{Burgess2019} & 86.5 \textcolor{blue}{12.4} / 78.3 \textcolor{blue}{13.3} / 81.1 \textcolor{blue}{11.6} / 86.8 \textcolor{blue}{12.2} & 82.9 \textcolor{blue}{12.3} / 77.8 \textcolor{blue}{7.6} / 86.3 \textcolor{blue}{1.6} / 83.7 \textcolor{blue}{11.9} & 80.0 \textcolor{blue}{9.4} / 68.8 \textcolor{blue}{13.5} / 74.5 \textcolor{blue}{15.0} / 82.0 \textcolor{blue}{5.6} \\ 
IODINE \cite{Greff2019} & 32.4 \textcolor{blue}{9.0} / 27.3 \textcolor{blue}{6.8} / 35.3 \textcolor{blue}{8.3} / 43.6 \textcolor{blue}{10.6} & 33.2 \textcolor{blue}{11.9} / 27.3 \textcolor{blue}{6.3} / 34.5 \textcolor{blue}{7.0} / 44.7 \textcolor{blue}{9.3} & 40.8 \textcolor{blue}{12.7} / 33.7 \textcolor{blue}{10.2} / 41.8 \textbf{\textcolor{blue}{10.4}} / 55.9 \textcolor{blue}{15.8} \\ 
SlotAtt \cite{Locatello2020} & 64.6 \textcolor{blue}{5.3} / 48.5 \textcolor{blue}{3.2} / 58.3 \textcolor{blue}{3.5} / 74.9\textcolor{blue}{4.6} & 20.5 \textcolor{blue}{22.3} / 18.2 \textcolor{blue}{20.1} / 22.9 \textcolor{blue}{23.1} / 33.2 \textcolor{blue}{31.4} & 31.8 \textcolor{blue}{17.8} / 28.0 \textcolor{blue}{11.2} / 35.2 \textcolor{blue}{12.4} / 50.1 \textcolor{blue}{19.8}\\ 
\midrule
& YCB-U  & ScanNet-U  & COCO-U   \\ \midrule
              & AP / PQ / Pre / Rec  & AP / PQ / Pre / Rec & AP / PQ / Pre / Rec  \\
AIR \cite{Eslami2016} & 9.8 \textcolor{blue}{3.5} / 12.2 \textcolor{blue}{2.6} / 18.6 \textcolor{blue}{4.1} / 19.9 \textcolor{blue}{2.8} & 7.1 \textcolor{blue}{1.4} / 10.4 \textcolor{blue}{0.2} / 19.9 \textcolor{blue}{2.1} / 14.0 \textcolor{blue}{1.0} & 12.9 \textcolor{blue}{4.8} / 16.3 \textcolor{blue}{4.6} / 28.6 \textcolor{blue}{1.8} / 24.3 \textcolor{blue}{9.3} \\ 
MONet \cite{Burgess2019} & 3.7 \textcolor{blue}{0.5} / 8.3 \textcolor{blue}{1.2} / 12.2 \textcolor{blue}{1.4} / 13.2 \textcolor{blue}{1.4} & 25.4 \textcolor{blue}{8.3} / 23.6 \textcolor{blue}{7.4} / 29.2 \textcolor{blue}{8.9} / 37.2 \textcolor{blue}{12.5} & 14.5 \textcolor{blue}{0.6} / 16.5 \textcolor{blue}{0.8} / 27.9 \textcolor{blue}{1.4} / 20.9 \textcolor{blue}{0.8} \\ 
IODINE \cite{Greff2019} & 2.5 \textcolor{blue}{1.4} / 4.9 \textcolor{blue}{2.0} / 7.6 \textcolor{blue}{2.9} / 9.3 \textcolor{blue}{3.6} & 16.9 \textcolor{blue}{0.8} / 17.2 \textcolor{blue}{0.0} / 23.8 \textcolor{blue}{0.8} / 30.2 \textcolor{blue}{0.1} & 5.5 \textcolor{blue}{1.0} / 7.7 \textcolor{blue}{0.0} / 11.7 \textcolor{blue}{0.6} / 13.5 \textcolor{blue}{0.3} \\ 
SlotAtt \cite{Locatello2020} & 28.9 \textcolor{blue}{2.0} / 21.9 \textcolor{blue}{1.2} / 30.7 \textcolor{blue}{3.1} / 37.3 \textcolor{blue}{2.4} & 9.4 \textcolor{blue}{4.8} / 10.4 \textcolor{blue}{3.8} / 15.4 \textcolor{blue}{4.6} / 18.1 \textcolor{blue}{7.2} & 4.4 \textcolor{blue}{1.1} / 7.9 \textcolor{blue}{0.3} / 11.9 \textcolor{blue}{0.9} / 14.7 \textcolor{blue}{0.7} \\ 
\midrule
& YCB-T+U  & ScanNet-T+U  & COCO-T+U   \\ \midrule
              & AP / PQ / Pre / Rec  & AP / PQ / Pre / Rec & AP / PQ / Pre / Rec  \\ 
AIR \cite{Eslami2016} & 18.9 \textcolor{blue}{13.7} / 21.0 \textcolor{blue}{9.9} / 32.5 \textcolor{blue}{15.7} / 32.6 \textcolor{blue}{13.2} & 12.3 \textcolor{blue}{5.4} / 16.8 \textcolor{blue}{5.6} / 37.4 \textcolor{blue}{20.8} / 20.4 \textcolor{blue}{1.1} & 25.3 \textcolor{blue}{13.0} / 28.3 \textcolor{blue}{9.0} / 49.4 \textcolor{blue}{13.2} / 38.2 \textcolor{blue}{12.4} \\ 
MONet \cite{Burgess2019} & 89.1 \textcolor{blue}{0.8} / 84.7 \textcolor{blue}{1.5} / 91.7 \textcolor{blue}{4.1} / 89.3 \textcolor{blue}{0.8} & 87.0 \textcolor{blue}{8.9} / 77.5 \textcolor{blue}{10.3} / 80.5 \textcolor{blue}{11.1} / 87.4 \textcolor{blue}{8.7} & 88.2 \textcolor{blue}{2.9} / 76.2 \textcolor{blue}{0.1} / 79.9 \textcolor{blue}{3.7} / 89.0 \textcolor{blue}{2.9} \\ 
IODINE \cite{Greff2019} & 35.3 \textcolor{blue}{1.2} / 26.4 \textcolor{blue}{0.3} / 34.9 \textcolor{blue}{0.6} / 42.8 \textcolor{blue}{0.2} & 27.4 \textcolor{blue}{0.8} / 25.0 \textcolor{blue}{0.8} / 32.1 \textcolor{blue}{0.5} / 41.8 \textcolor{blue}{1.8} & 53.3 \textcolor{blue}{16.7} / 35.6 \textcolor{blue}{8.3} / 45.0 \textcolor{blue}{9.7} / 59.8 \textcolor{blue}{13.0} \\ 
SlotAtt \cite{Locatello2020} & 76.8 \textcolor{blue}{1.2} / 57.1 \textcolor{blue}{3.8} / 72.4 \textcolor{blue}{3.4} / 77.7 \textcolor{blue}{1.8} & 47.3 \textcolor{blue}{11.2} / 33.8 \textcolor{blue}{9.2} / 42.6 \textcolor{blue}{9.1} / 57.3 \textcolor{blue}{8.4} & 34.5 \textcolor{blue}{17.8} / 25.3 \textcolor{blue}{7.6} / 33.7 \textcolor{blue}{8.3} / 43.0 \textcolor{blue}{14.1}\\ \bottomrule
               
  \end{tabular}
  \vspace{-0.4cm}
\end{table}

\subsubsection{Ablations on Object- and Scene-Level Factor Joint Ablations - Part 1}\label{sec:joint_ablation_1}
\paragraph{Ablated Datasets}

\begin{itemize}[leftmargin=*]
\setlength{\itemsep}{1pt}
\setlength{\parsep}{1pt}
\setlength{\parskip}{1pt}
    \item \textit{Ablation of Object Color Gradient, Object Shape Concavity and Inter-object Color Similarity}: In the three ablated datasets: YCB-C+S+T / ScanNet-C+S+T / COCO-C+S+T, for each image, we first find the smallest convex hull \cite{Eddins2011} for each object. Then each convex hull is filled with a distinctive texture as shown in Figure \ref{fig:dtd_texture}. Lastly, we calculate the average color of the  distinctive texture in each convex hull and change each pixel inside to be that color.    
    \item \textit{Ablation of Object Color Gradient, Object Shape Concavity and Inter-object Shape Variation}: In the three ablated datasets: YCB-C+S+U / ScanNet-C+S+U / COCO-C+S+U, 
    we use the (C+S) datasets created before and find the uniform scale of the object convex hull.
    \item \textit{Ablation of all four factors}: In the three ablated datasets: YCB-C+S+T+U / ScanNet-C+S+T+U / COCO-C+S+T+U, each image is ablated with the four factors together.
\end{itemize}

\begin{figure}[th]
    \setlength{\abovecaptionskip}{ 0 pt}
    \setlength{\belowcaptionskip}{ 0 pt}
    \centering
      \includegraphics[width=0.46\linewidth]{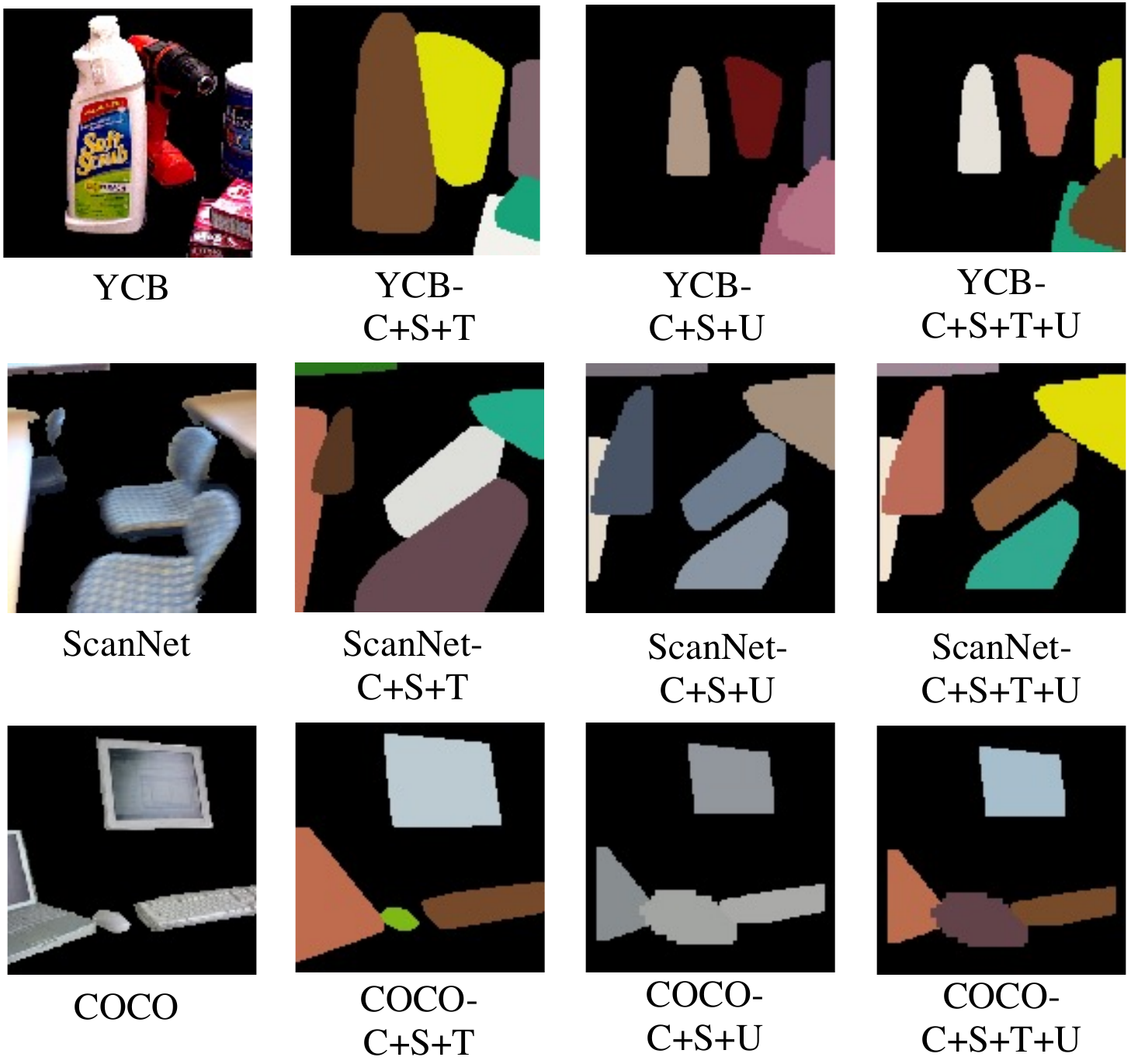}
      \vspace{0.1cm}
    \caption{Example images of datasets ablated with both object- and scene-level factors.}
    \label{fig:dataset_S_C_+}
    \vspace{-0.2cm}
\end{figure}

Figure \ref{fig:dataset_S_C_+} shows example images from the three real-world datasets with different types of joint object- and scene-level factor ablations.

\begin{figure}[ht]
    \setlength{\abovecaptionskip}{ 4 pt}
    \setlength{\belowcaptionskip}{ 0 pt}
    \centering
      \includegraphics[width=1\linewidth]{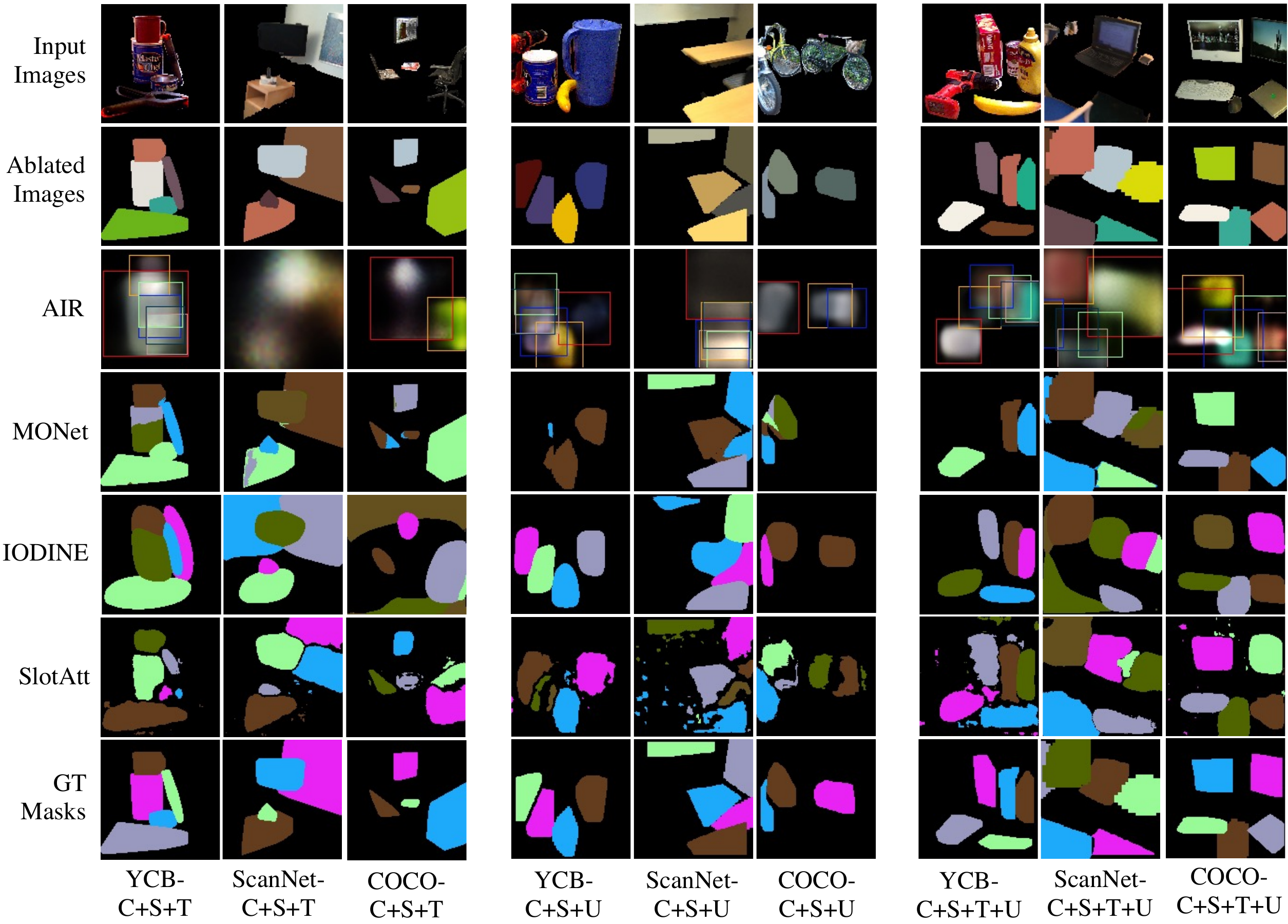}
    \caption{Qualitative results on the datasets ablated with both object- and scene-level factors.}
    \label{fig:app_hybrid_ablation_1}
\end{figure}

\paragraph{Qualitative and Quantitative Results}
As shown in Figure \ref{fig:app_hybrid_ablation_1} and Table \ref{tab:app_hybrid_ablation_1}, all methods have achieved a significant improvement in object segmentation on the datasets ablated with joint object- and scene-level factors. The datasets with all four factors ablated can lead to impressive performance similar to the synthetic datasets. This shows that the distribution gaps of the objectness biases represented by the four factors between the synthetic and real-world datasets lead to the failure of existing unsupervised models.

\setlength{\abovecaptionskip}{-6 pt}
\setlength{\belowcaptionskip}{4 pt}
\begin{table}[h]
\tiny
  \caption{Quantitative results on the datasets ablated with both object- and scene-level factors. Standard deviations of performance are calculated over 3 runs (marked with \textcolor{blue}{blue}). }
  \tabcolsep= 0.06cm 
  \label{tab:app_hybrid_ablation_1}
  \centering
  \begin{tabular}{r c|c|c} \toprule
      & YCB-C+S+T & ScanNet-C+S+T & COCO-C+S+T  \\ \midrule
              & AP / PQ / Pre / Rec  & AP / PQ / Pre / Rec & AP / PQ / Pre / Rec \\ 
AIR \cite{Eslami2016} & 6.3 \textcolor{blue}{3.7} / 10.7 \textcolor{blue}{3.3} / 24.0 \textcolor{blue}{13.2} / 13.1 \textcolor{blue}{0.5} & 1.9 \textcolor{blue}{3.9} / 6.4 \textcolor{blue}{5.2} / 27.6 \textcolor{blue}{10.7} / 6.3 \textcolor{blue}{13.9} & 10.8 \textcolor{blue}{5.6} / 18.6 \textcolor{blue}{7.0} / 45.1 \textcolor{blue}{14.4} / 20.3 \textcolor{blue}{6.9} \\ 
MONet \cite{Burgess2019} & 67.7 \textcolor{blue}{4.2} / 55.0 \textcolor{blue}{8.5} / 62.7 \textcolor{blue}{19.8} / 75.4 \textcolor{blue}{10.2} & 76.1 \textcolor{blue}{7.4} / 61.7 \textcolor{blue}{2.6} / 64.4 \textcolor{blue}{2.1} / 77.1 \textcolor{blue}{2.8} & 67.1 \textcolor{blue}{15.3} / 56.4 \textcolor{blue}{10.1} / 62.2 \textcolor{blue}{6.2} / 73.9 \textcolor{blue}{17.3} \\ 
IODINE \cite{Greff2019} & 82.6 \textcolor{blue}{2.6} / 63.8 \textcolor{blue}{0.9} / 75.5 \textcolor{blue}{5.1} / 86.0 \textcolor{blue}{1.6} & 69.0 \textcolor{blue}{14.9} / 53.5 \textcolor{blue}{14.1} / 60.7 \textcolor{blue}{14.5} / 77.3 \textcolor{blue}{9.4} & 77.6 \textcolor{blue}{7.8} / 59.8 \textcolor{blue}{6.8} / 72.5 \textcolor{blue}{1.8} / 83.0 \textcolor{blue}{7.5} \\ 
SlotAtt \cite{Locatello2020} & 65.9 \textcolor{blue}{23.2} / 51.6 \textcolor{blue}{18.5} / 62.2 \textcolor{blue}{19.8} / 72.9 \textcolor{blue}{17.6} & 60.2 \textcolor{blue}{8.7} / 45.5 \textcolor{blue}{8.1} / 52.9 \textcolor{blue}{9.2} / 68.3 \textcolor{blue}{7.3} & 51.8 \textcolor{blue}{11.0} / 41.5 \textcolor{blue}{7.7} / 48.6 \textcolor{blue}{7.5} / 62.9 \textcolor{blue}{8.5} \\ 
\midrule
& YCB-C+S+U  & ScanNet-C+S+U  & COCO-C+S+U   \\ \midrule
              & AP / PQ / Pre / Rec  & AP / PQ / Pre / Rec & AP / PQ / Pre / Rec  \\
AIR \cite{Eslami2016} & 12.5 \textcolor{blue}{6.4} / 17.7 \textcolor{blue}{7.0} / 27.5 \textcolor{blue}{11.2} / 29.6 \textcolor{blue}{10.5} & 8.2 \textcolor{blue}{1.5} / 11.1 \textcolor{blue}{0.5} / 20.4 \textcolor{blue}{0.8} / 15.0 \textcolor{blue}{1.3} & 23.7 \textcolor{blue}{9.8} / 27.5 \textcolor{blue}{8.0} / 51.3 \textcolor{blue}{9.6} / 32.8 \textcolor{blue}{8.3} \\ 
MONet \cite{Burgess2019} & 3.5 \textcolor{blue}{2.5} / 4.8 \textcolor{blue}{3.2} / 23.5 \textcolor{blue}{2.6} / 5.3 \textcolor{blue}{6.5} & 60.1 \textcolor{blue}{9,5} / 56.3 \textcolor{blue}{13.3} / 69.7 \textcolor{blue}{20.7} / 64.0 \textcolor{blue}{6.2} & 35.6 \textcolor{blue}{1.4} / 33.0 \textcolor{blue}{3.8} / 43.4 \textcolor{blue}{16.6} / 38.7 \textcolor{blue}{0.1} \\ 
IODINE \cite{Greff2019} & 65.3 \textcolor{blue}{9.2} / 55.9 \textcolor{blue}{10.4} / 73.9 \textcolor{blue}{4.6} / 67.7 \textcolor{blue}{8.4} & 64.6 \textcolor{blue}{3.2} / 52.1 \textcolor{blue}{5.1} / 61.3 \textcolor{blue}{7.6} / 68.2 \textcolor{blue}{2.2} & 64.8 \textcolor{blue}{7.3} / 53.4 \textcolor{blue}{6.5} / 65.8 \textcolor{blue}{8.5} / 68.2 \textcolor{blue}{2.5} \\ 
SlotAtt \cite{Locatello2020} & 58.1 \textcolor{blue}{3.7} / 39.1 \textcolor{blue}{2.8} / 52.4 \textcolor{blue}{2.3} / 62.0 \textcolor{blue}{3.6} & 50.5 \textcolor{blue}{2.9} / 38.2 \textcolor{blue}{3.6} / 51.8 \textcolor{blue}{6.9} / 55.0 \textcolor{blue}{1.3} & 20.2 \textcolor{blue}{5.2} / 18.1 \textcolor{blue}{0.3} / 26.9 \textcolor{blue}{0.1} / 33.4 \textcolor{blue}{0.6}\\ 
\midrule
& YCB-C+S+T+U  & ScanNet-C+S+T+U  & COCO-C+S+T+U   \\ \midrule
              & AP / PQ / Pre / Rec  & AP / PQ / Pre / Rec & AP / PQ / Pre / Rec  \\ 
AIR \cite{Eslami2016} & 24.8 \textcolor{blue}{20.1} / 23.2 \textcolor{blue}{11.9} / 33.7 \textcolor{blue}{16.8} / 39.1 \textcolor{blue}{20.1} & 10.0 \textcolor{blue}{1.8} / 13.5 \textcolor{blue}{1.8}/ 24.0 \textcolor{blue}{1.4} / 18.2 \textcolor{blue}{7.4} & 36.4 \textcolor{blue}{19.3} / 39.4 \textcolor{blue}{14.5} / 63.1 \textcolor{blue}{16.9} / 49.2 \textcolor{blue}{17.9} \\ 
MONet \cite{Burgess2019} & 70.8 \textcolor{blue}{9.5} / 79.2 \textcolor{blue}{13.0} / 96.8 \textcolor{blue}{9.4} / 70.8 \textcolor{blue}{8.5} & 79.2 \textcolor{blue}{3.4} / 65.0 \textcolor{blue}{3.6} / 69.1 \textcolor{blue}{4.2} / 82.4 \textcolor{blue}{2.2} & 76.2 \textcolor{blue}{3.0} / 75.9 \textcolor{blue}{3.8} / 90.9 \textcolor{blue}{6.4} / 76.5 \textcolor{blue}{3.3} \\ 
IODINE \cite{Greff2019} & 72.4 \textcolor{blue}{4.8} / 54.2 \textcolor{blue}{6.7} / 65.1 \textcolor{blue}{5.8} / 74.3 \textcolor{blue}{5.8} & 76.1 \textcolor{blue}{8.7} / 56.5 \textcolor{blue}{9.2} / 66.9 \textcolor{blue}{7.0} / 80.3 \textcolor{blue}{6.2} & 81.4 \textcolor{blue}{3.4} / 59.3 \textcolor{blue}{5.1} / 70.5 \textcolor{blue}{7.7} / 84.3 \textcolor{blue}{6.7} \\ 
SlotAtt \cite{Locatello2020} & 92.0 \textcolor{blue}{2.0} / 65.5 \textcolor{blue}{6.1} / 84.4 \textcolor{blue}{6.5} / 92.5 \textcolor{blue}{1.7} & 62.7 \textcolor{blue}{25.4} / 42.6 \textcolor{blue}{28.4} / 50.5 \textcolor{blue}{34.3} / 69.4 \textcolor{blue}{19.0} & 83.7 \textcolor{blue}{4.7} / 60.7 \textcolor{blue}{7.2} / 76.4 \textcolor{blue}{4.6} / 84.0 \textcolor{blue}{5.0}\\ \bottomrule
               
  \end{tabular}
\end{table}

\subsubsection{Ablations on Object- and Scene-Level Factor Joint Ablations - Part 2}\label{sec:joint_ablation_2}
\begin{figure}[ht]
    \setlength{\abovecaptionskip}{ 0 pt}
    \setlength{\belowcaptionskip}{ 0 pt}
    \centering
    \vspace{-0.2cm}
      \includegraphics[width=0.9\linewidth]{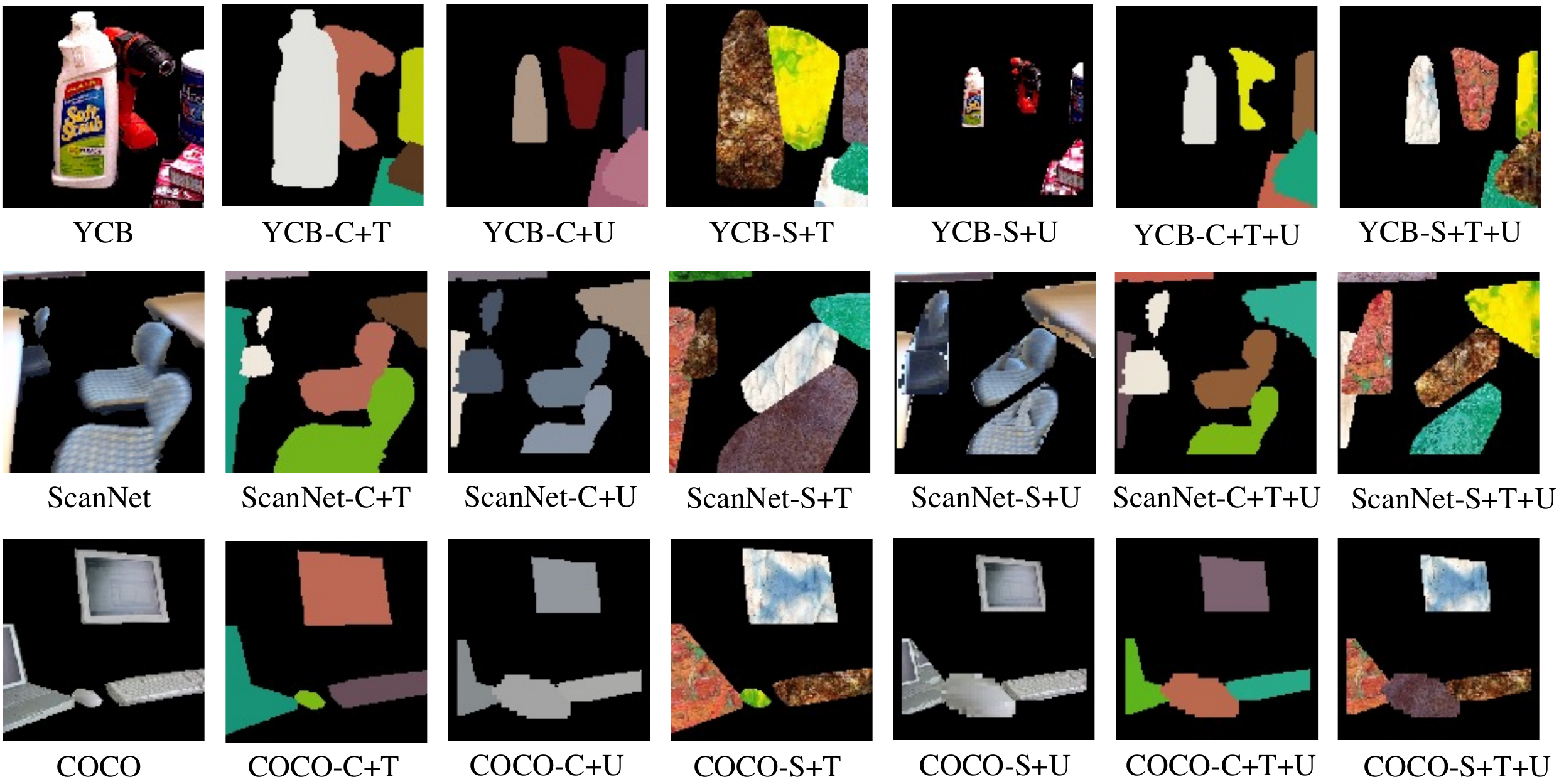}
      \vspace{0.1cm}
    \caption{Example images of additional datasets ablated with both object- and scene-level factors.}
    \label{fig:dataset_additional}
    \vspace{-0.2cm}
\end{figure}

In addition to the experiments in Section \ref{sec:joint_ablation_1}, we generate additional 6 groups of datasets ablated with different combinations of both object- and scene-level factors as detailed in Section \ref{sec:add_exp_joint_ablation}, and conduct extra experiments shown in Section \ref{sec:add_res_joint_ablation}. Figure \ref{fig:dataset_additional} shows example images of the additional ablated datasets. 

\paragraph{Ablated Datasets} \label{sec:add_exp_joint_ablation}
\begin{itemize}[leftmargin=*]
\setlength{\itemsep}{1pt}
\setlength{\parsep}{1pt}
\setlength{\parskip}{1pt}
    \item \textit{Ablation of Object Color Gradient and Inter-object Color Similarity}: In each image of the three real-world datasets, we replace the object color by averaging all pixels of a distinctive texture, and keep the original shape unchanged, getting three ablated datasets: YCB-C+T / ScanNet-C+T / COCO-C+T.   
    \item \textit{Ablation of Object Color Gradient and Inter-object Shape Variation}: In each image of the three real-world datasets, we replace the object color by averaging its own texture, and then apply size normalization on the original shape of objects, getting three ablated datasets: YCB-C+U / ScanNet-C+U / COCO-C+U.  
    \item \textit{Ablation of Object Color Gradient, Inter-object Color Similarity and Inter-object Shape Variation}: In each image of the three real-world datasets, we replace the object color by averaging all pixels of a distinctive texture, and then apply size normalization on the original shape of objects. The ablated datasets are denoted as: YCB-C+T+U / ScanNet-C+T+U / COCO-C+T+U.   
    
    \item \textit{Ablation of Object Shape Concavity and Inter-object Color Similarity}: In each image of the three real-world datasets, we replace the object color by a distinctive texture, and modify the object shape as a convex hull, getting three ablated datasets: YCB-S+T / ScanNet-S+T / COCO-S+T.   
    \item \textit{Ablation of  Object Shape Concavity and Inter-object Shape Variation}: In each image of the three real-world datasets, we keep the texture of object unchanged, and modify the object shape as a convex hull followed by size normalization, getting three ablated datasetss: YCB-S+U / ScanNet-S+U / COCO-S+U.
    \item \textit{Ablation of Object Shape Concavity, Inter-object Color Similarity and Inter-object Shape Variation}: In each image of the three real-world datasets, we replace the object color by a distinctive texture, and modify the object shape as a convex hull followed by size normalization. The ablated datasets are denoted as: YCB-S+T+U / ScanNet-S+T+U / COCO-S+T+U. 
\end{itemize}

\setlength{\abovecaptionskip}{-6 pt}
\setlength{\belowcaptionskip}{4 pt}
\begin{table}[h]
\tiny
  \caption{Quantitative results on additional datasets ablated with both object- and scene-level factors. Standard deviations of performance are calculated over 3 runs (marked with \textcolor{blue}{blue}).}
  \tabcolsep= 0.06cm 
  \label{tab:app_hybrid_ablation_additional}
  \centering
  \begin{tabular}{r c|c|c} \toprule

& YCB-C+T & ScanNet-C+T & COCO-C+T  \\ \midrule
              & AP / PQ / Pre / Rec  & AP / PQ / Pre / Rec & AP / PQ / Pre / Rec \\ 
AIR \cite{Eslami2016} & 8.0 \textcolor{blue}{6.1} / 13.2 \textcolor{blue}{7.3} / 26.6 \textcolor{blue}{17.6} / 17.9 \textcolor{blue}{7.3} & 2.8 \textcolor{blue}{4.7} / 6.7 \textcolor{blue}{6.5} / 14.8 \textcolor{blue}{4.0} / 8.1 \textcolor{blue}{14.8} & 12.4 \textcolor{blue}{9.4} / 30.2 \textcolor{blue}{23.0} / 40.1 \textcolor{blue}{25.6} / 24.2 \textcolor{blue}{13.8} \\ 
MONet \cite{Burgess2019} & 82.9 \textcolor{blue}{9.6} / 66.7 \textcolor{blue}{0.5} / 73.0 \textcolor{blue}{1.2} / 87.8 \textcolor{blue}{12.2} & 36.5 \textcolor{blue}{28.4} / 33.5 \textcolor{blue}{21.8} / 40.4 \textcolor{blue}{24.1} / 50.3 \textcolor{blue}{19.7} & 66.1 \textcolor{blue}{5.6} / 53.2 \textcolor{blue}{3.7} / 56.2 \textcolor{blue}{2.3} / 74.4 \textcolor{blue}{6.4} \\ 
IODINE \cite{Greff2019} & 78.9 \textcolor{blue}{2.6} / 59.7 \textcolor{blue}{2.1} / 71.5 \textcolor{blue}{0.6} / 83.6 \textcolor{blue}{2.7} & 65.1 \textcolor{blue}{2.9} / 51.1 \textcolor{blue}{1.0} / 62.4 \textcolor{blue}{2.9} / 74.2 \textcolor{blue}{0.5}& 55.1 \textcolor{blue}{10.4} / 42.2 \textcolor{blue}{7.5} / 56.3 \textcolor{blue}{6.4} / 66.2 \textcolor{blue}{9.2} \\ 
SlotAtt \cite{Locatello2020} & 58.7 \textcolor{blue}{12.1} / 43.2 \textcolor{blue}{5.0} / 57.8 \textcolor{blue}{9.4} / 71.0 \textcolor{blue}{9.4} & 29.2 \textcolor{blue}{2.9} / 27.6 \textcolor{blue}{1.3} / 34.4 \textcolor{blue}{1.4} / 49.2 \textcolor{blue}{1.2} & 22.1 \textcolor{blue}{8.6} / 19.6 \textcolor{blue}{4.1} / 25.2 \textcolor{blue}{4.1} / 35.8 \textcolor{blue}{8.1}\\
\midrule

& YCB-C+U  & ScanNet-C+U  & COCO-C+U   \\ \midrule
              & AP / PQ / Pre / Rec  & AP / PQ / Pre / Rec & AP / PQ / Pre / Rec  \\
AIR \cite{Eslami2016} & 11.4 \textcolor{blue}{4.9} / 16.4 \textcolor{blue}{5.4} / 25.8 \textcolor{blue}{8.7} / 25.1 \textcolor{blue}{6.5} & 5.4 \textcolor{blue}{0.5} / 12.3 \textcolor{blue}{2.6} / 35.1 \textcolor{blue}{15.3} / 12.6 \textcolor{blue}{0.4} & 20.5 \textcolor{blue}{10.8} / 24.9 \textcolor{blue}{7.1} / 47.7 \textcolor{blue}{10.2} / 30.5 \textcolor{blue}{8.8}\\ 
MONet \cite{Burgess2019} & 49.1 \textcolor{blue}{3.5} / 44.6 \textcolor{blue}{4.1} / 60.6 \textcolor{blue}{3.2} / 52.5 \textcolor{blue}{3.3} & 31.9 \textcolor{blue}{5.7} / 33.7 \textcolor{blue}{5.7} / 46.8 \textcolor{blue}{5.4} / 35.7 \textcolor{blue}{8.7} & 32.6 \textcolor{blue}{11.1} / 28.5 \textcolor{blue}{12.1} / 34.9 \textcolor{blue}{14.2} / 39.0 \textcolor{blue}{18.5}\\ 
IODINE \cite{Greff2019} & 66.3 \textcolor{blue}{0.1} / 54.6 \textcolor{blue}{0.0} / 71.3 \textcolor{blue}{4.1} / 70.0 \textcolor{blue}{0.2} & 34.8 \textcolor{blue}{20.9} / 30.0 \textcolor{blue}{14.7} / 36.4 \textcolor{blue}{20.9} / 47.2 \textcolor{blue}{14.0} & 44.4 \textcolor{blue}{8.9} / 32.6 \textcolor{blue}{7.9} / 41.7 \textcolor{blue}{12.3} / 51.1 \textcolor{blue}{7.6} \\ 
SlotAtt \cite{Locatello2020} & 45.6 \textcolor{blue}{7.9} / 31.2 \textcolor{blue}{4.2} / 41.9 \textcolor{blue}{7.7} / 51.3 \textcolor{blue}{6.4} & 30.4 \textcolor{blue}{8.9} / 24.2 \textcolor{blue}{6.2} / 33.8 \textcolor{blue}{9.5} / 40.6 \textcolor{blue}{11.0} & 12.7 \textcolor{blue}{7.7} / 11.3 \textcolor{blue}{3.7} / 15.8 \textcolor{blue}{5.0} / 23.5 \textcolor{blue}{6.8} \\ 
\midrule

& YCB-S+T & ScanNet-S+T & COCO-S+T  \\ \midrule
              & AP / PQ / Pre / Rec  & AP / PQ / Pre / Rec & AP / PQ / Pre / Rec \\ 
AIR \cite{Eslami2016} & 9.3 \textcolor{blue}{7.8} / 13.4 \textcolor{blue}{8.4} / 21.9 \textcolor{blue}{14.3} / 21.2 \textcolor{blue}{12.3} & 3.2 \textcolor{blue}{0.5} / 7.7 \textcolor{blue}{0.6} / 18.5 \textcolor{blue}{6.0} / 8.9 \textcolor{blue}{6.0} & 10.5 \textcolor{blue}{5.4} / 18.2 \textcolor{blue}{7.2} / 4.3 \textcolor{blue}{23.2} / 20.3 \textcolor{blue}{7.3} \\ 
MONet \cite{Burgess2019} & 86.7 \textcolor{blue}{2.2} / 78.9 \textcolor{blue}{1.6} / 81.9 \textcolor{blue}{1.8} / 87.0 \textcolor{blue}{2.2} & 83.2 \textcolor{blue}{3.7} / 77.8 \textcolor{blue}{13.0} / 86.0 \textcolor{blue}{20.0} / 83.8 \textcolor{blue}{2.9} & 80.2 \textcolor{blue}{0.0} / 68.5 \textcolor{blue}{7.2} / 73.2 \textcolor{blue}{11.7} / 82.3 \textcolor{blue}{1.0} \\ 
IODINE \cite{Greff2019} & 41.9 \textcolor{blue}{17.9} / 33.2 \textcolor{blue}{11.4} / 41.5 \textcolor{blue}{12.0} / 51.3 \textcolor{blue}{14.7} & 54.9 \textcolor{blue}{23.1} / 44.4 \textcolor{blue}{16.2} / 52.0 \textcolor{blue}{16.7} / 68.0 \textcolor{blue}{21.6} & 44.1 \textcolor{blue}{1.1} / 34.3 \textcolor{blue}{0.7} / 41.0 \textcolor{blue}{0.3} / 56.8 \textcolor{blue}{0.6} \\ 
SlotAtt \cite{Locatello2020} & 77.3 \textcolor{blue}{7.8} / 60.6 \textcolor{blue}{1.5} / 75.0 \textcolor{blue}{2.3} / 69.2 \textcolor{blue}{19.7} & 24.9 \textcolor{blue}{46.1} / 21.2 \textcolor{blue}{35.0} / 25.5 \textcolor{blue}{38.3} / 37.9 \textcolor{blue}{43.2} & 67.4 \textcolor{blue}{3.1} / 50.2 \textcolor{blue}{1.6} / 59.2 \textcolor{blue}{1.6} / 76.4 \textcolor{blue}{0.3} \\ 
\midrule

& YCB-S+U  & ScanNet-S+U  & COCO-S+U   \\ \midrule
              & AP / PQ / Pre / Rec  & AP / PQ / Pre / Rec & AP / PQ / Pre / Rec  \\
AIR \cite{Eslami2016} & 0.8 \textcolor{blue}{5.4} / 2.9 \textcolor{blue}{7.1} / 5.0 \textcolor{blue}{10.0} / 5.2 \textcolor{blue}{12.7} & 6.3 \textcolor{blue}{0.2} / 13.1 \textcolor{blue}{3.3} / 32.4 \textcolor{blue}{16.3} / 14.3 \textcolor{blue}{1.2} & 20.0 \textcolor{blue}{9.5} / 25.8 \textcolor{blue}{10.0} / 48.3 \textcolor{blue}{15.6} / 31.6 \textcolor{blue}{10.9} \\ 
MONet \cite{Burgess2019} & 5.5 \textcolor{blue}{1.2} / 9.8 \textcolor{blue}{0.4} / 14.1 \textcolor{blue}{0.8} / 17.0 \textcolor{blue}{0.2} & 36.9 \textcolor{blue}{2.9} / 31.4 \textcolor{blue}{2.6} / 38.2 \textcolor{blue}{3.6} / 50.1 \textcolor{blue}{0.1} & 26.1 \textcolor{blue}{3.5} / 23.8 \textcolor{blue}{1.8} / 29.6 \textcolor{blue}{1.6} / 41.2 \textcolor{blue}{10.0} \\ 
IODINE \cite{Greff2019} & 2.8 \textcolor{blue}{2.4} / 4.6 \textcolor{blue}{2.0} / 7.2 \textcolor{blue}{2.9} / 8.9 \textcolor{blue}{3.6} & 17.3 \textcolor{blue}{2.4} / 17.8 \textcolor{blue}{1.4} / 24.1 \textcolor{blue}{1.8} / 31.6 \textcolor{blue}{2.1} & 5.7 \textcolor{blue}{0.5} / 8.7 \textcolor{blue}{1.7} / 1.3 \textcolor{blue}{13.7} / 16.3 \textcolor{blue}{2.4} \\ 
SlotAtt \cite{Locatello2020} & 36.2 \textcolor{blue}{3.3} / 23.8 \textcolor{blue}{4.1} / 33.6 \textcolor{blue}{5.5} / 45.6 \textcolor{blue}{1.7} & 21.1 \textcolor{blue}{0.9} / 18.9 \textcolor{blue}{0.2} / 26.2 \textcolor{blue}{0.1} / 32.5 \textcolor{blue}{2.0} & 12.7 \textcolor{blue}{7.1} / 12.1 \textcolor{blue}{2.3} / 16.4 \textcolor{blue}{2.1} / 24.7 \textcolor{blue}{5.2} \\ 
\midrule
      
& YCB-C+T+U & ScanNet-C+T+U & COCO-C+T+U  \\ \midrule
              & AP / PQ / Pre / Rec  & AP / PQ / Pre / Rec & AP / PQ / Pre / Rec \\ 
AIR \cite{Eslami2016} & 18.2 \textcolor{blue}{12.0} / 20.5 \textcolor{blue}{7.9} / 32.5 \textcolor{blue}{13.5} / 30.9 \textcolor{blue}{9.0} & 13.5 \textcolor{blue}{6.6} / 20.2 \textcolor{blue}{9.1} / 42.4 \textcolor{blue}{25.5} / 24.1 \textcolor{blue}{5.6} & 24.2 \textcolor{blue}{11.1} / 28.3 \textcolor{blue}{8.5} / 50.5 \textcolor{blue}{13.1} / 36.4 \textcolor{blue}{9.7} \\ 
MONet \cite{Burgess2019} & 63.2 \textcolor{blue}{6.9} / 66.5 \textcolor{blue}{11.8} / 86.3 \textcolor{blue}{13.0} / 64.4 \textcolor{blue}{5.7} & 74.6 \textcolor{blue}{5.6} / 62.1 \textcolor{blue}{6.6} / 67.8 \textcolor{blue}{7.2} / 79.2 \textcolor{blue}{3.2} & 73.0 \textcolor{blue}{3.3} / 75.7 \textcolor{blue}{10.3} / 94.2 \textcolor{blue}{23.1} / 73.1 \textcolor{blue}{7.5} \\ 
IODINE \cite{Greff2019} & 54.9 \textcolor{blue}{23.1} / 38.7 \textcolor{blue}{18.2} / 58.0 \textcolor{blue}{12.0} / 59.4 \textcolor{blue}{20.8} & 65.5 \textcolor{blue}{3.8} / 48.0 \textcolor{blue}{3.2} / 59.8 \textcolor{blue}{5.9} / 70.7 \textcolor{blue}{2.9} & 47.9 \textcolor{blue}{11.7} / 35.5 \textcolor{blue}{7.4} / 53.8 \textcolor{blue}{2.1} / 53.4 \textcolor{blue}{14.2} \\ 
SlotAtt \cite{Locatello2020} & 73.8 \textcolor{blue}{4.6} / 53.0 \textcolor{blue}{2.9} / 67.6 \textcolor{blue}{7.8} / 75.2 \textcolor{blue}{4.5} & 58.5 \textcolor{blue}{8.1} / 45.5 \textcolor{blue}{10.2} / 52.9 \textcolor{blue}{8.9} / 68.3 \textcolor{blue}{11.9} & 30.5 \textcolor{blue}{27.6} / 20.3 \textcolor{blue}{18.0} / 27.8 \textcolor{blue}{23.0} / 38.9 \textcolor{blue}{23.3} \\ 
\midrule

& YCB-S+T+U  & ScanNet-S+T+U  & COCO-S+T+U   \\ \midrule
              & AP / PQ / Pre / Rec  & AP / PQ / Pre / Rec & AP / PQ / Pre / Rec  \\
AIR \cite{Eslami2016} & 24.6 \textcolor{blue}{20.0} / 24.8 \textcolor{blue}{13.7} / 37.4 \textcolor{blue}{20.7} / 37.6 \textcolor{blue}{18.6} & 16.7 \textcolor{blue}{10.1} / 23.9 \textcolor{blue}{12.0} / 47.8 \textcolor{blue}{30.2} / 28.5 \textcolor{blue}{7.8} & 29.4 \textcolor{blue}{16.5} / 34.6 \textcolor{blue}{18.4} / 58.3 \textcolor{blue}{35.7} / 42.2 \textcolor{blue}{13.4} \\ 
MONet \cite{Burgess2019} & 88.4 \textcolor{blue}{0.6} / 84.6 \textcolor{blue}{1.5} / 92.0 \textcolor{blue}{1.5} / 88.6 \textcolor{blue}{0.7} & 97.8 \textcolor{blue}{0.3} / 87.3 \textcolor{blue}{0.8} / 85.4 \textcolor{blue}{0.1} / 98.1 \textcolor{blue}{0.4} & 86.9 \textcolor{blue}{0.1} / 83.8 \textcolor{blue}{11.6} / 89.9 \textcolor{blue}{13.1} / 87.0 \textcolor{blue}{1.5} \\ 
IODINE \cite{Greff2019} & 42.3 \textcolor{blue}{15.0} / 32.1 \textcolor{blue}{9.7} / 41.0 \textcolor{blue}{11.5} / 52.0 \textcolor{blue}{15.7} & 39.1 \textcolor{blue}{4.7} / 29.9 \textcolor{blue}{1.8} / 41.8 \textcolor{blue}{7.2} / 47.9 \textcolor{blue}{2.4} & 64.3 \textcolor{blue}{9.7} / 45.1 \textcolor{blue}{6.9} / 51.4 \textcolor{blue}{5.6} / 72.0 \textcolor{blue}{11.7} \\ 
SlotAtt \cite{Locatello2020} & 88.6 \textcolor{blue}{6.5} / 67.0 \textcolor{blue}{9.6} / 80.9 \textcolor{blue}{13.0} / 89.6 \textcolor{blue}{5.6} & 82.6 \textcolor{blue}{8.5} / 61.5 \textcolor{blue}{10.1} / 73.2 \textcolor{blue}{10.2} / 83.9 \textcolor{blue}{7.7} & 84.2 \textcolor{blue}{0.9} / 59.6 \textcolor{blue}{6.8} / 74.7 \textcolor{blue}{3.8} / 85.3 \textcolor{blue}{0.6}\\  \bottomrule
               
  \end{tabular}
\end{table}

\begin{figure}[t]
    \setlength{\abovecaptionskip}{ 4 pt}
    \setlength{\belowcaptionskip}{ -6 pt}
    \centering
    \includegraphics[width=1\linewidth]{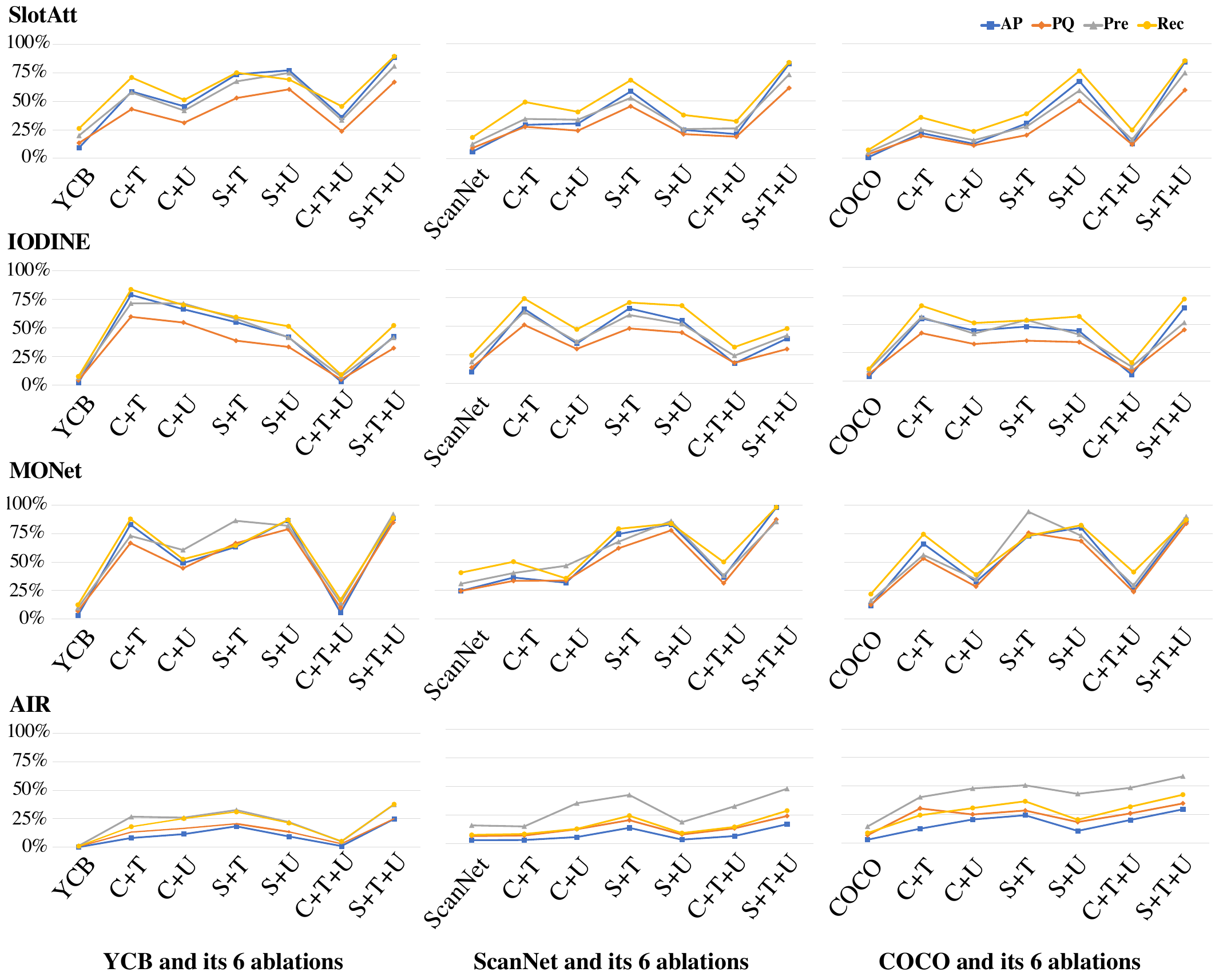} 
    \vspace{0.3cm}
    \caption{{Quantitative results of baselines on three real-world datasets and their variants in Sec} \ref{sec:joint_ablation_2}}
\label{fig:app_hybrid_ablation_stats}
\end{figure}

\begin{figure}[ht]
    \setlength{\abovecaptionskip}{ 0 pt}
    \setlength{\belowcaptionskip}{ 4 pt}
    \centering
    \vspace{-0.5cm}
      \includegraphics[width=0.75\linewidth]{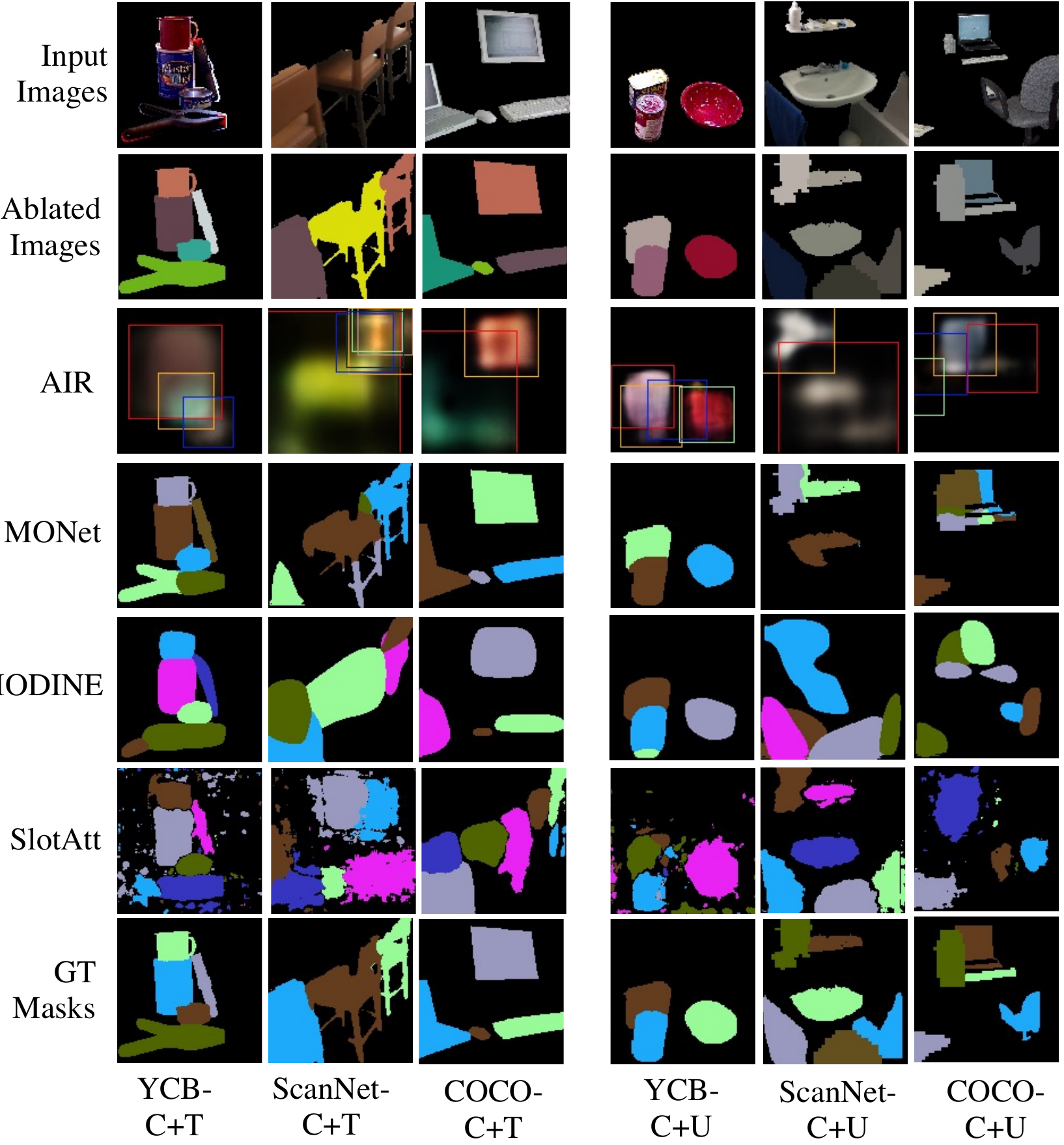}
    \caption{Qualitative results on the additional datasets ablated with object- and scene-level factors.}
    \label{fig:app_additional_experiments_vis_1}
\end{figure}

\begin{figure}[ht]
    \setlength{\abovecaptionskip}{ 0 pt}
    \setlength{\belowcaptionskip}{ 0 pt}
    \centering
    \vspace{-0.5cm}
      \includegraphics[width=0.75\linewidth]{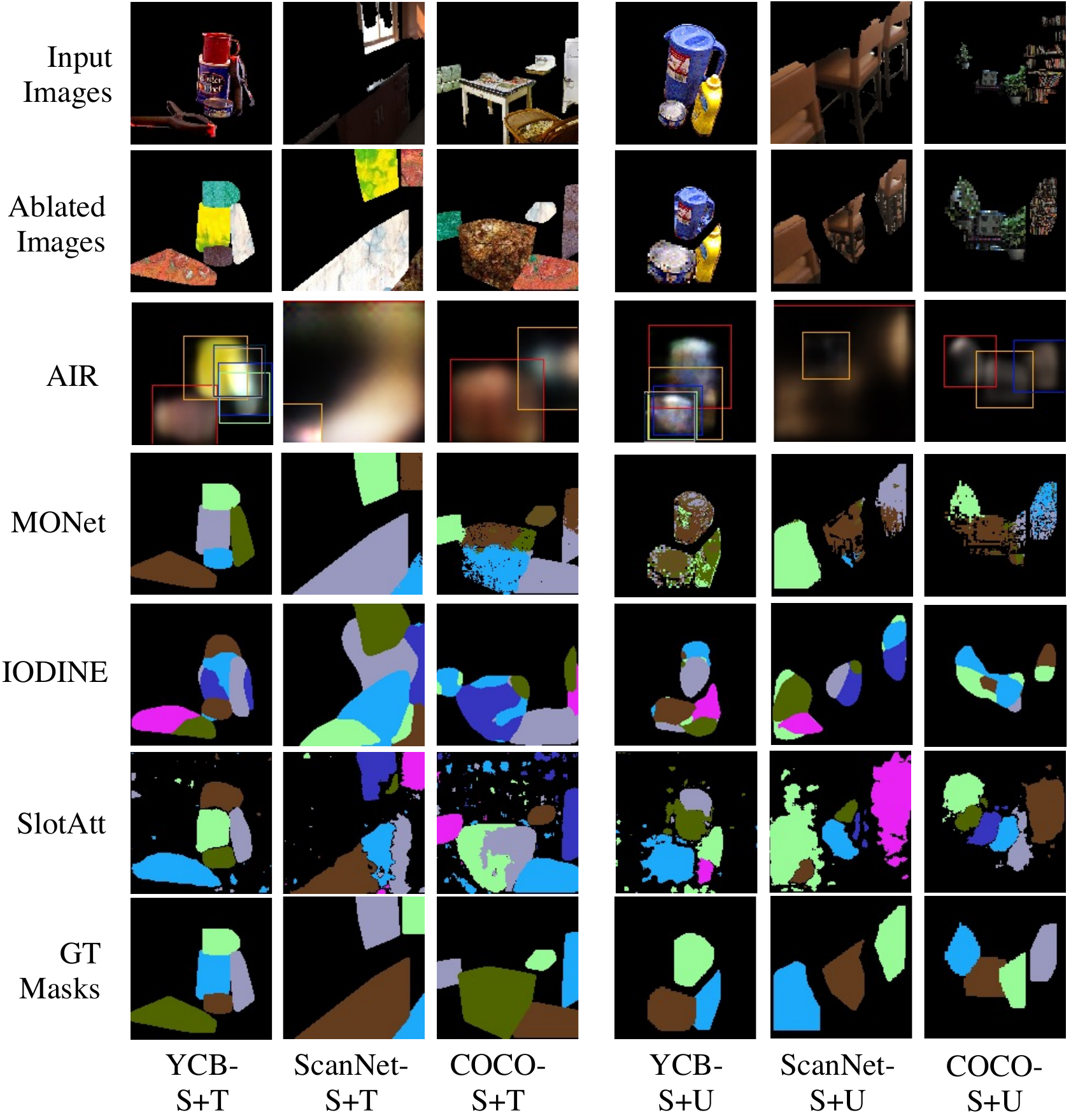}
    \caption{Qualitative results on the additional datasets ablated with object- and scene-level factors.}
    \label{fig:app_additional_experiments_vis_2}
\end{figure}

\begin{figure}[t]
    \setlength{\abovecaptionskip}{ 0 pt}
    \setlength{\belowcaptionskip}{ 0 pt}
    \centering
      \includegraphics[width=0.75\linewidth]{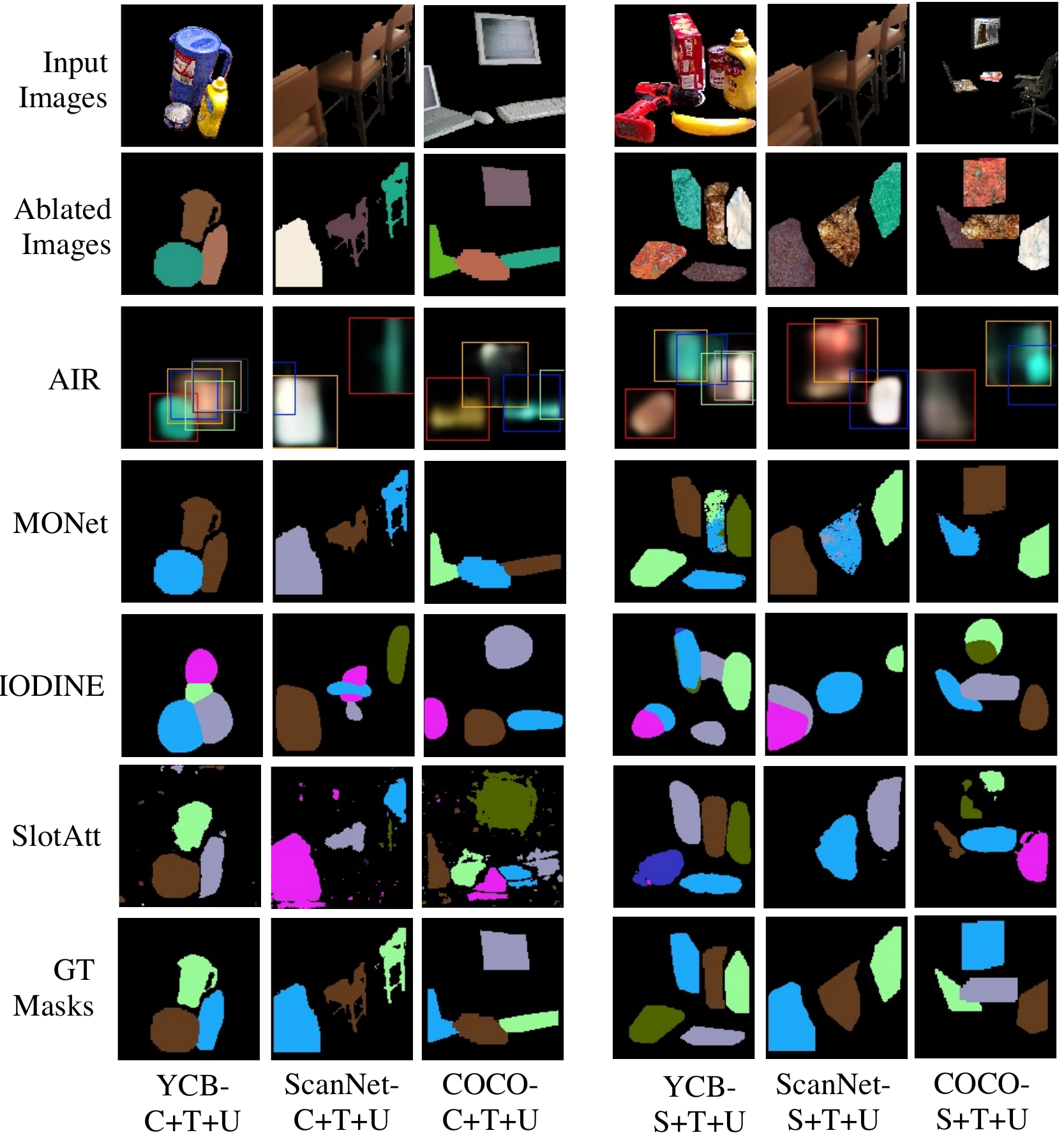}
    \caption{Qualitative results on the additional datasets ablated with object- and scene-level factors.}
    \label{fig:app_additional_experiments_vis_3}
    \vspace{11.cm}
\end{figure}

\paragraph{Qualitative and Quantitative Results} \label{sec:add_res_joint_ablation}
As shown in Table \ref{tab:app_hybrid_ablation_additional} and Figures \ref{fig:app_hybrid_ablation_stats}/\ref{fig:app_additional_experiments_vis_1}/\ref{fig:app_additional_experiments_vis_2}/\ref{fig:app_additional_experiments_vis_3}, all 6 additional combinations of object- and scene-level factors are explored, demonstrating consistent findings as our experiments in Section 4.4. Overall, all four methods show a high sensitivity to both object- and scene-level factors relating to appearance. This can be seen from the fact that for datasets without ablations in appearance, \ie{}, the (S+U) ablated datasets, the object segmentation performance is inferior. By contrast, the object segmentation accuracy can be greatly improved on the datasets only with appearance factors ablated, \ie{}, the (C+T) datasets. Meanwhile, more regular shapes and uniform scales of objects still have a significant positive influence on the success of object segmentation especially when the appearance factors are combined in ablated datasets. To be specific, AIR \cite{Eslami2016} is quite sensitive to the scale of objects apart from object color gradient and inter-object color similarity. MONet \cite{Burgess2019} can obtain comparable performance to the simple synthetic datasets once object color gradient and inter-object color similarity are ablated. All four factors are closely relevant the results of IODINE \cite{Greff2019} and SlotAtt \cite{Locatello2020}. 
        
\end{document}